\documentclass{tlp}
\nocite{*}

\submitted{4 October 2005}
\revised{21 March 2006}
\accepted{14 April 2006}

\usepackage{latexsym}
\usepackage{verbatim}
\usepackage{url}
\usepackage{epsfig}
\usepackage{graphics}
\usepackage{graphicx}
\usepackage{times}
\usepackage{url}
\usepackage{fancyvrb}  

\newtheorem{example}{Example}
\newtheorem{theorem}{Theorem}
\newtheorem{proposition}{Proposition}
\newtheorem{lemma}{Lemma}

\newtheorem{definition}{Definition}
\newtheorem{remark}{Remark}

\newtheorem{corollary}{Corollary}[section]

\newcommand{\http}[1]{\begin{small} {\url{#1}} \end{small}}

\def\ni{\noindent}

\def\beq{\begin{equation}}
\def\eeq#1{\label{#1}\end{equation}}
\def\la{\leftarrow}

\def\la{{\leftarrow}}
\def\naf{{\; not \;}}

\def\initially{\hbox{\bf initially}}
\def\causes{\hbox{\bf causes}}
\def\caused{\hbox{\bf if}}
\def\oneof{\hbox{\bf oneof}}
\def\executable{\hbox{\bf executable}}
\def\determines{\hbox{\bf determines}}

\def\naf{{not} \;}

\def\ni{\noindent}
\def\beq{\begin{equation}}
\def\eeq#1{\label{#1}\end{equation}}
\def\la{\leftarrow}

\def\qed{\hfill$\Box$}

\def\cala{{\cal A}^c}

\def\cald{{\cal D}}
\def\calg{{\cal G}}

\def\calp{{\cal P}}

\def\cali{{\cal I}}
\def\calp{{\cal P}}

\def\determines{\hbox{\bf determines}}

\def\after{\; \hbox{\bf after } \; }

\def\kwhether{\; \hbox{ \bf whether } \;}

\def\knows{\; \hbox{\bf knows \ }}

\def\dlvk{${\tt DLV}^{\tt K}$}
\def\sgp{SGP}

\def\cmbp{CMBP}
\def\cps{{\sc ascp}}
\def\pond{POND}
\def\pddl{PDDL}
\def\ack{${\cal A}^c_K$}
\def\ar{${\cal AR}$}

\def\kcases{{\bf cases}}
\def\smodels{{\tt smodels}}
\def\cmodels{{\tt cmodels}}

\def\reduct{{\tt reduct}}
\def\min{{\tt min}}
\def\limit{{\tt limit}}

\newcommand{\myneg}[1]{\neg #1}

\title[ASP with Sensing Actions, Incomplete 
Information, and Static Causal Laws]
{
Reasoning and Planning with Sensing Actions, Incomplete 
Information, and Static Causal Laws using Answer Set Programming
}

\author[Phan Huy Tu, Tran Cao Son, and Chitta Baral]
{
PHAN HUY TU and TRAN CAO SON\\
Department of Computer Science\\ 
New Mexico State University \\
PO Box 30001, MSC CS \\
Las Cruces, NM 88003, USA\\
\email{\{tphan,tson\}@cs.nmsu.edu}\\
\and
CHITTA BARAL\\
Department of Computer Science and Engineering\\ 
Arizona State University \\
Tempe, AZ 85287, USA\\
\email{chitta@asu.edu}
}

\begin{document}

\maketitle

\begin{abstract}
We extend the 0-approximation of sensing actions and incomplete 
information in \cite{sonbaral00} to action theories with static 
causal laws and prove its soundness with respect to the possible
world semantics. We also show that the conditional planning problem 
with respect to this approximation is {\bf NP}-complete. We then 
present an answer set programming based conditional planner, 
called \cps, that is capable of generating both 
conformant plans and conditional plans in the presence of sensing actions,
incomplete information about the initial state, and static causal laws.  
We prove the correctness of our implementation and argue that our 
planner is sound and complete with respect to the proposed approximation. 
Finally, we present experimental results comparing \cps\ to other planners.
\end{abstract}


\section{Introduction}
\label{sec1}

Classical planning assumes that agents have complete information
about the world. For this reason, it is often labeled as
unrealistic because agents operating in real-world environment
often do not have complete information about their environment.
Two important questions arise when one wants to remove this
assumption: {\em how to reason about the knowledge of agents} and
{\em what is a plan} in the presence of incomplete information.
The first question led to the development of several approaches to
reasoning about effects of sensing (or knowledge producing)
actions \cite{gold96b,lobo97,moo85b,ScherlL03,sonbaral00,thi00a}.
The second question led to the notions of {\em conditional plan}
and {\em conformant plan} whose execution is guaranteed to 
achieve the goal regardless of the values of unknown 
fluents in the initial situation. The former contains sensing actions
and conditionals such as the well-known ``if-then-else'' or ``cases''
construct, while the latter is just a sequence of actions. 
In this paper, we refer to {\em conditional planning} and
{\em conformant planning} as planning approaches that generate
conditional plans and conformant plans, respectively. 
We use {\em plan} as a generic term for both conditional and 
conformant plan when the distinction between the two is 
not important.

Approaches to conditional planning can be characterized by the
techniques employed in their search process or by the action
formalism that supports their reasoning process. Most of the early
conditional planners implemented a partial-order planning algorithm
\cite{gol98,gold96,pry96,peo92} and used Situation Calculus
or STRIPS as their underlying formalism in representing and reasoning about
actions and their effects. Among them, CoPlaS
\cite{lob98}, which is implemented in Sicstus Prolog, 
is a regression planner that uses a high-level action 
description language to represent
and reason about effects of actions, including sensing actions;
and FLUX \cite{thi00}, a constraint logic programming based planner, 
is capable of generating and verifying conditional
plans. Another conditional planner based on a QBF theorem prover 
was developed in \cite{rin99b}. Some other planners, for example, 
SGP \cite{weld98b} or POND \cite{bryceKS04}, extended the planning graph 
algorithm \cite{bf95} to deal with sensing actions.
The main difference between SGP and POND is that the former 
searches solutions within the planning graph, whereas 
the latter uses it as a means of computing the heuristic 
function.

Conformant planning \cite{gef98,brafman:hoffmann:icaps-04,CimattiRB04,castellini-cplan,eit03,smi98} is another
approach to deal with incomplete information. In conformant
setting, a solution is simply a sequence of actions that achieves the
goal from every possible initial situation. A recent study
\cite{CimattiRB04} shows that conformant planning based on model
checking is computationally competitive with other
approaches to conformant planning such as those based on heuristic
search algorithms \cite{gef98,brafman:hoffmann:icaps-04} or those that 
extend Graphplan
\cite{smi98}. A detailed comparison in \cite{eit03} demonstrates that 
a logic programming based conformant planner is able to compete 
with other approaches to planning. 

The most important difference between conditional planners 
and conformant planners lies in the fact that 
conditional planners can deal with sensing actions whereas
conformant planners cannot. Consequently, there are 
planning problems solvable by conditional planners but 
not by conformant planners. The following example demonstrates 
this issue. 
\begin{example}
\label{ex01} 
Consider a security window with a lock that behaves 
as follows. The window can be in one of the three 
states {\em opened, closed}\footnote{
  The window is closed and unlocked. 
}  or {\em locked}\footnote{
  The window is closed and locked.
}. 
When the window is closed or opened, 
pushing it {\em up} or {\em down} will {\em open} or 
{\em close} it respectively. 
When the window is closed or locked, flipping the lock will 
lock or close it respectively. 

Now, consider a security robot that needs to make sure that the window
is locked after 9 pm. Suppose that the robot has been told that
the window is not open (but whether it is locked or closed is unknown). 

Intuitively, the robot can achieve its goal by performing 
the following steps. First, (1) it checks the window 
to determine the window's status. If the window is closed, 
(2.a) it locks the window; otherwise (i.e., the window is 
already locked), simply (2.b) it does nothing.

Observe that no sequence of actions 
can achieve the goal from every possible initial situation. In other
words, {\em there exists no conformant plan} achieving the goal.
\qed
\end{example}

In this paper, we investigate the application of {\em answer set 
programming} (see e.g. \cite{Baral03,Lifschitz02,mar99,nie99}) in conformant and 
conditional planning. To achieve our goal, we 
first define an approximation semantic for action theories with 
static causal laws and sensing actions based on the 0-approximation 
in \cite{sonbaral00}. It is an alternative to the possible world 
semantics for reasoning about effects of actions in the 
presence of incomplete information and sensing actions \cite{moo85b}.
The basic idea of this approach is to {\em approximate} the set of 
possible world states by a set of fluent literals that is true 
in every possible world state. The main advantage of the 
approximation-based approach is its low complexity in reasoning 
and planning tasks ({\bf NP}-complete) comparing to those 
based on the possible world semantics 
$\Sigma_2\mathbf{P}$-complete \cite{baral:ijcai99:aij}. The trade-off
for this low complexity is incompleteness. As we will demonstrate in 
our experiments, this is not really an issue with the benchmarks in the 
literature. 

We prove that the entailment relationship 
for action theories based on this approximation is sound with respect
to the possible world semantics for action theories with incomplete 
initial situation. We then show that the planning problem 
with respect to the newly developed approximation is {\bf NP}-complete. 
This facilitates the development of \cps, an 
answer set programming based planner that is capable of generating 
both conditional and conformant plans. 
Given a planning problem instance with incomplete information 
about the initial situation and sensing actions,
we translate it into a logic program whose answer sets 
\cite{GL88}  -- 
which can be computed using existing answer set 
solvers (e.g. \cmodels \ \cite{cmodels2}, \smodels \ \cite{sim02},
{\bf dlv} \cite{dlvsystem}, ASSAT \cite{lin02a}, NoMore \cite{AngerKL02}, 
etc.)  --
correspond to conformant or conditional plans that satisfy the goal. 
We compare our planner against state-of-the-art planners.
The results of our experiments show that conditional and conformant 
planning based on answer set programming can be competitive with
other approaches. 
To the best of our knowledge, no answer set based conditional 
planner has been developed except a previous version of the planner
presented in an earlier version of this paper \cite{SonTB04}.

The paper is organized as follows. Section \ref{preliminaries}
presents the basics of an action language with sensing actions
and static causal laws, including its syntax and the 0-approximation, 
as well as the notions of conditional plans and queries. It also contains the 
complexity result of the conditional planning problem with respect to
the 0-approximation. Section \ref{planner} describes
a logic programming encoding of a conditional/conformant 
planner, called \cps. Section
\ref{property} discusses several properties of \cps. 
Section \ref{experiments} experimentally compares \cps\ with
some other state-of-the-art conformant/conditional planners.
Section \ref{conclusion} discusses some desirable extensions 
of the current work. The proofs of theorems and propositions 
are given in Appendices A and B. An example of encoding
is given in Appendix C.

\section{${\cal A}_K^c$ --- An Action Language with Sensing Actions and Static Causal Laws}
\label{preliminaries}
The representation language, $\cala_K$, for our planner
is an extension of the action language ${\cal A}_K$ in
\cite{sonbaral00}. While ${\cal A}_K$ extends the high-level action description 
language ${\cal A}$ from \cite{GL93a} by introducing two new types of
propositions called {\em knowledge producing proposition} and the
{\em executability condition}, $\cala_K$ extends ${\cal A}_K$
by adding {\em static causal laws} and allowing a sensing action to
sense more than one fluent. Loosely speaking, $\cala_K$ is a subset of
the language ${\cal L}_{DS}$ in \cite{bar00}. Nevertheless, 
like ${\cal A}_K$, ${\cal L}_{DS}$ considers sensing actions
that sense only one fluent. The semantics given for $\cala_K$ in this paper 
is an approximation of the semantics of  ${\cal L}_{DS}$.

\subsection{Action Language $\cala_K$ -- Syntax} 
The alphabet of an action theory in $\cala_K$ 
consists of a set of actions {\bf A} and a set of fluents {\bf F}.
A {\em fluent literal} (or {\em literal} for short) 
is either a fluent $f \in \mathbf{F}$ or its negation $\neg f$. 
$f$ and $\neg f$ are said to be complementary. For a literal $l$, 
by $\myneg{l}$, we mean its complement. A {\em fluent formula} is
a propositional formula constructed from the set of literals 
using operators $\wedge$, $\vee$, and/or $\myneg$.
To describe an action theory, propositions of the following 
forms are used:
\begin{eqnarray}
&  \initially(l) \label{init} \\
&  \executable(a,\psi) \label{exec}\\
&  \causes(a,l,\phi) \label{dynamic}\\
&  \caused(l,\varphi) \label{static}\\
&  \determines(a,\theta) \label{knowledge} 
\end{eqnarray}
where $a \in {\bf A}$ is an action, $l$ is a literal, 
and $\psi,\phi,\varphi,\theta$ are
sets of literals\footnote{A set of literals is
interpreted as the conjunction of its members. The empty set
$\emptyset$ denotes {\em true}.}. 

The initial situation is described by a set of propositions 
(\ref{init}), called {\em v-propositions}. (\ref{init})
says that $l$ holds in the initial situation. 
A proposition of form (\ref{exec}) is called {\em executability condition}. 
It says that $a$ is executable in
any situation in which $\psi$ holds (the precise meaning 
of {\em hold} will be given later). 
A proposition (\ref{dynamic}), called a {\em dynamic causal law}, 
represents a conditional effect of an action. 
It says that performing $a$ in a situation in which $\phi$
holds causes $l$ to hold in the successor situation. 
A proposition (\ref{static}), called a {\em static causal law}, 
states that $l$ holds in any situation in which $\varphi$ holds.
A {\em knowledge proposition} (or {\em k-proposition} for short) 
(\ref{knowledge}) states that the values of literals in $\theta$,
sometimes referred to as {\em sensed-literals}, 
will be known after $a$ is executed. Because the execution of 
$a$ will determine the truth value of at least one fluent, 
without loss of generality, we assume that $\theta$
contains at least two literals. Furthermore, we require that
if $\theta$ is not a set of two contrary literals $f$ and $\neg f$
then the literals in $\theta$ are mutually exclusive, i.e.,
\begin{enumerate}
\item 
for every pair of literals $g$ and $g'$ in $\theta$, $g \ne g'$, 
the theory contains the static causal law 
\[
\caused(\neg g',\{g\}) 
\]
and 

\item for every literal $g$ in $\theta$, the theory contains the 
static causal law
\[
\caused(g,\{\neg g' \mid g' \in \theta \setminus \{ g \}\}).
\]
\end{enumerate}
For convenience, we use the abbreviation 
$$\oneof(\theta)$$
to denote the above set of static causal laws. Apart from this,
we will sometime write   
$$
\determines(a,f)
$$
to stand for
$$
\determines(a,\{f,\neg f\}).
$$

Actions appearing in (\ref{dynamic}) and (\ref{knowledge}) are
called non-sensing actions and sensing actions, respectively. In
this paper, we assume that they are disjoint from each other. 
In addition, we also assume that each sensing action appears
in at most one k-proposition. 

An {\em action theory} is given
by a pair $(\cald,\cali)$ where $\cald$
is a set of propositions (\ref{exec})--(\ref{knowledge}) 
and $\cali$ is a set of propositions (\ref{init}). 
$\cald$ and $\cali$ are called the {\em
domain description} and {\em initial situation}, respectively.
A {\em planning problem instance} is a 3-tuple $(\cald,\cali,\calg)$, where
$(\cald,\cali)$ is an action theory and $\calg$ is a
{\em conjunction} of fluent literals. It is worth mentioning that
with a proper set of rules for checking the truth value of 
a fluent formula (see e.g. \cite{SonBTM05}), the framework 
and all results presented in this paper can be extended to 
allow $\calg$ to be an arbitrary fluent formula as well.

\begin{example}
\label{ex04}
The planning problem instance $\calp_1 = (\cald_1,\cali_1,\calg_1)$
in Example \ref{ex01} can be represented as follows.
\[
\cald_1 = \left \{
\begin{array}{lll}
\executable(check, \{\}) \\
\executable(push\_up, \{ closed\}) \\
\executable(push\_down, \{ open\}) \\
\executable(flip\_lock, \{ \neg open\}) \\
\\
\causes(push\_down, closed, \{\}) \\
\causes(push\_up, open,  \{\}) \\
\causes(flip\_lock, locked,  \{closed\}) \\
\causes(flip\_lock, closed,  \{locked\}) \\
\\
\oneof(\{open, locked, closed\})\\
\\
\determines(check, \{open,closed,locked\}) 
\end{array}
\right \}
\]
\[
\cali_1 \;\; = \; \left \{
\begin{array}{lll}
\initially(\neg open)
\end{array}
\right \}
\]
$$
\calg_1 = \{ locked \}
$$
\qed
\end{example}

\begin{remark}
For an action theory $(\cald, \cali)$, 
$\caused(l,\emptyset) \in \cald$ implies that 
literal $l$ holds in every situation.
Since $l$ is always true, queries about the truth value 
of $l$ (or $\neg l$) have a trivial answer and 
the theory can be simplified by removing all instances 
of $l$ in other propositions. 
Furthermore, if the theory also contains a 
dynamic law of the form $\causes(a,\neg l,\phi)$ then the 
execution of $a$ in a state satisfying $\phi$ 
will result in an inconsistent state of the world. 
Thus, the introduction of $l$ 
in the action theory is either redundant or erroneous. 
For this reason, without loss of generality, we will assume 
that action theories in 
this paper do not contain any static causal law (\ref{static}) 
with $\varphi = \emptyset$.
\end{remark}

\begin{remark}
Since an empty plan can always be used to achieve an empty goal,
we will assume hereafter that planning problem instances considered 
in this paper have non-empty goals.
\end{remark}
\subsection{Conditional Plan}
\label{akquery}
In the presence of incomplete information and sensing 
actions, we need to extend the notion of a plan from a sequence
of actions so as to allow conditional statements such as
{\bf if-then-else}, {\bf while-do}, or {\bf case-endcase} (see e.g.
\cite{lev96,lobo97,sonbaral00}). Notice that an if-then-else 
statement can be replaced by a case-endcase statement. 
Besides, if we are only interested in plans with bounded 
length then whatever can be represented by a while-do 
statement with a non-empty body can also be represented 
by a set of case-endcase statements as well. 
Therefore, in this paper, we limit ourselves 
to conditional plans with the case-endcase construct only. 
Formally, we consider conditional plans defined as follows.
We note that our notion of conditional plans in this paper is 
fairly similar to the ones introduced in \cite{lev96,lobo97,sonbaral00}. 
\begin{definition}[Conditional Plan]
\label{condplan}
\begin{itemize}
\item[1. ]
$[]$ is a conditional plan, denoting the empty plan, i.e.,
the plan containing no action.
\item[2. ]
if $a$ is a non-sensing action and $p$ is a conditional plan then
$[a;p]$ is a conditional plan.
\item[3. ]
if $a$ is a sensing action with proposition (\ref{knowledge}),
where $\theta = \{ g_1, \dots, g_n \}$, and $p_j$'s 
are conditional plans then $[a; \kcases(\{g_j \rightarrow p_j\}_{j=1}^{n})]$
is a conditional plan.
\item[4. ]
Nothing else is a conditional plan.
\end{itemize}
\end{definition}

By this definition, clearly a sequence of actions is also a 
conditional plan. The execution of a conditional plan of the 
form $[a;p]$, where $a$ is a non-sensing action and $p$ is another 
conditional plan, is done sequentially,
i.e., $a$ is executed first, followed by $p$. 
To execute a conditional plan of the form 
$[a; \kcases(\{g_j \rightarrow p_j\}_{j=1}^{n})]$, 
we first execute $a$ and then
evaluate each $g_j$ with respect to our current knowledge. If 
one of the $g_j$'s, say $g_k$, holds, we execute 
the corresponding sub-plan $p_k$. Observe that because 
fluent literals in $\theta$ are mutual exclusive, such 
$g_k$ uniquely exists. 

\begin{example}
\label{ex07}
The following are conditional plans
of the action theory in Example \ref{ex04}: 
$$
p_1 = [push\_down;flip\_lock] \label{plan_1}
$$
$$p_2 = 
check; \;  
\kcases \left (
\begin{array}{lll} 
open & \rightarrow & [] \\
closed & \rightarrow & [flip\_lock] \\
locked & \rightarrow & []
\end{array}
\right)
$$
$$
p_3 = 
check; \;  
\kcases \left (
\begin{array}{lll} 
open & \rightarrow & [push\_down;flip\_lock] \\
closed & \rightarrow & [flip\_lock;flip\_lock;flip\_lock] \\
locked & \rightarrow & []
\end{array}
\right)
$$
$$
p_4 = 
check; \;  
\kcases \left (
\begin{array}{lll} 
open & \rightarrow & [] \\
closed & \rightarrow & p_2 \\
locked & \rightarrow & []
\end{array}
\right)
$$
Among those, $p_2$, $p_3$ and $p_4$ are conditional
plans that achieve the goal $\calg_1$\footnote{Note that $p_2$ and $p_4$
can achieve the goal because the first case 
``the window is open'' {\em cannot} happen}.
\qed
\end{example}
In the rest of the paper, the terms ``plan'' and 
``conditional plan'' will be used alternatively.

\subsection{Queries}
\label{query}
A query posed to an $\cala_K$ action theory $(\cald,\cali)$ is of 
the form 
\begin{equation} \label{query1}
\knows \rho \after p 
\end{equation}
or 
\begin{equation} \label{query2}
\kwhether \rho \after p 
\end{equation}
where $p$ is a conditional plan and $\rho$ is a fluent formula.
Intuitively, the first (resp. second) 
query asks whether $\rho$ is true (resp. known) 
after the execution of $p$ from the initial situation.

\subsection{0-Approximation Semantics of $\cala_K$}
\label{approx}

We now define an approximation semantics of $\cala_K$, called
0-approximation, which extends 
the 0-approximation in \cite{sonbaral00} to deal with static causal laws. 
It is  defined by a transition function $\Phi$ that maps actions and 
a-states into sets of a-states (the meaning of a-states will follow).
Before providing the formal definition of 
the transition function, we introduce some notations
and terminology.

For a set of literals $\sigma$, $\myneg{\sigma}$ denotes the 
set $\{\myneg{l} \mid l \in \sigma \}$. 
$\sigma$ is said to be {\em consistent} if it does 
not contain two complementary literals. 
A literal $l$ (resp. set of literals $\gamma$) {\em holds} in a  
set of literals $\sigma$
if $l \in \sigma$ (resp. $\gamma \subseteq
\sigma$); $l$ (resp. $\gamma$) {\em possibly holds} in $\sigma$ if
$\myneg{l} \not \in \sigma$ (resp. $\myneg{\gamma} \cap \sigma =
\emptyset$).

Given a consistent set of literals $\sigma$, 
the truth value of a formula $\rho$, denoted by $\sigma(\rho)$, 
is defined as follows. If $\rho \equiv l$ for some literal $l$
then $\sigma(\rho) = {\tt T}$ if $l \in \sigma$; 
$\sigma(\rho) = {\tt F}$ if $\myneg{l} \in \sigma$;
$\sigma(\rho) = {\tt unknown}$ otherwise.
If $\rho \equiv \rho_1 \wedge \rho_2$ then $\sigma(\rho) = {\tt T}$
if $\sigma(\rho_1) = {\tt T}$ and $\sigma(\rho_2) = {\tt T}$;
$\sigma(\rho) = {\tt F}$ if $\sigma(\rho_1) = {\tt F}$ 
or $\sigma(\rho_2) = {\tt F}$;
$\sigma(\rho) = {\tt unknown}$ otherwise.
If $\rho \equiv \rho_1 \vee \rho_2$ then $\sigma(\rho) = {\tt T}$
if $\sigma(\rho_1) = {\tt T}$ or $\sigma(\rho_2) = {\tt T}$;
$\sigma(\rho) = {\tt F}$ if $\sigma(\rho_1) = {\tt F}$ and 
$\sigma(\rho_2) = {\tt F}$;
$\sigma(\rho) = {\tt unknown}$ otherwise.
If $\rho \equiv \neg \rho_1$ then $\sigma(\rho) = {\tt T}$
if $\sigma(\rho_1) = {\tt F}$;
$\sigma(\rho) = {\tt F}$ if $\sigma(\rho_1) = {\tt T}$;
$\sigma(\rho) = {\tt unknown}$ otherwise.

We say that $\rho$ is known to be true (resp. false) in $\sigma$ and 
write $\sigma \models \rho$ (resp. $\sigma \models \myneg \rho$)
if $\sigma(\rho) = {\tt T}$ (resp. $\sigma(\rho) = {\tt F}$).
When $\sigma \models \rho$ or $\sigma \models \neg \rho$ we say that
$\rho$ is {\em known} in $\sigma$; otherwise, $\rho$ is {\em unknown}
in $\sigma$. 
We will say that $\rho$ holds in $\sigma$ if it is
 known to be true in $\sigma$.

A set of literals $\sigma$ satisfies a static causal law (\ref{static}) if 
either (i) $\varphi$ does not hold in $\sigma$; or (ii)
$l$ holds in $\sigma$ (i.e., $\varphi$ holds in $\sigma$ implies 
that $l$ holds in $\sigma$).
By $Cl_{\cald}(\sigma)$, we denote the smallest 
set of literals that includes $\sigma$ and 
satisfies all static causal laws in ${\cald}$. 
Note that $Cl_{\cald}(\sigma)$ might be inconsistent but it
is unique ({\em see} Lemma \ref{lm1}, Appendix A).

An {\em interpretation} $I$ of a domain description $\cald$ 
is a complete and consistent set of literals in $\cald$, i.e.,
for every fluent $f \in \mathbf{F}$, 
(i) $f \in I$ or $\neg f \in I$; and 
(ii) $\{f, \neg f\} \not \subseteq I$. 

A {\em state} $s$ is an interpretation satisfying
all static causal laws in $\cald$.
An action $a$ is {\em executable} in $s$ if 
there exists an executability condition (\ref{exec}) such that 
$\psi$ holds in $s$. For a non-sensing action $a$ executable 
in $s$, let
\begin{eqnarray}
& E(a,s) = \{ l \mid \exists \textnormal{ a dynamic 
causal law } (\ref{dynamic}) \textnormal{ such that } 
\phi \textnormal{ holds in } s \} \label{e_a_s}
\end{eqnarray}
The set $E(a,s)$ is often referred to as the {\em direct effects} of 
$a$. When the agent has complete information about the world,
the set of possible next states after 
the execution of $a$ in $s$, denoted by $Res_{\cald}^c(a,s)$, 
is defined as follows.
\begin{definition}
[Possible Next States, \cite{mcc95a}]
\label{def_compl}
Let ${\cald}$ be a domain description. For any state $s$
and non-sensing action $a$ executable in $s$, $Res_{\cald}^c(a,s) =
\{ s' \mid s' \textnormal{ is a state such
that } s' = Cl_{\cald}(E(a,s) \cup (s \cap s'))\}$.
\end{definition}
The intuitive meaning of this definition is that
a literal $l$ holds in a possible next state
$s'$ of $s$ after $a$ is executed iff either
(i) it is a direct effect of $a$, i.e., $l \in E(a,s)$
(ii) it holds by inertia, i.e., $l \in (s \cap s')$, or
(iii) it is an indirect effect\footnote{Indirect effects
are those caused by static causal laws.} of $a$, i.e., $l$ holds
because of the operator $Cl_{\cald}$.

Note that the $Res^c_{\cald}$-function can be 
{\em non-deterministic}, i.e.,
 $Res^c_{\cald}(a,s)$ might contain more than one element.
The following example illustrates this point.
\begin{example}
\label{ex10}
Consider the following domain description
\[
\cald_2 = \left \{
\begin{array}{lll}
\executable(a, \{\}) \\
\causes(a, f, \{\}) \\
\caused(g, \{f,\myneg h\}) \\
\caused(h, \{f,\myneg g\})\\
\caused(k, \{\myneg f\})
\end{array}
\right \}
\]
Let $s = \{ \myneg f, \myneg g, \myneg h, k \}$. Clearly $s$ is
a state since it satisfies all static laws in $\cald_2$. Executing
$a$ in $s$ results in two possible next states
$$Res^c_{\cald_2}(a,s) = \{\{f, \myneg g, h, k\},
\{f, g, \myneg h, k\}\}$$
In the first possible next state $s_1 = \{f, \myneg g, h, k\}$,
$f$ holds because it is a direct effect of $a$, i.e., $f \in
E(a,s)$; $\neg g$ and $k$ hold because of inertia ($
s \cap s_1 = \{\neg g,k\}$); and $h$ holds because it is an 
indirect effect of $a$ (in particular, $h$ holds because of the 
static causal law $\caused(h, \{f,\myneg g\})$).

Likewise, we can explain why each literal in the second possible
next state holds.
\qed
\end{example}
\begin{definition}
[Consistent Domains]
\label{def_cons}
A domain description $\cald$ is {\em consistent} if
for every state $s$ and action $a$ executable in $s$,
$Res_{\cald}^c(a,s) \ne \emptyset$.
\end{definition}
In the presence of incomplete information, an agent, however,
does not always know exactly which state it is currently in. 
One possible way to deal with this problem is to represent 
the agent knowledge by a set of possible states (a.k.a. belief state)  
that are consistent with the agent's current knowledge and extend 
Definition \ref{def_compl} to define a mapping from 
pairs of actions and belief states into belief states 
as in \cite{bar00}. The main problem with this approach is its
high complexity \cite{baral:ijcai99:aij}, even for the computation
of what is true/false after the execution of one action.
We address this problem by defining an approximation 
of the set of states in Definition \ref{def_compl} as follows.

First, we relax the notion of a state in Definition \ref{def_compl} 
to be an approximate state defined as follows.
\begin{definition} 
[Approximate State]
A consistent set of literals $\delta$ is called 
an approximate state (or {\em a-state}, for short) 
if $\delta$ satisfies all static causal laws in $\cald$.
\end{definition}
Intuitively, $\delta$ represents the (possibly incomplete) 
current knowledge of the agent, i.e., it contains all fluent 
literals that are known to be true to the agent. 
When $\delta$ is a subset of some state $s$, we say that 
it is {\em valid}. An action $a$ is {\em executable} in $\delta$ if
there exists an executability condition (\ref{exec})
in $\cald$ such that $\psi$ holds in $\delta$.

Next, we define what are the possible next a-states after 
the execution of an action $a$ in a given a-state $\delta$,
provided that $a$ is executable in $\delta$. 
Consider the case that $a$ is a non-sensing action. Let
\begin{eqnarray}
& e(a,\delta) = Cl_{\cald}(\{ l \mid \exists \textnormal{ a dynamic 
causal law } (\ref{dynamic}) \textnormal{ such that } 
\phi \textnormal{ holds in } \delta \}) \label{e_a}
\end{eqnarray}
and
\begin{eqnarray} \label{sem-pc1}
& pc(a,\delta) = \bigcup_{i=0}^{\infty} pc^i(a,\delta) \label{pc_a}
\end{eqnarray}
where
\begin{eqnarray} \label{sem-pc2}
pc^0(a,\delta) = \{ l \mid 
\exists \textnormal{ a dynamic causal law }
(\ref{dynamic}) \textnormal{ s.t. }
l \not \in \delta \textnormal{ and } \phi
\textnormal{ possibly holds in } \delta \} \label{pc_a_0}
\end{eqnarray}
and for $i \ge 0$,
\begin{eqnarray} \label{sem-pc3}
pc^{i+1}(a,\delta) = pc^i(a,\delta) \cup &
\{ l \mid \exists \textnormal{ a static causal law }
(\ref{static}) \textnormal{ s.t. }
l \not \in \delta, \varphi \cap pc^i(a,\delta)
\ne \emptyset, \nonumber \\
& \textnormal{ and } \varphi 
\textnormal{ possibly holds in } e(a,\delta) 
\label{pc_a_k} \}
\end{eqnarray}
Intuitively, $e(a,\delta)$ and $pc(a,\delta)$ denote what 
{\em definitely holds} and what {\em may change} in the next situation
respectively \footnote{Note that the operator $Cl_{\cald}$ is used
in the definition of $e(a,\delta)$ to {\em maximize} what 
definitely holds in the next situation.}. Specifically, 
$l \in e(a,\delta)$ means that $l$ holds in the next situation and 
$l \in pc(a,\delta)$ means that $l$ is not in $\delta$ but possibly holds 
in the next situation.
This implies that $\delta \setminus \neg pc(a,\delta)$ is an approximation 
of the set of literals that hold by inertia after the execution of $a$
in $\delta$. Taking into account the effects of the static causal 
laws, we have that the set of literals 
$
\delta' = Cl_{\cald}(e(a,\delta) {\cup}
(\delta {\setminus} \neg pc(a,\delta)))
$ 
must hold in the next situation. This leads us to the following
definition of the possible next a-states after 
a non-sensing action gets executed. 

\begin{definition} 
[0-Result Function] \label{res}
For every a-state $\delta$ and non-sensing action $a$ executable 
in $\delta$, let $$\delta' = Cl_{\cald}(e(a,\delta) {\cup}
(\delta {\setminus} \neg pc(a,\delta))).$$ 
Define 
\begin{enumerate}

\item $Res_\cald(a,\delta) = \{\delta'\}$ 
if  $\delta'$  is consistent. 

\item $Res_\cald(a,\delta) =  \emptyset$ 
if  $\delta'$  is inconsistent.
\end{enumerate}
\end{definition}
The next examples illustrate this definition.
\begin{example}
\label{ex13}
{\rm
Consider the domain description ${\cald}_1$ in Example \ref{ex04}.
Let $\delta = \{\neg open, closed, \neg locked\}$. 
We can easily check that $\delta$ is an a-state of $\cald_1$.
We have 
\[
e(flip\_lock,\delta) = Cl_{\cald_1}(\{ locked \}) =
\{ \neg open, \neg closed, locked \}
\]
and
\[
pc^0(flip\_lock,\delta) = \{ locked \}
\]
Because $\caused(\neg open,\{locked\}) \in \cald_1$,
and $\caused(\neg closed,\{locked\}) \in \cald_1$,
by (\ref{pc_a_k}), we have
$$pc^1(flip\_lock,\delta) = \{locked, \neg closed \}$$
Note that $\neg open \not \in pc^1(flip\_lock,\delta)$ because
it is already in $\delta$.

It is easy to see that $pc^i(flip\_lock,\delta) = pc^1(flip\_lock,\delta)$
for all $i > 1$. Hence, we have
$$pc(flip\_lock,\delta) = 
\bigcup_{i=0}^{\infty} pc^i(flip\_lock,\delta) =
\{\neg closed, locked\}$$
Accordingly, we have
$$Res_{\cald_1}(flip\_lock,\delta) = \{ Cl_{\cald_1}(e(flip\_lock,\delta) 
\cup (\delta \setminus \neg pc(flip\_lock,\delta))) \} = $$
$$\{ Cl_{\cald_1}(\{\neg open, \neg closed, locked\}) \} =
\{\{\neg open, \neg closed, locked\}\}$$
\qed}
\end{example}
\begin{example}
\label{ex14}
For the domain description $\cald_2$ in Example \ref{ex10},
we have
$$e(a,s) = Cl_{\cald_2}(\{ f \})=\{ f \}$$
$$pc^0(a,s) = \{ f \}$$
As $\caused(g,\{f,\neg h\}) \in \cald_2$ and
$\caused(h,\{f,\neg g\}) \in \cald_2$, we have
$$pc^1(a,s) = \{ f, g, h \}$$
Note that $k \not \in pc^1(a,s)$ since $\neg f$ does not
hold in $e(a,s)$. We can check that
$pc^i(a,s) = pc^1(a,s)$ for all $i > 1$. Hence, we have
$$pc(a,s) = \{ f,g,h\}$$
As a result, we have
$$Res_{\cald_2}(a,s) = \{ Cl_{\cald_2}(e(a,s) 
\cup (s \setminus \neg pc(a,s))) \} = \{ Cl_{\cald_2}(\{ f,k \}) \}
= \{ \{ f,k \} \}$$
\qed
\end{example}
The following proposition shows that when a non-sensing
action is executed, the $Res$-function is {\em deterministic}
in the sense that it returns at most one possible next a-state;
furthermore, it is ``sound'' with respect to the $Res^c$-function.

\begin{proposition}
\label{prop-res-nonsense}
Let $\cald$ be a consistent domain description. For any state $s$,
a-state $\delta \subseteq s$, and non-sensing action $a$
executable in $\delta$, there exists an a-state $\delta'$
such that (i) $Res_\cald(a,\delta) = \{ \delta' \}$, and
(ii) $\delta'$ is a subset of every state $s' \in Res^c_\cald(a,s)$.
\end{proposition}
\begin{proof}
{\em see Appendix A}.
\end{proof}

We have specified what are the possible next a-states after
a non-sensing action is performed. Let us move to the case 
when a sensing action is executed. Consider an a-state
$\delta$ and a sensing action $a$ with k-proposition (\ref{knowledge})
in $\cald$. Intuitively, after $a$ is executed, the agent will know
the values of literals in $\theta$. Thus, the
set of possible next a-states can be defined as follows.
\begin{definition} 
[0-Result Function] \label{res-sense}
For every a-state $\delta$ and sensing action $a$ 
with proposition (\ref{knowledge}) such that $a$ is executable 
in $\delta$ , 
$$
Res_\cald(a,\delta) = \{Cl_{\cald}(\delta \cup \{g \}) \mid
g \in \theta \textnormal{ and }
Cl_{\cald}(\delta \cup \{g \}) \textnormal{ is consistent}\}
$$
\end{definition}
Roughly speaking, executing $a$ will result in several possible 
next a-states, in each of which exactly one sensed-literal in $\theta$
holds. However, some of them might be inconsistent with what is currently 
known. For example, if the security robot in Example \ref{ex01} 
knows that the window is not open then after it {\em checks} the window,
it should not consider the case that the window is {\em open} because
this is inconsistent with its current knowledge. 
Thus, in defining the set of possible next a-states resulting from the 
execution of a sensing action, we need to exclude such inconsistent 
a-states. The following example illustrates this.
\begin{example}
\label{ex16}
Consider again the domain description ${\cald}_1$ in 
Example \ref{ex04} and an a-state $\delta_1 = \{\neg open\}$.

We have
\[
Cl_{\cald_1}(\delta_1 \cup \{open\}) = 
\{ open, \neg open, closed, \neg closed, locked, \neg locked\} = \delta_{1,1}
\]
\[
Cl_{\cald_1}(\delta_1 \cup \{closed\}) = 
\{ \neg open, closed, \neg locked \} = \delta_{1,2}
\]
\[
Cl_{\cald_1}(\delta_1 \cup \{locked\}) = 
\{ \neg open, \neg closed,  locked \} = \delta_{1,3}
\]
Among those, $\delta_{1,1}$ is inconsistent. Therefore,
we have
$$Res_{\cald_1}(check,\delta_1)=\{\delta_{1,2},\delta_{1,3}\}$$
\qed
\end{example}
The next proposition shows that if a sensing action is 
performed in a valid a-state then the set of possible
next a-states will contain at least one valid a-state.
This corresponds to the fact that if the current knowledge 
of the world of the agent is consistent with the state of the world,
it will remain consistent with the state of the world after 
the agent acquires additional knowledge through the execution of a
sensing action. 
\begin{proposition}
\label{prop-res-sense}
Let $\cald$ be a consistent domain description. For any a-state 
$\delta$, and a sensing action $a$ executable in $\delta$, 
if $\delta$ is valid then $Res_\cald(a,\delta)$ contains
at least one valid a-state.
\end{proposition}
\begin{proof}
{\em see Appendix A}.
\end{proof}
The transition function $\Phi$ that maps actions 
and a-states into sets of a-states is defined as follows.
\begin{definition}
[Transition Function]
\label{trans}
Given a domain description $\cald$, for any action $a$ and a-state $\delta$, 
\begin{itemize}
\item[1.] if $a$ is not executable in $\delta$
then 
$$\Phi(a,\delta) = \perp$$
\item[2.] otherwise,
$$\Phi(a,\delta) = Res_\cald(a,\delta)$$
\end{itemize}
\end{definition}

The transition function $\Phi$ returns the set
of possible next a-states after performing a single 
action in a given a-state. We now extend it to define 
the set of possible next a-states after the execution of a plan. 
The extended transition function, called $\hat\Phi$, is 
given in the following definition.

\begin{definition}
[Extended Transition Function]
\label{d:weak:extend}
Given a domain description $\cald$, for any plan $p$ and 
a-state $\delta$,
%
\begin{itemize}
\item[1.]
if $p = []$ then
$$\hat\Phi(p,\delta) =\{\delta\}$$
\item[2.]
if $p = [a;q]$, where $a$ is a non-sensing action 
and $q$ is a sub-plan, then
$$\hat\Phi(p, \delta) =
\left \{
\begin{array}{ll}
\bot & \textnormal{if } \Phi(a,\delta) = \bot\\
\bigcup_{\delta' \in \Phi(a,\delta)} \hat\Phi(q, \delta') & 
\textnormal{otherwise} \\
\end{array}
\right.
$$
\item[3.]
if $p = [a;\kcases(\{g_j \rightarrow p_j\}_{j=1}^n)]$,
where $a$ is a sensing action and $p_j$'s are sub-plans, then
$$\hat\Phi(p, \delta) = 
\left \{
\begin{array}{ll}
\bot & \textnormal{if } \Phi(a,\delta) = \bot\\
\bigcup_{1 \le j \le n, \delta' \in
\Phi(a,\delta), g_j \textnormal{\footnotesize{ holds in }} \delta'} 
\hat \Phi(p_j,\delta') & 
\textnormal{otherwise} \\
\end{array}
\right.
$$
\end{itemize}
where, by convention, $\dots \cup \bot \cup \dots = \bot$.
\end{definition}

Items (2) and (3) of the above definition deserve some elaboration. 

\begin{remark}
\label{rem-nonsense}
During the execution of a plan $p$, when a non-sensing action 
$a$ is encountered (Item 2), by Definitions
\ref{res} and \ref{trans}, there are three possibilities:
$\Phi(a,\delta) = \bot$, $\Phi(a,\delta) = \emptyset$,
or $\Phi(a,\delta) = \{ \delta' \}$ for some a-state $\delta'$.
If the first case occurs then the result of execution of
$p$ in $\delta$ by the definition is also $\bot$. In this case, we say that
$p$ is not executable in $\delta$; otherwise, $p$ is {\em executable}
in $\delta$. If the second case occurs then by the definition,
$\hat \Phi(p,\delta) = \emptyset$. One may notice that, by
Proposition \ref{prop-res-nonsense}, this case takes place
only if there exists no state $s$ such that $\delta \subseteq
s$ (i.e., $\delta$ is invalid), or the domain is inconsistent. 
When $\Phi(a,\delta) = \{ \delta' \}$,
then the result of the execution of $p$ in $\delta$ is exactly 
as the result of the execution of the rest of $p$ in $\delta'$.
\end{remark}
\begin{remark}
\label{rem-sense}
If $p = [a;\kcases(\{g_j \rightarrow p_j\}_{j=1}^n)]$,
where $a$ is a sensing action and $p_j$'s are sub-plans (Item 3),
and $\Phi(a,\delta) \ne \bot$ then by 
Definitions \ref{res-sense} and \ref{trans}, we know that 
$\Phi(a,\delta)$ may contain several a-states $\delta_j$'s.
Each $\delta_j$ corresponds to an a-state in which literal $g_j$
holds. Therefore, we define $\hat \Phi(p,\delta)$ to be 
the union of the sets of possible a-states that are the 
results of the execution of $p_j$ in $\delta_j$. Note that
when we add $g_j$ to the current state $\delta$ to generate
$\delta_j$, we {\em assume} that $g_j$ holds. However, if
later on, during the execution of the rest of $p$, which
is $p_j$, we discover that $\hat \Phi(p_j,\delta_j) = \emptyset$, then 
our assumption about $g_j$ is not correct. Therefore, such a
$\delta_j$ contributes nothing to the set of possible a-states of  
$\hat\Phi(a,\delta)$. To see how this can happen, consider
the following domain description

\[
\cald_3 = \left \{
\begin{array}{lll}
\executable(a, \{\}) \\
\executable(b, \{\}) \\
\\
\causes(b,h, \{\}) \\
\caused(f,\{g,h\})\\
\caused(f,\{g,\neg h\})\\
\determines(a,f)  
\end{array}
\right \}
\]

and suppose that the set of fluents is $\{f,g,h\}$.
Let us see what are the final possible a-states after the execution
of plan $p = [a;\kcases(\{f \rightarrow b; \neg f \rightarrow b\})]$
in a-state $\delta = \{ g \}$ as defined by the extended transition function.

When $a$ is performed, we generate two possible
next a-states $\delta_1 = \{g, f \}$, and $\delta_2 = \{ g, \neg f\}$.
Executing $b$ in $\delta_2$ results in no possible next a-state
because $Cl_{\cald_3}(\{g,\neg f,h\}) = \{g,\neg f,h,f\}$ 
is not consistent. This means that $\Phi(b,\delta_2)$, and thus
$\hat \Phi([b],\delta_2)$, become $\emptyset$. Therefore, the 
set of possible final a-states is 
$\hat \Phi(p,\delta) = \hat \Phi([b],\delta_1)= \{\{f, g, h \}\}$.

Note that in this example, we did not notice that
$\delta_2$ is inconsistent at the time the action $a$
was performed. Rather, its inconsistency was only realized after 
the execution of $b$. In other words, our assumption that
$\neg f$ holds was not correct.

Similarly to the execution of a non-sensing action, when a 
sensing action $a$ is performed, by Proposition \ref{prop-res-sense}, 
$\Phi(a,\delta) = \emptyset$ only if the domain is inconsistent or 
$\delta$ is invalid.
\end{remark}
The above remarks imply that in some cases, for a plan $p$ and
an a-state $\delta$, $\hat \Phi(p,\delta)$ may be
empty. Intuitively, this is because
either $\delta$ is invalid or the domain is inconsistent.
We will show that under reasonable assumptions about $\delta$
and the domain, this cannot happen.
\begin{definition}
[Consistent Action Theories]
An action theory $(\cald,\cali)$ is {\em consistent} if
$\cald$ is consistent and its initial a-state, defined 
by $Cl_{\cald}(\{l \mid \initially(l) \in \cali\})$,
is valid. 
\end{definition}
The next proposition says that the execution of an executable 
plan from a valid a-state of a consistent action theory 
will result in at least one valid a-state. 
\begin{proposition}
\label{prop-cons-action-theory}
Let $(\cald,\cali)$ be a consistent action theory and let
$\delta$ be its initial a-state. For every conditional 
plan $p$, if $\hat \Phi(p,\delta) \ne \bot$ 
then $\hat \Phi(p,\delta)$ contains at least
one valid a-state.
\end{proposition}

\begin{proof}
{\em see Appendix A}.
\end{proof}
The above proposition implies that if the action theory 
$(\cald,\cali)$ is consistent and $\delta$ is
its initial a-state then the execution of $p$ in $\delta$
will yield at least a valid trajectory\footnote{A trajectory is 
an alternate sequence of a-states and actions, 
$\delta_0 a_1 \delta_1 a_2 \ldots a_n \delta_n$, such that 
$\delta_i \in \Phi(a_i, \delta_{i-1})$ for $i=1,\ldots,n$;
A trajectory is valid if $\delta_i$'s are valid a-states.},
provided that $p$ is executable in $\delta$. This is consistent
with the fact that if the initial a-state is complete (i.e.,
if we have complete information) then the execution of an  
executable plan in the initial a-state 
would return a valid trajectory. {\em From now on, we only consider 
consistent action theories.} 

We next define the entailment 
relationship between action theories and queries.

\begin{definition}
[Entailment]
\label{0ent}
Let $(\cald,\cali)$ be an action theory and $\delta$ be its 
initial a-state. For a plan $p$ and a fluent formula $\rho$, we say that 
\begin{itemize}
\item 
$(\cald,\cali)$ entails the query $\knows \rho \after p$ and write
\[
\cald \models_{\cali} \knows \rho \after p
\]
if $\hat\Phi(p,\delta) \ne \bot$ and $\rho$ is true in every
a-state in $\hat\Phi(p,\delta)$; and 
\item 
$(\cald,\cali)$ entails the query $\kwhether \rho \after p$ and write
\[
\cald \models_{\cali} \kwhether \rho \after p
\]
if $\hat\Phi(p,\delta) \ne \bot$ and $\rho$ is known in 
every a-state in $\hat\Phi(p,\delta)$.
\end{itemize}
%
\end{definition}
\begin{example}
\label{ex19} 
For the action theory $(\cald_1,\cali_1)$ in
Example \ref{ex04}, we will show that 
\begin{equation}
\cald_1 \models_{\cali_1} \knows locked \after p_2
\label{ex04_plan_2}
\end{equation}
where $p_2$ is given in Example \ref{ex07}.

Let $p_{2,1} = []$, $p_{2,2} = [flip\_lock]$ and
$p_{2,3} = []$. It is easy to see that the initial a-state 
of $(\cald_1,\cali_1)$ is $\delta_1 = \{ \neg open\}$.

It follows from Example \ref{ex16} that
$$\Phi(check,\delta_1) = \{ \delta_{1,2}, \delta_{1,3} \}$$

On the other hand, we have
$$\hat \Phi(p_{2,2},\delta_{1,2}) = 
\{ \{ locked, \neg open, \neg closed \} \}$$
and
$$\hat \Phi(p_{2,3},\delta_{1,3}) = 
\{ \{ locked, \neg open, \neg closed \} \}$$

Therefore, we have
$$\hat \Phi(p_2,\delta_1) = \hat \Phi(p_{2,2},\delta_{1,2})
\cup \hat \Phi(p_{2,3},\delta_{1,3}) = \{ 
\{ locked, \neg open, \neg closed \} \}$$

Since $locked$ is true in $\{ locked, \neg open, \neg closed \}$,
we have (\ref{ex04_plan_2}) holds.
On the other hand, because $closed$ is false in
$\{ locked, \neg open, \neg closed \}$, we have
\begin{eqnarray*}
\cald_1 \not\models_{\cali_1} \knows closed \after p_2\;\;\;\;\;\; 
\textnormal{ but  } \;\;\;\;\;\; 
\cald_1 \models_{\cali_1} \knows \neg closed \after p_2.
\end{eqnarray*}
Likewise, we can prove that
\begin{eqnarray*}
\cald_1 \models_{\cali_1} \knows locked \after p_3 
\;\;\;\;\;\; 
\textnormal{ and  } \;\;\;\;\;\; 
\cald_1 \models_{\cali_1} \knows locked \after p_4. 
\end{eqnarray*}
\end{example}
\begin{definition}
[Solutions]
\label{solutions}
A plan $p$ is called a {\em solution} to 
a planning problem instance $\calp = (\cald,\cali,\calg)$ 
iff $$\cald \models_{\cali} \knows \calg \after p$$
When $p$ is a solution to $\calp$, we say that $p$ 
is a plan that {\em achieves} the goal $\calg$.
\end{definition}
According to this definition, it is easy to see that 
plans $p_2$, $p_3$, and $p_4$ in Example \ref{ex07}
are solutions to $\calp_1 = (\cald_1,\cali_1,\calg_1)$ in
Example \ref{ex04}.

\subsection{Properties of the 0-Approximation}

We will now discuss some properties of the 0-approximation. 
For a domain description $\cald$, we define the size
of $\cald$ to be the sum of {\em (1)} the number of fluents;
{\em (2)} the number of actions; and {\em (3)} the number of 
propositions in $\cald$. 
The size of a planning problem instance $\calp = (\cald,\cali,\calg)$ 
is defined as the size of $\cald$.  The size of a plan $p$, 
denoted by $size(p)$, is defined as follows.  
\begin{enumerate}
\item $size([]) = 0$; 
\item $size([a;p]) = 1 + size(p)$ 
if $a$ is a non-sensing action and $p$ is a plan; and
\item $size([a; \kcases(\{g_j \rightarrow p_j\}_{j=1}^{n})]) 
= 1 + \Sigma^n_{j=1} (1 + size(p_j))$ 
if $a$ is a sensing action and $p_j$'s 
are plans. 
\end{enumerate}
Then, we have the following proposition.
\begin{proposition}
\label{prop-trans1}
For a domain description $\cald$, an action $a$, 
and an a-state $\delta$, computing $\Phi(a,\delta)$ 
can be done in polynomial time in the size of $\cald$. 
\end{proposition}
\begin{proof}
{\em see Appendix A.}
\end{proof}
   From this proposition, we have the following corollary. 
\begin{corollary}\label{cor1}
Determining whether or not a plan $p$ is a solution of 
the planning problem instance $\calp = (\cald, \cali, \calg)$ 
from an a-state $\delta$ can be done 
in polynomial time in the size of $p$ and $\calp$. 
\end{corollary}

\begin{definition}
The {\em conditional planning problem} is defined as follows.
\begin{itemize}
\item {\em Given:} A planning problem instance $\calp = (\cald,\cali,\calg)$ 
        of size $n$ and a polynomial $Q(n)\ge n$;
\item {\em Determine:} whether there exists a conditional 
      plan, whose size is bounded by $Q(n)$, 
      that achieves $\calg$ from $\cali$ (with respect to 
     Definition \ref{solutions}).
\end{itemize}
\end{definition}

\begin{theorem}
\label{t-pp}
The conditional planning problem is {\bf NP}-complete.
\end{theorem}
\begin{proof}
{\em see Appendix A}.
\end{proof}

The above theorem shows that planning using the 0-approximation
has lower complexity than planning with respect to the full 
semantics. Here, by the full semantics we mean the possible world 
semantics extended to domains with sensing actions. Yet, the price 
one has to pay is the incompleteness of this approximation, i.e., there are 
planning instances which have solutions 
with respect to the full semantics but do not have solutions with 
respect to the approximation. This can be seen in the following example.

\begin{example} \label{incomplete}
Consider the planning problem instance 
$\calp{=}(\cald,\cali,\calg)$ with 
$$\cald = \{\causes(a,f,\{g\}),\causes(a,f,\{\neg g\})\}, \;
\cali = \emptyset, \textnormal{ and } \calg = \{f\}.$$ 
We can easily check that $p=[a]$ is a plan achieves $f$
from every initial situation (with respect to the possible world 
semantics developed for $\cala_K$ in \cite{bar00}). 
However, $p$ is not a solution with respect to Definition \ref{solutions}, 
because $\cald \not\models_{\cali} \knows f \after a$.
\end{example}

The above example highlights the main weakness of this approximation 
in that it does not allow for reasoning by cases for non-sensing actions 
or in the presence of disjunctive initial situation. In our experiments 
with the benchmarks, we observe that most of the benchmarks that our planner
could not solve fall into the second category, i.e., they require the 
capability of reasoning with disjunctive information about the initial 
state. Given that we do not consider action theories with disjunctive 
initial state, this should not come as a surprise. 

\section{A Logic Programming Based Conditional Planner}
\label{planner}
\begin{figure}[bht]
\begin{center}
\epsfig{file=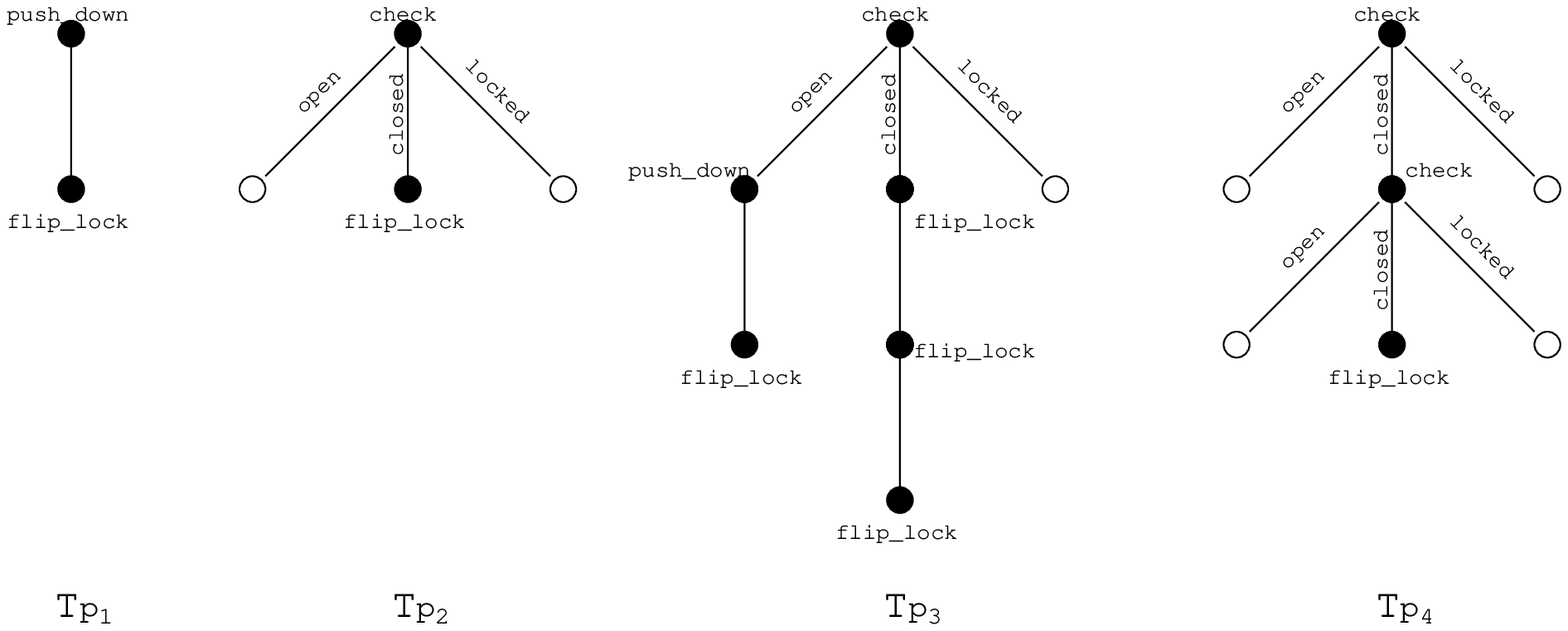, scale = 0.6}
\end{center}
\caption{
Sample plan trees
} 
\label{tree}
\end{figure}

This section describes an answer set programming based 
conditional planner, called \cps. Given a planning problem 
instance $\calp = (\cald,\cali,\calg)$, 
we translate it into a logic program $\pi_{h,w}(\calp)$,
where $h$ and $w$ are two input parameters whose meanings will become
clear shortly, and then use an answer set solver 
(e.g., \smodels \ or \cmodels) to compute its answer sets.
The answer sets of $\pi_{h,w}(\calp)$ represent 
solutions to $\calp$. Our intuition
behind this task rests on the observation that each plan $p$ 
(Definition \ref{condplan}) corresponds to a labeled plan 
tree $T_p$ defined as below.
\begin{list}{$\bullet$}{\topsep=1pt \parsep=0pt \itemsep=1pt}
\item If $p$ = [] then $T_p$ is a tree with a single node.
\item If $p = [a]$, where $a$ is a non-sensing action, then
$T_p$ is a tree with a single node and this node is labeled with $a$.
\item If $p = [a;q]$, where $a$ is a non-sensing action and $q$ is 
a non-empty plan, then $T_p$ is a tree whose root is labeled with 
$a$ and has only one subtree which is $T_q$. Furthermore, the 
link between $a$ and $T_q$'s root is labeled with an empty string.
\item If $p = [a;\kcases(\{g_j \rightarrow p_j\}_{j=1}^n)]$, where
$a$ is a sensing action that determines $g_j$'s, then $T_p$ is
a tree whose root is labeled with $a$ and has $n$ subtrees 
$\{T_{p_j} \mid j \in \{1,\dots,n\}\}$. 
For each $j$, the link from $a$ to the root 
of $T_{p_j}$ is labeled with $g_j$. 
\end{list}
Observe that each trajectory of the plan $p$ corresponds to 
a path from the root to a leave of $T_p$.
As an example, Figure \ref{tree} depicts the labeled trees 
for plans $p_1$, $p_2$, $p_3$ and $p_4$ in
Example \ref{ex07} (black nodes 
indicate that there exists an action occurring
at those nodes, while white nodes indicate
that there is no action occurring at those nodes).

For a plan $p$, let $\alpha$ be the number of leaves of 
$T_p$ and $\beta$ be the number of nodes along the longest path 
from the root to the leaves of $T_p$. $\alpha$ and $\beta$ 
will be called the {\em width} and {\em height} of $T_p$ respectively.
Suppose $w$ and $h$ are two integers that such that
$\alpha \le w$ and $\beta \le h$.

\begin{figure}[bht]
\begin{center}
\epsfig{file=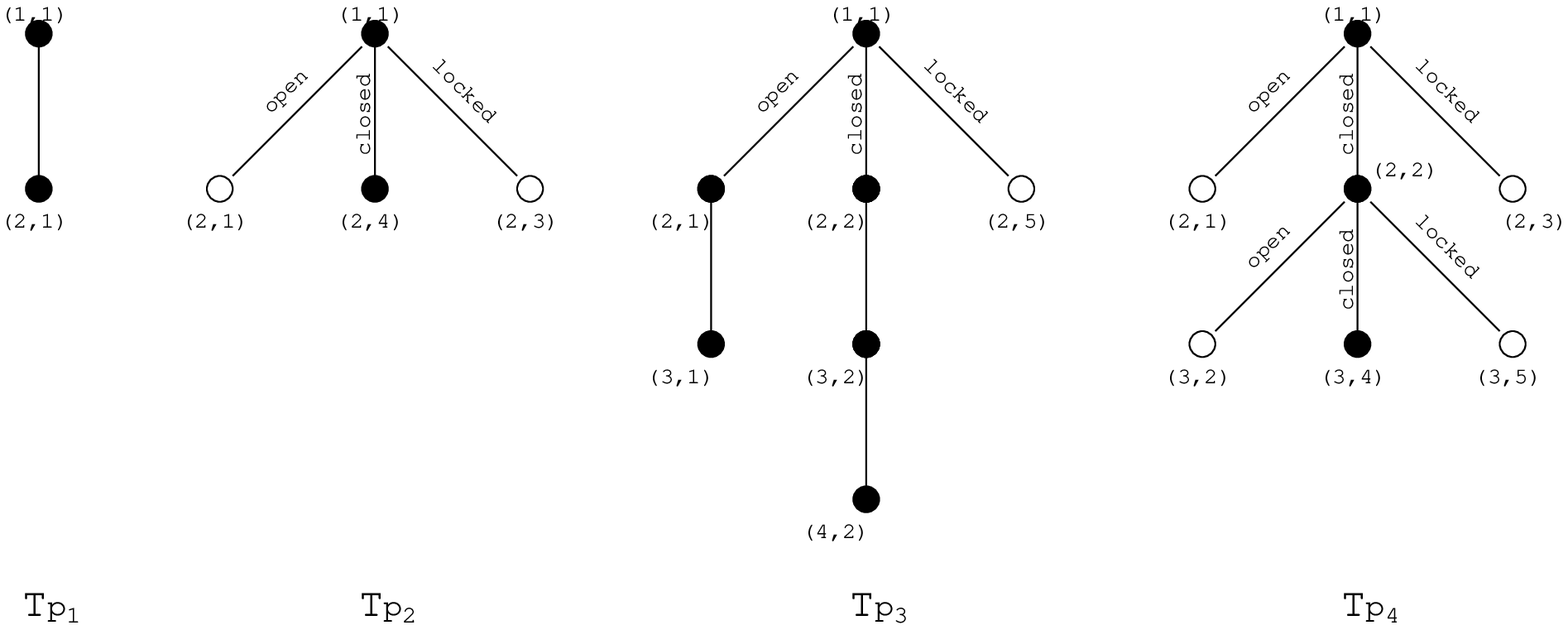, scale = 0.6}
\end{center}
\caption{Possible mappings for the trees in Figure \ref{tree}} 
\label{tree1}
\end{figure}

Let us denote the leaves of $T_p$ by $x_1,\ldots,x_{\alpha}$. 
We map each node $y$ of $T_p$ to a pair of integers $n_y$ = ($t_y$,$p_y$), 
where $t_y$ is the number of nodes along the path from
the root to $y$, and $p_y$ is defined in the following way.
\begin{itemize}
\item For each leaf $x_i$ of $T_p$, $p_{x_i}$ is
an arbitrary integer between $1$ and $w$. Furthermore,
there exists a leaf $x$ with $p$-value of $1$, i.e., $p_x = 1$, and
there exist no $i \ne j$ such that $p_{x_i} = p_{x_j}$.
\item For each interior node $y$ of $T_p$ with children 
$y_1,\ldots,y_r$, $p_y = \min\{p_{y_1},\ldots,p_{y_r}\}$.
\end{itemize}
For instance, Figure \ref{tree1} shows some possible mappings 
with $h=4$ and $w=5$ for the trees in Figure \ref{tree}. 
It is easy to see that if $\alpha \le w$ and $\beta \le h$ then
such a mapping always exists. Furthermore, from the construction 
of $T_p$, independently of how the leaves of $T_p$ are numbered, we 
have the following properties.
\begin{enumerate}
\item For every node $y$, $t_y \le h$ and $p_y \le w$.
\item For a node $y$, all of its children have the same $t$-value. 
That is, if $y$ has $r$ children $y_1,\ldots,y_r$ then $t_{y_i} = t_{y_j}$ 
for every $1 \le i,j \le r$. Furthermore, the $p$-value of $y$
is the smallest one among the $p$-values of its children.
\item The root of $T_p$ is always mapped to the pair $(1,1)$. 
\end{enumerate}

\begin{figure}[bht]
\begin{center}
\epsfig{file=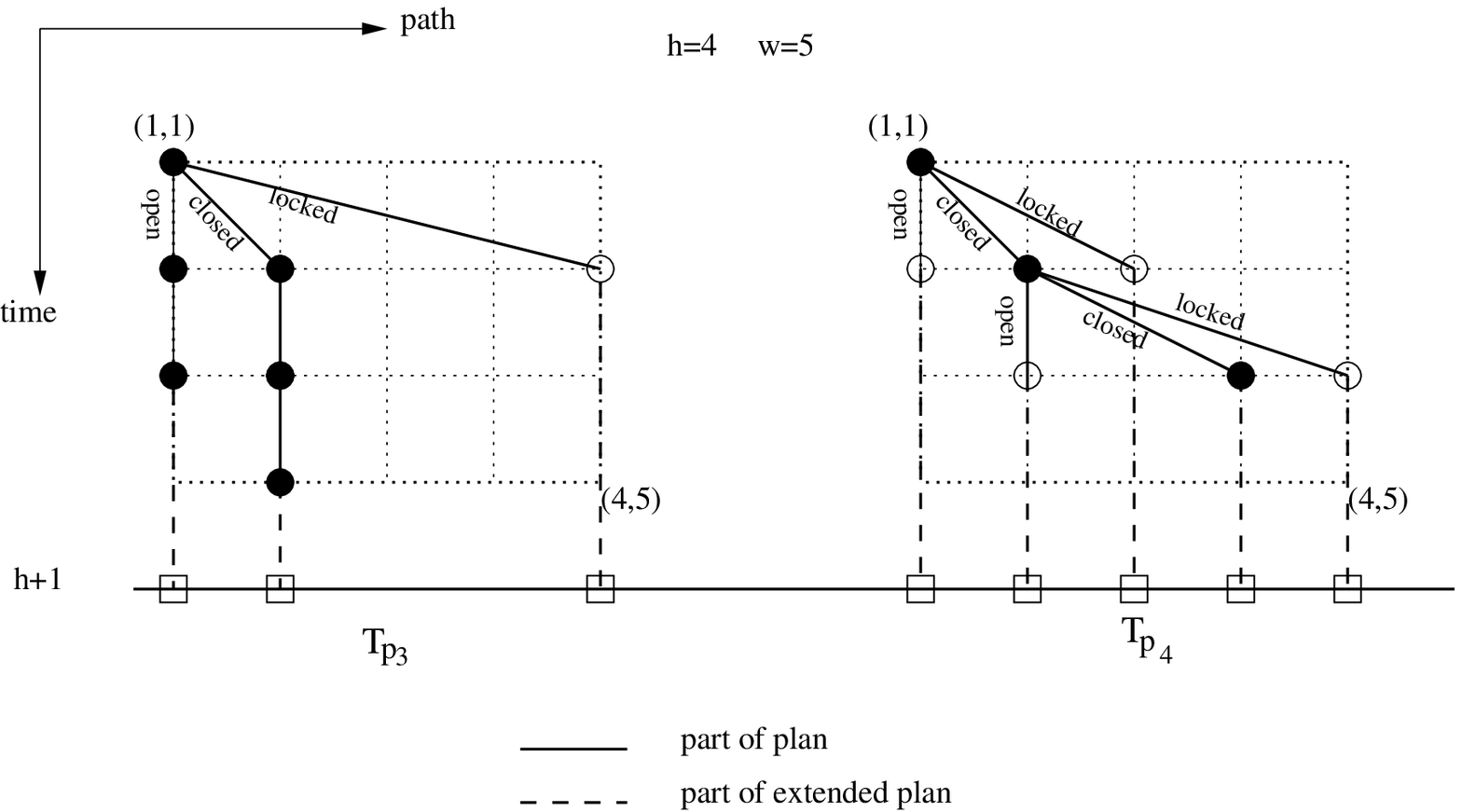, scale = 0.7}
\end{center}
\caption{Grid representation of conditional plans} 
\label{grid}
\end{figure}

Our encoding is based on the above mapping. We observe that a conditional plan
$p$ can be represented on a grid $h \times w$ 
where each node $y$ of $T_p$ is placed at the position 
$(t_y,p_y)$ relative to the leftmost top corner of the grid.
This way, it is guaranteed that the root of $T_p$ is always
placed at the leftmost top corner. Figure \ref{grid} depicts 
the $4 \times 5$ grid representation of conditional plans 
$T_{p_3}$ and $T_{p_4}$ in Figure \ref{tree1}. As it can be seen in 
Figure \ref{grid}, each path (trajectory) of the plan 
can end at an arbitrary time point. For example, the leftmost and 
rightmost trajectories of $T_{P_4}$ end at 2, whereas the others end at 3.
On the other hand, to check if the plan is indeed a solution, we need
to check the satisfaction of the goal at every leaf node of 
the plan, that is, at the end of each trajectory. In our encoding,
this task is simplified by extending all the trajectories of the plan 
so that they have the same height $h+1$ and then checking the 
goal at the end of each extended trajectory (see Figure \ref{grid}). 
Note that an a-state associated with each node on the extended part 
of each trajectory in our encoding will be guaranteed to be
the same as the one associated with the end node of the 
original trajectory.

We now describe the program $\pi_{h,w}(\calp)$ in the syntax 
of \smodels\ (for a concrete example, see Appendix C).
In $\pi_{h,w}(\calp)$, variables of sorts $time$ and
$path$ correspond to rows and columns of the grid. 
Instead of using the predicate $holds(L,T)$ ({\em see}, e.g., 
\cite{dim97,lif99d}) to denote that a literal $L$ 
holds at the time $T$, we use the predicate $holds(L,T,P)$ 
to represent the fact that $L$ holds at node $(T,P)$
(the time moment $T$, the path number $P$ on the grid).

The program $\pi_{h,w}(\calp)$ contains the following elements.
\begin{enumerate}
\item {\bf Constants.} There are two constants used in the program
$h$ and $w$ which serve as the input parameters of the program. In
addition, we have constants to denote fluents, literals and actions 
in the domain. Due to the fact that \smodels\ does not allow symbol 
$\neg$, to represent a literal constant $\neg f$, we will use $neg(f)$.
\item {\bf Predicates. } The program uses the following predicates.
\begin{itemize}
\item $time(T)$ is true if $1 \le T \le h$.
\item $time1(T_1)$ is true if $1 \le T_1 \le h+1$.
\item $path(P)$ is true if $1 \le P \le w$.
\item $fluent(F)$ is true if $F$ is a fluent.
\item $literal(L)$ is true if $L$ is a literal.
\item $contrary(L,L_1)$ is true if $L$ and $L_1$ are
two complementary literals.
\item $sense(L)$ is true if $L$ is a sensed literal.
\item $action(A)$ is true if $A$ is an action
\item $holds(L,T,P)$ is true if literal $L$ holds at $(T,P)$.
\item $poss(A,T,P)$ is true if action $A$ is executable at $(T,P)$.
\item $occ(A,T,P)$ is true if action $A$ occurs at $(T,P)$.
That means the node $(T,P)$ in $T_p$ is labeled with action $A$.
\item $e(L,T,P)$ is true if literal $L$ is an effect of a
non-sensing action occurring at $(T,P)$.
\item $pc(L,T,P)$ is true if literal $L$ may change at $(T+1,P)$. 
\item $goal(T,P)$ is true if the goal is satisfied at $(T,P)$.
\item $br(G,T,P,P_1)$ is true if there exists a branch from $(T,P)$
to $(T+1,P_1)$ labeled with $G$ in $T_p$. For example, in the grid 
representation of $T_{p_3}$ (Figure \ref{grid}), we have $br(open,1,1,1)$, 
$br(closed,1,1,2)$, and $br(locked,1,1,5)$.

\item $used(T,P)$ is true if $(T,P)$ belongs to some extended 
trajectory of the plan. This allows us to know
which paths are used in the construction of the plan and thus to be
able to check if the plan satisfies the goal.
As an example, for $T_{p_3}$ in Figure \ref{grid},
we have $used(t,1)$ for $1 \le t \le 5$, and
$used(t,2)$ and $used(t,5)$ for $2 \le t \le 5$. The goal 
satisfaction, hence, will be checked at nodes $used(5,1)$, $used(5,2)$,
and $used(5,5)$.

\end{itemize}
\item {\bf Variables.} 
The following variables are used in the program.
\begin{itemize}
\item $F$: a $fluent$ variable.
\item $L$ and $L_1$: $literal$ variables.
\item $T$ and $T_1$: $time$ variables,
in ranges $1..h$ and $1..h+1$ respectively, 
\item $G$, $G_1$ and $G_2$:  $sensed{-}literal$ variables.
\item $A$: an $action$ variable.
\item $P$, $P_1$, and $P_2$: $path$ variables,
in range $1..w$.
\end{itemize}
The domains of these variables are declared in \smodels\ using 
the keyword {\tt \#domain} (see Appendix C for more details).
Observe that the type of each variable has to be declared accordingly
if this feature of \smodels\ is not used. 

\item {\bf Rules. }
The program has the following facts to define variables of sort
$time$ and $path$:
\begin{eqnarray*}
time(1..h) & \la\\
time1(1..h+1) & \la \\
path(1..w) & \la
\end{eqnarray*}

For each action $a$, fluent $f$, or sensed-literal $g$ in the domain, 
$\pi_{h,w}(\calp)$ contains the following facts respectively
\begin{eqnarray*}
action(a) & \la & \\
fluent(f) & \la & \\
sense(g) & \la &
\end{eqnarray*}
The remaining rules of $\pi_{h,w}(\calp)$ are divided into three 
groups: (i) domain dependent rules; (ii) goal representation and 
(iii) domain independent rules, which are given next. Note that they are shown in a
shortened form in which the following shortening conventions 
are used.
\begin{itemize}
\item Two contrary literal variables are written as $L$ and $\neg L$.
\item For a predicate symbol $p$, and a set $\gamma$ of 
literals or actions, we will write $p(\gamma,\dots)$ to denote
the set of atoms $\{ p(x,\dots) \mid x \in \gamma \}$.
\item For a literal constant $l$, $\neg l$ stands for $neg(f)$
(resp. $f$) if $l = f$  (resp. $l = \neg f$) for some fluent $f$.
\end{itemize}
For example, the rule (\ref{r_14}) stands for the
following rule
$$holds(L,T{+}1,P) \la holds(L,T,P), contrary(L,L_1), \naf pc(L_1,T,P)$$
\end{enumerate}
\subsection{Domain dependent rules}
\begin{itemize}
\item {\bf Rules encoding the initial situation.} 
For each v-proposition (\ref{init}) in $\cali$, 
$\pi_{h,w}(\calp)$ contains the fact
\begin{eqnarray}
holds(l,1,1)& \la & \label{r_1}
\end{eqnarray}
\item 
{\bf Rules encoding actions' executability conditions.} 
For each executability condition (\ref{exec}) in $\cald$,
$\pi_{h,w}(\calp)$ contains the rule
\begin{eqnarray}
poss(a,T,P)& \la & holds(\psi,T,P) \label{r_2}
\end{eqnarray}

\item 
{\bf Rules for reasoning about the effect of non-sensing actions. }
For each dynamic causal law (\ref{dynamic}) in $D$,
we add to $\pi_{h,w}(\calp)$ the following rules: 
\begin{eqnarray}
e(l,T,P) & \la &
    occ(a,T,P), holds(\phi,T,P)\label{r_3}\\
pc(l,T,P)& \la &
     occ(a,T,P), \naf holds(l,T,P), 
     \naf holds(\neg \phi,T,P) \label{r_4} 
\end{eqnarray}
Here, $a$ is a non-sensing action. Its execution changes the 
world according to the
$Res$-function. The first rule, when used along with
(\ref{r_9}), encodes what definitely holds
as the effect of $a$ in the next a-state. 
The second rule, when used along with (\ref{r_8}),
describes what would potentially be changed by $a$
(see the definitions of $e(a,\delta)$ and $pc(a,\delta)$
in Subsection \ref{approx}). Note that in the
second rule, $\naf holds(\neg \phi,T,P)$ stands
for $\{ \naf holds(\neg l) \mid l \in \phi \}$,
meaning that $\phi$ possibly holds at $(T,P)$.
These rules will be used in cooperation with (\ref{r_10}),
(\ref{r_13}), and (\ref{r_14}) to define the next 
a-state after the execution of a non-sensing 
action.
\item 
{\bf Rules for reasoning about the effect of sensing actions. } 
For each k-proposition (\ref{knowledge}) in $\cald$, $\pi_{h,w}(\calp)$ 
contains the following rules:
\begin{eqnarray}
& \la & occ(a,T,P), \naf br(\theta,T,P,P) \label{r_5} \\ 
 1\{br(g,T,P,X){:}new\_br(P,X)\}1  & \la &  
    occ(a,T,P) \label{r_6} \\
& & \;\; (g \in \theta) \nonumber \\
& \la & occ(a,T,P), holds(g, T,P) \label{r_7} \\
& & \;\; (g \in \theta) \nonumber
\end{eqnarray}
The first rule assures that if a sensing action $a$ occurs 
at $(T,P)$ then there must be a branch from $(T,P)$ to
$(T+1,P)$. The second rule ensures that a new branch, corresponding
to a new successor a-state, will be created for each literal 
sensed by the action. The last rule is a constraint that 
prevents $a$ from taking place if one of the literals 
sensed by the action is already known. With this rule,
the returned plan is guaranteed to be optimal in the sense that a 
sensing action should not occur if one of the literals sensed by 
the action already holds. Observe that the semantics of ${\cala}_K$ 
does not prevent a sensing action to execute when some of 
its sensed-fluents is known. For this reason, some solutions 
to a planning problem instance might not be found using 
this encoding. However, as we will see later, 
the program will generate an ``equivalent'' plan
to those solutions. Subsection \ref{soundcomp} will elaborate 
more on this issue.

\item 
{\bf Rules for reasoning about static causal laws. }
For each static causal law (\ref{static}) in $\cald$,
$\pi_{h,w}(\calp)$ contains the rules
\begin{eqnarray}
pc(l,T,P) & \la & \naf holds(l,T,P), pc(l',T,P),\nonumber \\
  & & \naf e(\neg \varphi,T,P) \label{r_8}\\
  & & (l' \in \varphi) \nonumber\\
e(l,T,P)& \la & e(\varphi,T,P) \label{r_9}\\
holds(l,T_1,P)& \la &
    holds(\varphi,T_1,P) \label{r_10}
\end{eqnarray}
Rules in this group encode the equations (\ref{sem-pc1})-(\ref{sem-pc3})
and the operator $Cl_\cald$.
\end{itemize}

\subsection{Goal representation}
The following rules encode the goal and make sure that it is 
always achieved at the end of every possible branch 
created by the execution of the plan. 
\begin{eqnarray}
goal(T_1,P) & \la & holds(\calg,T_1,P)  \label{r_11}\\
goal(T_1,P) & \la & holds(L,T_1,P), holds(\neg L,T_1,P)  \label{r_11_2}\\
& \la & used(h{+}1,P), \naf goal(h{+}1,P) \label{r_12}
\end{eqnarray}
The first rule says that the goal is satisfied at a node
if all of its subgoals are satisfied at that node.
The last rule guarantees that if a path $P$ is used in the construction 
of a plan then the goal must be satisfied at the end of this path,
that is, at node $(h+1,P)$.

Rule (\ref{r_11_2}) deserves some explanation. Intuitively, 
the presence of $ holds(L,T,P)$ and $ holds(\neg L,T,P)$ indicates 
that the a-state at the node $(T,P)$ is inconsistent. This means that
no action should be generated at this node as inconsistent a-states 
will be removed by the extended transition function 
(Definition \ref{d:weak:extend}). To achieve this effect\footnote{
   The same effect can be achieved by (i) introducing a new predicate,
   say $stop(T,P)$, to represent that the a-state at $(T,P)$ is 
   inconsistent;  (ii) adding $\naf stop(T,P)$ in the body of 
   rule (\ref{r_22}) to prevent action to occur at $(T,P)$; 
   and (iii) modifying the rule (\ref{r_12}) accordingly.
}, 
we say that the ``goal'' has been achieved at $(T,P)$. 
The inclusion of this rule might raise the question:
is it possible for the program to generate a plan 
whose execution yields inconsistent a-states only. Fortunately, due to 
Proposition \ref{prop-cons-action-theory}, this will not be the 
case for consistent action theories. 

\subsection{Domain independent rules}
\begin{itemize}
\item 
{\bf Rules encoding the effect of non-sensing actions.}
Rules (\ref{r_3}) -- (\ref{r_4}) specify what definitely
holds and what could potentially be changed in the next 
a-state as the effect of a non-sensing action.
The following rules encode the effect and frame axioms 
for non-sensing actions. 
\begin{eqnarray}
holds(L,T{+}1,P)& {\la} & e(L,T,P) \label{r_13}\\
holds(L,T{+}1,P)& {\la} &
    holds(L,T,P), \naf pc(\myneg{L},T,P) \label{r_14}
\end{eqnarray}
When used in conjunction with (\ref{r_3}) -- (\ref{r_4}), 
they define the $Res$ function. 

\item 
{\bf Inertial rules for sensing actions.}
This group of rules encodes the fact that the execution of 
a sensing action does not change the world. However,
there is one-to-one correspondence between the set of sensed
literals and the set of possible next a-states after the 
execution of a sensing action. 
\begin{eqnarray}
& \la & P_1 < P_2, P_2 < P, br(G_1,T,P_1,P), \nonumber \\
  & & br(G_2,T,P_2,P) \label{r_16}  \\
& \la & P_1 \le P, G_1 \ne G_2, br(G_1,T,P_1,P), \nonumber \\
  & & br(G_2,T,P_1,P) \label{r_17}  \\
& \la & P_1 <  P, br(G,T,P_1,P), used(T,P) \label{r_18} \\
used(T{+}1,P) & \la &  P_1 < P, br(G,T,P_1,P)\label{r_19} \\
holds(G,T{+}1,P) & \la & P_1 \le P, br(G,T,P_1,P)   \label{r_20} \\
holds(L,T{+}1,P)& \la & P_1 < P, br(G,T,P_1,P), 
holds(L,T,P_1) \label{r_21}
\end{eqnarray}
The first three rules make sure that there is no cycle in
the plan that we are encoding. The next rule is to mark
a node as used if there exists a branch in the plan that
coming to that node. This allows us to know which paths on
the grid are used in the construction of the plan and thus
to be able to check if the plan satisfies the goal (see rule
(\ref{r_12})).

The last two rules, along with rule (\ref{r_10}),
encode the possible next a-state corresponding to the branch 
denoted by literal $G$ after a sensing action is performed in 
a state $\delta$. They say that such a-state should contain $G$ 
(rule (\ref{r_20})) and literals that hold in $\delta$ 
(rule (\ref{r_21})). 

Note that because for each literal $G$ sensed by a 
sensing action $a$, we create a corresponding branch 
(rules (\ref{r_5}) and (\ref{r_6})), the rules of this 
group guarantee that all possible next a-states 
after $a$ is performed are generated.

\item 
{\bf Rules for generating action occurrences.} 
\begin{eqnarray}
1\{occ(X,T,P) : action(X)\} 1 & \la &
    used(T,P), \naf goal(T,P)  \label{r_22}\\
& \la &  occ(A,T,P), \naf poss(A,T,P) \label{r_23}
\end{eqnarray}
The first rule enforces exactly one action to
take place at a node that was used but the goal has not 
been achieved. The second one guarantees that 
only executable actions can occur.

\item 
{\bf Auxiliary Rules. }
\begin{eqnarray}
literal(F)& \la & \label{r_24}\\
literal(\neg F)& \la &  \label{r_25}\\
contrary(F,\neg F) & \la & \label{r_26} \\
contrary(\neg F,F) & \la & \label{r_27} \\
new\_br(P,P_1) & \la &  P \le P_1  \label{r_28} \\
used(1,1) & \la &   \label{r_29} \\
used(T{+}1,P)& \la &  used(T,P) \label{r_30}
\end{eqnarray}
The first four rules define literals and contrary literals.
Rule (\ref{r_28}) says that a newly created branch should outgo
to a path number greater than the current path.
The last two rules mark nodes that have been used.
\end{itemize}

\section{Properties of \cps}
\label{property} 
This section discusses some important properties of \cps. 
We begin with how to extract a solution from an answer set 
returned by \cps. Then, we argue that \cps \ is sound 
and complete with respect to the 0-approximation semantics.
We also show that \cps\ can be used as a conformant planner.
Finally, we present how to modify \cps\ to act as a reasoner.
\subsection{Solution Extraction}
In some previous answer set based planners
\cite{dim97,eit03,lif99d}, reconstructing a plan from an 
answer set for a logic program encoding the planning problem instance 
is quite simple: we only need to collect the action occurrences in the model
and then order them by the time they occur. In other
words, if the answer set contains $occ(a_1,1)$, $\ldots$, $occ(a_m,m)$ then
the plan is $a_1,\ldots,a_m$. For $\pi_{h,w}(\calp)$, the reconstruction
process is not that simple because each answer set for
$\pi_{h,w}(\calp)$ represents a conditional plan which may contain
conditionals in the form $br(l,t,p,p_1)$. 
The following procedure describes how to extract such a plan from 
an answer set.

Let $\calp = (\cald,\cali,\calg)$ be a planning problem instance
and $S$ be an answer set for $\pi_{h,w}(\calp)$. For any pair of integers,
$1 \le i \le h+1, 1 \le k \le w$, we define $p^k_i(S)$ as follows:
\[
p^k_i(S) = \left \{
\begin{array}{ll}
[] & \textnormal{if } i = h + 1 \textnormal{ or } occ(a,i,k) \not \in S
   \textnormal { for all } a   \\
a;p^k_{i+1}(S) & \textnormal{if } occ(a,i,k) \in S \textnormal { and } \\
   & a  \textnormal{ is a non-sensing action}\\
a;cases(\{g_j \rightarrow p_{i+1}^{k_j}(S)\}^n_{j=1}) &
\textnormal{if } occ(a,i,k) \in S, \\
& a  \textnormal{ is a sensing action},  \textnormal {and }\\
& br(g_j,i,k,k_j) \in S 
\textnormal{ for } 1 \le j \le n
\end{array}
\right.
\]
Intuitively, $p^k_i(S)$ is the conditional plan whose corresponding 
tree is rooted at node $(i,k)$ on the grid $h \times w$. $p^1_1(S)$
is, therefore, a solution to $\calp$. This is stated in 
Theorem \ref{t1} in the next subsection.

\subsection{Soundness and Completeness}
\label{soundcomp}
\begin{theorem}
\label{t1} Let $(\cald,\cali)$ be a consistent action 
theory, $\calp = (\cald,\cali,\calg)$ be a planning problem instance and
$h \ge 1$ and $w \ge 1$ be integers. If $\pi_{h,w}(\calp)$
returns an answer set $S$ then $p^1_1(S)$ is a solution to $\calp$.
\end{theorem}
\begin{proof}
{\em see Appendix B}.
\end{proof}
Theorem \ref{t1} shows the soundness of $\pi_{h,w}(\calp)$. 
We will now turn our attention to the completeness of 
$\pi_{h,w}(\calp)$. Observe that solutions generated by 
$\pi_{h,w}(\calp)$ are optimal in the following sense 
\begin{enumerate}
\item actions do not occur once the goal is 
achieved or a possible next a-state does not exist; and 
\item sensing actions do not occur if one 
of its sensed literals holds.
\end{enumerate}
The first property holds because of rule (\ref{r_22})
and the second property holds because of constraint
(\ref{r_7}). Since the definition of a conditional plan in general
does not rule out {\em non-optimal} plans, obviously $\pi_{h,w}(\calp)$ 
will not generate all possible solutions to $\calp$. 

For example, consider the planning problem instance $\calp_1$ in Example
\ref{ex04}. We have seen that plans $p_2$, $p_3$, and $p_4$ 
in Example \ref{ex07} are all solutions to $\calp_1$. 
However, $p_3$ and $p_4$ are not optimal because they do not 
satisfy the above two properties.

The above example shows that $\pi_{h,w}(\calp)$  is not complete 
w.r.t. the 0-approximation in the sense that no one-to-one correspondence 
between its answer sets and solutions to $\calp$ exists. 
However, we will show next that it is complete 
in the sense that for each solution $p$ to $\calp$, there exist
two integers $h$ and $w$ such that $\pi_{h,w}(\calp)$ will 
generate an answer set $S$ whose corresponding plan,
$p^1_1(S)$, can be obtained from $p$ by applying the
following transformation (called the $\reduct$ operation).

\begin{definition}
[Reduct of a plan]
\label{def_reduct}
Let $\calp = (\cald,\cali,\calg)$ be a planning problem instance,
$p$ be a plan and $\delta$ be an a-state such that 
$\hat\Phi(p, \delta) \ne \bot$.  A reduct of $p$ with respect 
to $\delta$, denoted by $\reduct_{\delta}(p)$, is defined as follows.
\begin{itemize}
\item[1.] if $p = []$ or $\delta \models \calg$ then
$$\reduct_{\delta}(p) = []$$
\item[2.] if $p = [a;q]$, where $a$ is a non-sensing action and
$q$ is a plan, then 
$$
\reduct_\delta(p) = 
\left \{
\begin{array}{ll}
a;\reduct_{\delta'}(q) & \textnormal{ if } \Phi(a, \delta) = \{\delta'\}\\
a & \textnormal{ otherwise }
\end{array}
\right.
$$
\item[3.] if $p = [a;\kcases(\{g_j \rightarrow p_j\}_{j=1}^n)]$,
where $a$ is a sensing action that senses $g_1,\ldots,g_n$, then 
$$\reduct_\delta(p) = 
\left \{
  \begin{array}{ll}
    \reduct_\delta(p_k) & \textnormal{ if } g_k \textnormal{ holds in }
    \delta \textnormal{ for some } k \\
    a; \kcases(\{g_j \rightarrow q_j\}_{j=1}^n) & \textnormal{ otherwise}
  \end{array}
\right.
$$
where
$$q_j = 
\left \{
  \begin{array}{ll}
    [] & \textnormal{ if } Cl_{\cald}(\delta \cup \{g_j\})
    \textnormal{ is inconsistent} \\
    \reduct_{Cl_{\cald}(\delta \cup \{g_j\})} (p_j) 
    & \textnormal{ otherwise}
  \end{array}
\right.
$$
\end{itemize}
\end{definition}
\begin{example}
\label{ex22}
Consider the planning problem instance $\calp_1$ in Example
\ref{ex04} and plans $p_2$, $p_3$, and $p_4$
in Example \ref{ex07}. 
Let $\delta = \{ \neg open \}$. We will show that
\begin{equation}
\reduct_{\delta}(p_3) = p_2 \label{eq_reduct_1}
\end{equation}
and
\begin{equation}
\reduct_{\delta}(p_4) = p_2 \label{eq_reduct_2}
\end{equation}
Because $open$, $closed$, and $locked$ do not hold in
$\delta$, we have
$$\reduct_{\delta}(p_3)=check;\kcases(\{open \rightarrow q_1,
closed \rightarrow q_2, locked \rightarrow q_3\})$$
where $q_j$'s are defined as in Definition \ref{def_reduct}.

Let 
\begin{eqnarray*}
\delta_1 & = & Cl_{\cald_1}(\delta \cup \{open\})
= \{ open, \neg open, closed, \neg closed, locked, \neg locked \} \\
\delta_2 & = & Cl_{\cald_1}(\delta \cup \{closed\}) = \{
\neg open, closed, \neg locked\} \\
\delta_3 & = &  Cl_{\cald_1}(\delta \cup \{locked\}) = \{
\neg open, \neg closed, locked\}
\end{eqnarray*}
It is easy to see that $q_1 = []$ (because $\delta_1$ is 
inconsistent) and $q_3 = []$ (because the sub-plan corresponding
to the branch ``locked'' in $p_3$ is empty). 

Let us compute $q_2$. We have
\begin{eqnarray*}
q_2 & = & \reduct_{\delta_2}(flip\_lock;
flip\_lock;flip\_lock)
\end{eqnarray*}
Because $\delta_2$ does not satisfy $\calg$ and
$\Phi(flip\_lock,\delta_2) = \{\delta_{2,1}\} \ne \emptyset$,
where $$\delta_{2,1} = \{\neg open, \neg closed, locked\},$$
we have
\begin{eqnarray*}
q_2 & = & flip\_lock;\reduct_{\delta_{2,1}}(flip\_lock;flip\_lock)
\end{eqnarray*}
As $\delta_{2,1}$ satisfies $\calg$, we have
$\reduct_{\delta_{2,1}}(flip\_lock;flip\_lock) = []$. Hence,
$q_2 = flip\_lock$. Accordingly, we have
$$\reduct_{\delta}(p_3)=check;\kcases(\{open \rightarrow [],
closed \rightarrow [flip\_lock], locked \rightarrow []\})
= p_2$$
That is, (\ref{eq_reduct_1}) holds.

We now show that (\ref{eq_reduct_2}) holds. It is easy to
see that
$$\reduct_{\delta}(p_4) = check;\kcases(\{open \rightarrow [],
closed \rightarrow \reduct_{\delta_{2}}(p_2), 
locked \rightarrow []\})$$
Because $closed$ holds in $\delta_2$, we have
$$\reduct_{\delta_{2}}(p_2)= \reduct_{\delta_{2}}(flip\_lock)
= flip\_lock$$
Thus,
$$\reduct_{\delta}(p_4) = check;\kcases(\{open \rightarrow [],
closed \rightarrow flip\_lock, 
locked \rightarrow []\}) = p_2$$
As a result, we have (\ref{eq_reduct_2}) holds.
\end{example}

We have the following proposition.

\begin{proposition} 
\label{prop-reduct}
Let $\calp = (\cald,\cali,\calg)$ be a planning problem instance and 
$\delta$ be its initial a-state. Then, for every solution $p$
to $\calp$, $\reduct_\delta(p)$ is unique and also 
a solution to $\calp$.
\end{proposition}
\begin{proof}
{\em see Appendix B}.
\end{proof}

The following theorem shows the completeness of our planner with respect
to the 0-approximation semantics.

\begin{theorem}
\label{t2} 
Let $\calp = (\cald,\cali,\calg)$ be a planning problem instance, and $p$ be 
a solution to $\calp$. Then, there exist two integers $h$ and 
$w$ such that $\pi_{h,w}(\calp)$ 
has an answer set $S$ and $p^1_1(S) = \reduct_{\delta}(p)$, 
where $\delta$ is the initial a-state of $(\cald,\cali)$. 
\end{theorem}
\begin{proof}
{\em see Appendix B}.
\end{proof}

\subsection{Special Case: \cps \ as a Conformant Planner}
Since conformant planning deals only with incomplete information, 
it is easy to see that $\pi_{h,1}(\calp)$ can be used to generate 
conformant plans for $\calp$. 

Let $S$ be an answer set for $\pi_{h,1}(\calp)$. 
Recall that we assume that each sensing action senses 
at least two literals. Hence, $w = 1$ implies $S$ does not
contain $occ(a,\dots)$ where $a$ is a sensing action because
if otherwise rules (\ref{r_6}) and (\ref{r_17}) cannot be satisfied. 
Thus, $p^1_1(S)$ is a sequence of non-sensing actions. By
Theorem \ref{t1}, we know that $p^1_1(S)$ achieves the goal of $\calp$ 
from every possible initial a-state of the domain, which implies that 
$p_1^1(S)$ is a conformant plan. In Section \ref{experiments},
we compare the performance of $\pi_{h,1}(\calp)$  against some of 
the state-of-the-art conformant planners. 

\subsection{Special Case: \cps\ as a Reasoner}
It is easy to see that with minor changes, \cps\ can be used to compute
the consequences of a plan. This can be done as follows. 
Given an action theory $(\cald,\cali)$,
for any integers $h,w$, let $\pi_{h,w}(\cald,\cali)$ be the set of rules: 
$\pi_{h,w}(\calp) \setminus \{ (\ref{r_5})-(\ref{r_7}), 
(\ref{r_11})-(\ref{r_12}), (\ref{r_16})-(\ref{r_18}),
(\ref{r_22}), (\ref{r_28}) \}$.
Intuitively, $\pi_{h,w}(\cald,\cali)$ is the program 
obtained from $\pi_{h,w}(\calp)$ by removing the rules for  
{\em (i)} generating
the branches when sensing actions are executed; {\em (ii)} checking the 
satisfaction of the goal; {\em (iii)} representing the constraints on 
branches; and {\em (iv)} generating action occurrences.
For a plan $p$, let $T_p$ be the corresponding tree for $p$ that
is numbered according to the principles described in 
the previous section. We define $\epsilon(p)$ to be the 
following set of atoms
\begin{eqnarray*}
& \{occ(a,t,p) \mid \exists \textnormal{ a node } x 
\textnormal{ in } T_p \textnormal{ labeled with action } a 
\textnormal{ and numbered with } (t,p)\} \cup\\
& \{br(g,t,p,p') \mid \exists \textnormal{a link } 
\textnormal{ labeled with } g \textnormal{ that connects
the node numbered with } (t,p) \\
& \textnormal{ to the node numbered with } (t+1,p') \textnormal{ in } T_p\}.
\end{eqnarray*}
It is easy to see that the program 
$\pi_{h,w}(\cald,\cali) \cup \epsilon(p)$ 
has a unique answer set which corresponds to $\hat\Phi(p, s_0)$.
This is detailed in the following proposition.
\begin{proposition}
Let $(\cald,\cali)$ be an action theory, $p$ be a plan, 
$\rho$ be a fluent formula,
$T_p$ be the plan tree for $p$ with a given numbering, 
and $h$ and $w$ be the height and
width of $T_p$ respectively. Let 
\[
\Pi = \pi_{h,w}(\cald,\cali) \cup \epsilon(p).
\]
We have that 
\begin{itemize}
\item $\Pi$ has a unique answer set $S$;

\item $\cald \models_{\cali} \knows \rho \after p \;$ if and only if

\begin{itemize}
\item there exists some $j$, $1 \le j \le w$, $\delta_{h+1,j}(S) \ne \bot$; 
and 
\item for every $j$, $1 \le j \le w$ and $\delta_{h+1,j}(S) \ne \bot$, 
$\rho$ is known to be true in $\delta_{h+1,j}(S)$. 
\end{itemize}

\item $\cald \models_{\cali} \kwhether \rho \after p \;$ if and only if

\begin{itemize}
\item there exists some $j$, $1 \le j \le w$, $\delta_{h+1,j}(S) \ne \bot$; 
and 
\item for every $j$, $1 \le j \le w$ and $\delta_{h+1,j}(S) \ne \bot$, 
$\rho$ is known in $\delta_{h+1,j}(S)$. 
\end{itemize}

\end{itemize}
where
$$
\delta_{t,j}(S) = \left \{
\begin{array}{ll}
\{ l \mid holds(l,t,j) \in S \} & \textnormal{if } used(t,j) \in S
\textnormal{ and } \\
& \{ l \mid holds(l,t,j) \in S \} \textnormal{ is consistent}\\
\bot & \textnormal{otherwise}
\end{array}
\right.
$$
\end{proposition}

\begin{proof}
The proof of this theorem is very similar to the 
proof of Theorem \ref{t1} so we omit it for brevity.
\end{proof}

\section{Evaluation}
\label{experiments}

In this section, we evaluate \cps\ against other planners 
using planning benchmarks from the literature. We first
briefly summarize the features of the systems that are used
in our experiments. We then describe the benchmarks.
Finally, we present the experimental results. 

\subsection{Planning Systems}
The planning systems that we compared with are
the following.
\begin{itemize}
\item \dlvk: \dlvk\ is a declarative, logic-based planning 
system built on top of the ${\tt DLV}$ system 
(\http{http://www.dbai.tuwien.ac.at/proj/dlv/}).
The input language ${\cal K}$ is a logic-based planning language
described in \cite{eit03}. The version we used for testing is
available at \http{http://www.dbai.tuwien.ac.at/proj/dlv/K/}.
\dlvk\ is capable of generating both concurrent 
and conformant plans. It, however, does not support sensing actions
and cannot generate conditional plans.

\item \cmbp \ (Conformant Model Based Planner) 
\cite{cim99,cimatti00}: \cmbp\ 
is a conformant planner developed by Cimatti and Roveri. A planning
domain in CMBP is represented as a finite state automaton. BDD 
(Binary Decision Diagram) techniques are employed to represent and
search the automaton. \cmbp \ allows nondeterministic domains with 
uncertainty in both the initial state and action effects. 
Nevertheless, it does not have the capability of generating 
concurrent and conditional plans. 
The input language to \cmbp \ is \ar \ described in \cite{lif97:aij}.
The version used for testing was downloaded from
\http{http://www.cs.washington.edu/research/jair/contents/v13.html}.

\item KACMBP \cite{CimattiRB04}: Similarly to CMBP, KACMBP uses
techniques from symbolic model checking to search in the
belief space. However, in KACMBP, the search is guided by a 
heuristic function which is derived based on knowledge associated 
with a belief state. 

KACMBP is designated for sequential and conformant setting. It, 
however, does not support concurrent planning and conditional
planning. The input language of KACMBP is SMV. 
The system was downloaded from 
\url{http://sra.itc.it/tools/mbp/AIJ04/}.

\item Conformant-FF (CFF) \cite{brafman:hoffmann:icaps-04}: CFF\footnote{
  We would like to thank J{\"o}rg Hoffmann for providing us with 
  an executable version of the system for testing.
}, 
to our best knowledge, is one of the current fastest conformant 
planners in most of the benchmark
domains in the literature. It extends the classical FF 
planner \cite{HoffmanN01} to deal with uncertainty in the initial state.
The basic idea is to represent a belief state $s$ just by
the initial belief state (which is described as 
a CNF formula) together with the action sequence that 
leads to $s$. In addition, the reasoning 
is done by checking the satisfiability of CNF formulae.

The input language of CFF is a subset of PDDL with
a minor change that allows the users to specify the 
initial state as a CNF formula. Both 
sequential and and conformant planning are supported in CFF. However, 
it does not support concurrent and conditional planning.

\item MBP \cite{bertol01heuristic}: MBP is a previous version 
of CMBP. Unlike CMBP which only deals with conformant planning, 
MBP supports conditional planning as well. The version used for testing
was downloaded from \url{http://sra.itc.it/tools/mbp/}.

\item \sgp \ (Sensory Graph Plan) \cite{weld98b,anders98conditional}:
\sgp \ is a planner based on the planning graph algorithm proposed by
Blum and Furst in \cite{bf95}. \sgp \ supports conditional effects,
universal and existential quantification. It also handles 
uncertainty and sensing actions. \sgp \ has the capability
of generating both conformant and conditional plans, as well as
concurrent plans. Nevertheless, static laws are not allowed in 
\sgp. The input syntax is \pddl \ (Planning Domain Definition Language). 
The version used for testing
is 1.0h (dated January 14th, 2000), written in Lisp, available
at \http{http://www.cs.washington.edu/ai/sgp.html}. 

\item POND \cite{bryceKS04}:  POND extends the planning graph 
algorithm \cite{bf95} to deal with sensing actions. Conformant planning 
is also supported as a feature of POND. The input language is a 
subset of PDDL. POND was downloaded from 
\url{http://rakaposhi.eas.asu.edu/belief-search/}.

\end{itemize}
Table~\ref{tab:features} summarizes the features of these planning
systems.

\begin{small}
\begin{table}[h]
\centering
\begin{tabular}{|l|llllllll|}
\cline{1-9}
                       & {\footnotesize \cps} & {\footnotesize \dlvk} & {\footnotesize MBP}  & 
{\footnotesize \cmbp} & {\footnotesize \sgp} & {\footnotesize \pond} & 
{\footnotesize CFF} & {\footnotesize KACMBP} \\
\cline{1-9}
Input Language         & \ack & $K$    & \ar  & \ar & \pddl & \pddl & \pddl & SMV\\
Sequential planning    & yes  & yes    & yes  & yes & no & yes  & yes & yes\\
Concurrent planning    & no   & yes    & no   & no  & yes & no & no & no\\
Conformant planning    & yes  & yes    & yes  & yes & yes & yes & yes & yes\\
Conditional planning   & yes  & no     & yes  & no  & yes & yes & no & no\\
\cline{1-9}
\end{tabular}
\caption{Features of Planning Systems\label{tab:features}}
\end{table}
\end{small}
\subsection{Benchmarks}
To test the performance of the planners, we prepared two test suites
for conformant and conditional planning, separately.
In our preparation, we attempt to encode the planning problem 
instances given to the systems in a uniform way (in terms of the number of 
actions, fluents, and effects of actions). Due to the 
differences in the representation languages of these systems,
there are situations in which the encoding of the problems 
might be different for each system. 


\subsubsection{Conformant Planning}

We tested the systems on the following domains\footnote{
  The system is available at \url{http://www.cs.nmsu.edu/~tson/ASPlan/Sensing}.
}:

\begin{itemize}

\item 
{\bf Bomb in the Toilet (BT):}
This set of problems was introduced in \cite{dermot87critique}: ``It has been 
alarmed that there is a bomb in a lavatory. There are $m$ suspicious
packages, one of which contains the bomb. The bomb can be defused 
if we dunk the package that contains the bomb into a toilet.'' 
Experiments were made with $m=2$, $4$, $6$, $8$, and $10$.

\item {\bf Bomb in the Toilet with Multiple Toilets (BMT):}
This set of problems is similar to the {\bf BT} problem but 
we have multiple toilets. There are five problems in this set,
namely $BMT(2,2)$, $BMT(4,2)$, $BMT(6,2)$, $BMT(8,4)$, 
and $BMT(10,4)$, where the first parameter is the number of
suspicious packages and the second parameter is the number of
toilets.

\item {\bf Bomb in the Toilet with Clogging (BTC):}
This set of problems is similar to BTs but we assume that dunking 
a package clogs the toilet and flushing the toilet unclogs it. 
We know that in the beginning, the toilet is unclogged. We
did experiments with $m=2, 4, 6, 8, $ and $10$, where $m$ is
the number of suspicious packages.

\item {\bf Bomb in the Toilet with Multiple Toilets and Clogging (BMTC):}
This set of problems is similar to BTC but we have multiple
toilets. We did experiments with five problems $BMTC(2,2)$,
$BMTC(4,2)$, $BMTC(6,2)$, $BMTC(8,4)$, 
and \\$BMTC(10,4)$, where the first parameter is the number of
suspicious packages and the second parameter is the number of
toilets.

\item 
{\bf Bomb in the Toilet with Clogging and Uncertainty in Clogging (BTUC):}
This set of problems is similar to BTC except that we do not know
whether the toilet is clogged or not in the beginning.

\item {\bf Bomb in the Toilet with Multiple Toilets and 
Uncertainty in Clogging (BMTUC):}
This set of problems is similar to BMTC except that we do not know
whether or not each toilet is clogged in the beginning.

\item {\bf Ring:} This set of problems is from \cite{CimattiRB04}. 
In this domain, one can move in a cyclic fashion (either forward or backward) 
around a $n$-room building to lock windows. 
Each room has a window and the window can be locked 
only if it is closed. Initially, the robot is in the first room
and it does not know the state (open, closed or locked) of the windows.
The goal is to have all windows locked. A possible
conformant plan is to perform a sequence of
actions {\em forward, close, lock} repeatedly. 
In this domain, we tested with $n=$2,4,6,8, and 10.

\item {\bf Domino (DOM): } This domain is very
simple. We have $n$ dominos standing on a line in such a way that 
if one of them falls then the domino on its right also falls.
There is a ball hanging close to the leftmost one. Touching
the ball causes the first domino to fall. Initially,
the states of dominos are unknown. The goal is to have the 
rightmost one to fall. The solution is obviously to touch
the ball. In this domain, we tested with $n=$10,20,50,100,
1000, and 10000.
\end{itemize}

\subsubsection{Conditional Planning}

The set of problems for testing includes:

\begin{itemize}


\item {\bf Bomb in the Toilet with Sensing Actions (BTS):}
This set of examples is taken from \cite{weld98b}. They are variations
of the BTC problem that allow sensing actions to be used to 
determine the existence of a bomb in a specific package. There are
$m$ packages and only one toilet. We can use one of the following methods 
to detect a bomb in a package:
(1) use a metal detector (action $detect\_metal$); 
(2) use a trained dog to sniff the bomb (action $sniff$); 
(3) use an x-ray machine (action $xray$); and, finally,
(4) listen for the ticking of the bomb (action $listen\_for\_ticking$).

This set of examples contains four subsets of problems, namely 
$BTS1(m)$,\\$BTS2(m)$, $BTS3(m)$, and 
$BTS4(m)$ respectively, where $m$ is the number of suspicious
packages. These subsets differ from each other in 
which ones of the above methods are allowed to use.
The first subset allows only one sensing action (1);
the second one allows sensing actions (1)-(2); and so on.

\item
{\bf Medical Problem (MED):}
This set of problems is from \cite{weld98b}. A patient is sick
and we want to find the right medication for her. Using
a wrong medication may be fatal. Performing a throat culture will
return either $red$, $blue$, or $white$, which determines the group
of illness the patient is infected with. Inspecting the color (that can
be performed only after the throat culture is done) allows us
to observe the color returned by a throat culture, depending on
the illness of the patient. Analyzing a blood sample tells us whether
or not the patient has a high white cell count. This can be done
only after a blood sample is taken. In addition, we know that 
in the beginning, the patient is not dead but infected. In addition,
none of the tests have been done.

There are five problems in this set, namely, $MED1$, $\dots$,
$MED5$. These problems are different from each other in how much
we know about the illness of the patient in the beginning. 

\item {\bf Sick Domain (SICK):}
This set of problems is similar to MED. A patient is sick and we 
need to find a proper medication for her. There are $n$ kinds
of illness that she may be infected with and each requiring a particular
medication. Performing throat culture can return a particular
color. Inspecting that color determine what kind of illness the 
patient has. Initially, we do not know the exact illness that
the patient is infected with.

The characteristic of this domain is that the length of the plan
is fixed (only 3) but the width of the plan may be large, depending
on the number of illnesses. We did experiments with five problems
in the domain, namely, $SICK(2)$, $SICK(4)$, ..., $SICK(10)$.
They differ from each other in the number of illnesses that the 
patient may have.

\item {\bf Ring (RINGS)}: This domain is a modification
of the $RING$ domain. In this modified version, the agent can close
a window only if it is open. It can lock a window only if
it is closed. The agent can determine the status of a window
by observing it (sensing action $observe\_window$).

\item {\bf Domino (DOMS)}:
This is a variant of the $DOM$ domain in which some dominos
may be glued to the table. Unlike the original version
of the $DOM$ domain, in this variant, 
when a domino falls, the next one falls only if
it is not glued. The agent can do an action to unglue
a glued domino. We introduce a new sensing action
$observe\_domino(X)$ to determine whether a
domino $X$ is glued or not.
\end{itemize}

\subsection{Performance}
We ran our experiments on a 2.4 GHz CPU, 768MB RAM, DELL machine, 
running Slackware 10.0 operating system. We compared \cps\ with 
\dlvk, \cmbp, \sgp, CFF and KACMBP on the conformant benchmarks 
and with \sgp, POND, and MBP on the conditional benchmarks.
Time limit was set to 30 minutes. 
The CMU Common Lisp version 19a was used to run \sgp\ examples.
We ran \cps\ examples on both \cmodels\ and \smodels. 
By convention, in what follows, we will use \cps$^c$ and \cps$^s$ to 
refer to the planner \cps\ when it was run on \cmodels\ and 
\smodels\ respectively. Sometimes, if the distinction between 
the two is not important, by \cps\ we mean both.

The experimental results for conformant and conditional planning
are shown in Tables \ref{tab:conf} and
\ref{tab:cond} respectively. Times are in seconds.
``TO/AB'' indicates that the corresponding planner does not 
return a solution within the time limit or stopped abnormally due
to some reasons, for example, out of memory or segmentation fault.

\begin{table}[h]
\begin{small}
\begin{tabular}[c]{|l|c|r|r|r|r|r|r|r|}
\cline{1-9}
Problem & Min. & \multicolumn{2}{c|}{\cps} & \dlvk & \cmbp & \sgp & CFF & 
KA- \ \\
\cline{3-4} & PL & cmodels & smodels & & & & & CMBP\\
\cline{1-9}
{\em BT(2)}       & 2  & 0.06 & 0.03 & 0.01  & 0.03 & 0.04 & 0.02 & 0.12\\
{\em BT(4)}       & 4  & 0.04 & 0.06 & 0.03  & 0.03 & 0.27 & 0.04 & 0.12\\
{\em BT(6)}       & 6  & 0.05 & 0.12 & 0.18  & 0.04 & 0.42 & 0.09 & 0.1\\
{\em BT(8)}       & 8  & 0.10 & 0.33 & 1.47  & 0.10 & 1.04 & 0.10 & 0.11\\
{\em BT(10)}      & 10 & 0.12 & 2.54 & 11.37 & 0.50 & 2.13 & 0.13 & 0.11\\
\cline{1-9}
{\em BMT(2,2)}    & 2  & 0.04 & 0.04 & 0.01  & 0.03 & 0.07 & 0.02 & 0.07\\
{\em BMT(4,2)}    & 4  & 0.05 & 0.09 & 0.03  & 0.04 & 0.28 & 0.03 & 0.12\\
{\em BMT(6,2)}    & 6  & 0.11 & 0.23 & 0.19  & 0.05 & 0.29 & 0.07 & 0.10\\
{\em BMT(8,4)}    & 8  & 0.41 & 4.70 & 1.70  & 0.11 & 3.14 & 0.09 & 0.11\\
{\em BMT(10,4)}   & 10 & 0.51 &152.45& 12.18 & 0.53 & 5.90 & 0.12 & 0.14\\
\cline{1-9}
{\em BTC(2)}      & 2  & 0.04 & 0.04 & 0.01  & 0.03 & 0.44 & 0.05 & 0.12\\
{\em BTC(4)}      & 7  & 0.04 & 0.12 & 0.33  & 0.04 & 21.62& 0.06 & 0.10\\
{\em BTC(6)}      & 11 & 0.06 & 0.33 & TO & 0.1 & TO       & 0.07 & 0.11\\
{\em BTC(8)}      & 15 & 0.11 & 0.53 & TO & 0.79 & TO      & 0.07 & 0.13\\
{\em BTC(10)}     & 19 & 0.12 & 468.04 & TO & 9.76 & TO    & 0.13 & 0.14\\
\cline{1-9}
{\em BMTC(2,2)}   & 2  & 0.06 & 0.06 & 0.01 & 0.03 & 0.18  & 0.05 & 0.12\\
{\em BMTC(4,2)}   & 6  & 0.10 & 0.19 & 0.17 & 0.05 & 2.03  & 0.04 & 0.09\\
{\em BMTC(6,2)}   & 10 & 0.14 & 0.63 & 20.02 & 0.24 & TO   & 0.07 & 0.12\\
{\em BMTC(8,4)}   & 12 & 0.56 & 60.56 & TO & TO & TO       & 0.10 & 0.12\\
{\em BMTC(10,4)}  & 16 & 1.44 & TO & TO & TO & TO          & 0.13 & 0.17\\
\cline{1-9}
{\em BTUC(2)}     & 4  & 0.05 & 0.04 & 0.02 & 0.02 & 0.59  & 0.03 & 0.09\\
{\em BTUC(4)}     & 8  & 0.04 & 0.11 & 0.94 & 0.04 & TO    & 0.04 & 0.11\\
{\em BTUC(6)}     & 12 & 0.06 & 0.22 & 524.3 & 0.11 & TO   & 0.06 & 0.11\\
{\em BTUC(8)}     & 16 & 0.11 & 4.7 & TO & 0.96 & TO       & 0.08 & 0.12\\
{\em BTUC(10)}    & 20 & 0.12 & TO & TO & 11.58 & TO       & 0.13 & 0.16\\
\cline{1-9}
{\em BMTUC(2,2)}  & 4  & 0.06 & 0.07 & 0.03 & 0.03 & 16.11 & 0.06 & 0.11\\
{\em BMTUC(4,2)}  & 8  & 0.10 & 0.23 & 0.24 & 0.07 & TO    & 0.09 & 0.14\\
{\em BMTUC(6,2)}  & 12 & 0.14 & 19.88 & 1368.28 & 0.43 & TO& 0.08 & 0.14\\
{\em BMTUC(8,4)}  & 16 & 0.56 & TO & TO & TO & TO          & 0.13 & 0.18\\
{\em BMTUC(10,4)} & 20 & 0.63 & TO & TO & TO & TO          & 0.16 & 0.16\\
\cline{1-9}
{\em RING(2)}     & 5  & 0.12 & 0.47 & 0.201 & 0.04 & 0.14 & 0.05 & 0.00 \\
{\em RING(4)}     & 11 & 0.21 & 6.76 & 0.638 & 0.05 & 2.28 & 0.09 & 0.12 \\
{\em RING(6)}     & 17 & 31.73& TO   & TO    & 0.40 & 77.10& 0.20 & 0.13 \\
{\em RING(8)}     & 23&1246.58& TO   & TO    &832.73& TO   & 0.74 & 0.18\\
{\em RING(10)}    & 29 &  TO  & TO   & TO    & TO   & TO   & 2.46 & 0.18\\
\cline{1-9}
{\em DOM(10)}  & 1  & 0.11 & 0.08 & 0.03   & 0.04& 2.24 & 0.05 & 0.13    \\
{\em DOM(20)}  & 1  & 0.14 & 0.07 & 0.24   & 0.05& 33.4 & 0.29 & 0.14   \\
{\em DOM(50)}  & 1  & 0.47 & 0.40 & 1368.28 & 0.06& 1315.98&4.44 &1.34 \\
{\em DOM(100)} & 1  & 1.70 & 1.64 & TO     & 0.11& TO & TO & 2.56\\
{\em DOM(500)} & 1  &31.28 & 32.52& TO     & 2.16& TO & TO & 29.10\\
{\em DOM(1000)}& 1  &121.91&129.96& TO     & 9.83& TO & TO & TO\\
\cline{1-9}
\end{tabular}
\end{small}
\caption{Conformant Planning Performance\label{tab:conf}}
\end{table}

In conformant setting (Table \ref{tab:conf}), it is noticeable that
\cps$^{c}$ behaves better than \cps$^{s}$ in all the 
conformant benchmark domains, especially in large problems.
Furthermore, CFF and KACMBP are superior to all the other planners
on most of the testing problems. Especially, both of them
scale up to larger instances very well, compared 
with the others. Yet, it is interesting to observe that 
\cps$^c$ does not lose out a whole lot against these two planners 
in many problems. In the following, we will discuss the performance
\cps\ in comparison with \cmbp, \dlvk, and \sgp.

It can be seen that \cps$^{c}$ is competitive with \cmbp\ and 
outperforms \dlvk\ and \sgp\ in most of problems. 
Specifically, in the $BT$ domain, \cps$^{c}$
took only 0.12 seconds to solve the last problem,
while \dlvk, \cmbp, and \sgp\ took 11.37, 0.5 and 2.13
seconds respectively. \cps$^{s}$ however is slower
than \cmbp\ and \sgp\ in this domain. 

\begin{table}[h]
\begin{small}
\begin{tabular}{|l|c|r|r|r|r|r|}
\cline{1-7}
Problem & Min. Plan & \multicolumn{2}{c|}{\cps} & \sgp & POND & MBP\\
\cline{3-4}
        & Length \& Width & cmodels & smodels & & & \\
\cline{1-7}
{\em BTS1(2)} & 2x2 & 0.166 & 0.088 & 0.11 & 0.188 & 0.047 \\
{\em BTS1(4)} & 4x4 & 0.808 & 1.697 & 0.22 & 0.189 & 0.048 \\
{\em BTS1(6)} & 6x6 & 5.959 & 83.245 & 2.44 & 0.233 & 0.055 \\
{\em BTS1(8)} & 8x8 & 25.284 & TO & 24.24 & 0.346 & 0.076 \\
{\em BTS1(10)} & 10x10 & 85.476 & TO & TO & 0.918 & 0.384 \\
\cline{1-7}
{\em BTS2(2)} & 2x2 & 0.39 & 0.102 & 0.19 & 0.186 & 0.038 \\
{\em BTS2(4)} & 4x4 & 1.143 & 3.858 & 0.32 & 0.198 & 0.067 \\
{\em BTS2(6)} & 6x6 & 19.478 & 1515.288 & 3.23 & 0.253 & 2.163 \\
{\em BTS2(8)} & 8x8 & 245.902 & TO & 25.5 & 0.452 & 109.867 \\
{\em BTS2(10)} & 10x10 & 345.498 & TO & TO & 1.627 & 178.823 \\
\cline{1-7}
{\em BTS3(2)} & 2x2 & 0.357 & 0.13 & 0.22 & 0.185 & 0.082 \\
{\em BTS3(4)} & 4x4 & 1.099 & 5.329 & 0.44 & 0.195 & 1.93 \\
{\em BTS3(6)} & 6x6 & 7.055 & TO & 3.89 & 0.258 & 147.76 \\
{\em BTS3(8)} & 8x8 & 56.246 & TO & 28.41 & 0.549 & AB \\
{\em BTS3(10)} & 10x10 & 248.171 & TO & TO & 2.675 & AB \\
\cline{1-7}
{\em BTS4(2)} & 2x2 & 0.236 & 0.149 & 0.26 & 0.194 & 0.098 \\
{\em BTS4(4)} & 4x4 & 1.696 & 3.556 & 0.64 & 0.191 & AB \\
{\em BTS4(6)} & 6x6 & 13.966 & 149.723 & 4.92 & 0.264 & AB \\
{\em BTS4(8)} & 8x8 & 115.28 & TO & 30.34 & 0.708 & AB \\
{\em BTS4(10)} & 10x10 & 126.439 & TO & TO & 4.051 & AB \\
\cline{1-7}
{\em MED(1)} & 1x1 & 1.444 & 1.434 & 0.09 & 0.187 & 0.048 \\
{\em MED(2)} & 5x5 & 35.989 & 9.981 & 0.59 & 0.193 & 0.047 \\
{\em MED(3)} & 5x5 & 42.791 & 9.752 & 1.39 & 0.2 & 0.049 \\
{\em MED(4)} & 5x5 & 39.501 & 10.118 & 7.18 & 0.205 & 0.049 \\
{\em MED(5)} & 5x5 & 35.963 & 9.909 & 44.64 & AB & 0.05 \\
\cline{1-7}
{\em SICK(2)} & 3x2 & 0.234 & 0.121 & 0.21 & 0.189 & 0.045 \\
{\em SICK(4)} & 3x4 & 0.901 & 0.797 & 10.29 & 0.19 & 0.048 \\
{\em SICK(6)} & 3x6 & 5.394 & 3.9 & TO & 0.201 & 0.059 \\
{\em SICK(8)} & 3x8 & 17.18 & 14.025 & TO & 0.221 & 0.129 \\
{\em SICK(10)} & 3x10 & 82.179 & 43.709 & TO & 0.261 & 0.778 \\
\cline{1-7}
{\em RINGS(1)} & 3x3 & 0.768 & 0.14 & 0.67 & 0.198 & 0.045 \\
{\em RINGS(2)} & 7x9 & 1386.299 & TO & TO & 0.206 & 0.057 \\
{\em RINGS(3)} & 11x27 & TO & TO & TO & 0.391 & 0.207 \\
{\em RINGS(4)} & 15x64 & TO & TO & TO & 3.054 & 3.168 \\
\cline{1-7}
{\em DOMS(1)} & 3x1 & 0.117 & 0.203 & 0.11 & 0.08 & 0.043 \\
{\em DOMS(2)} & 5x4 & 0.306 & 0.325 & 48.82 & 0.183 & 0.048 \\
{\em DOMS(3)} & 7x8 & 3.646 & 53.91 & TO & 0.19 & 0.057 \\
{\em DOMS(4)} & 9x16 & 87.639 & TO & TO & 0.248 & 0.101 \\
{\em DOMS(5)} & 11x32 & TO & TO & TO & 0.687 & 0.486\\
\cline{1-7}
\end{tabular}
\end{small}
\caption{Conditional Planning Performance\label{tab:cond}}
\end{table}

In the $BMT$ domain, \cps$^{s}$ is the worst. 
\cps$^s$ took more than two minutes to solve the 
largest problem in this domain, while \cmbp\ took only
0.53 seconds. \cps$^{c}$, however, is competitive
with \cmbp\ and outperforms both \dlvk\ and \sgp.

In the $BTC$ domain, although
\cps$^{s}$ is better than \dlvk\ and \sgp, its performance
is far from that of \cmbp. The time for \cps$^{s}$ to
solve the largest problem is nearly 8 minutes, while 
that for \cmbp\ is just 9.76 seconds. Again, \cps$^{c}$
is the best. It took only 0.12 seconds to solve
the same problem.

The $BMTC$ domain turns out to be hard for \dlvk, \cmbp, and \sgp. 
None of them were able to solve the $BMTC(8,4)$ within the time limit.
Although \cps$^{s}$ was able to solve this instance,
it could not solve the last instance. \cps$^{c}$ on the 
contrary can solve these instances very quickly, 
less than two seconds for each problem.

In the $BTUC$ and $BMTUC$ domains, although not competitive with
\cps$^{c}$, \cmbp\ outperforms both \dlvk\ and \sgp.
For example, \cmbp\ took less than 12 seconds to solve the largest
instance in the $BTUC$ domain, while \cps$^{s}$, \dlvk, and 
\sgp\ indicated a timeout. \cps$^{s}$ is competitive 
with \dlvk\ and much better than \sgp. Its performance is 
worse than \cmbp\ in these domains however. 

The $RING$ domain is really hard for the planners except
CFF and KACMBP. CFF and KACMBP took just a few minutes
to solve the largest problem; however, KACMBP seems
to scale up better than CFF on this domain. None of
the other planners could solve the last problem. Among
the others, \cmbp\ is the best, followed by \cps$^{c}$. 
\cmbp\ took around 14 minutes to solve $RING(8)$ while
\cps$^{c}$ took more than 20 minutes. \cps$^s$ is 
outperformed by both \dlvk\ and \sgp.

In the last domain, $DOM$, again, \cmbp\ outperforms
\cps, \dlvk, and \sgp. The solving time of \cps\ for the last problem is
around 2 minutes, while that for \cmbp\ is just less
than 10 seconds. \dlvk\ and \sgp\ were able to solve the
first three instances of this domain only. It is
worth noting here that the not-very-good performance
of CFF and KACMBP on this domain is because that
this domain is in nature very rich in static causal laws, a 
feature that is not supported by CFF and KACMBP. 
Therefore, to encode the domain in CFF and KACMBP,
we had to compile away static causal laws. 

The performance of \cps\ in the conditional
benchmarks is not as good as in the conformant benchmarks,
compared with other testing planners. As can
be seen in Table \ref{tab:cond},
it was outperformed by both POND and
MBP in the benchmarks, except in the last two problems
of the $BTS3$ domain or in the last three of the $BTS4$, 
where $MBP$ had a problem with segmentation fault or 
memory excess, or in $MED(5)$ problem
where POND stopped abnormally. Both POND and MBP
did very good at testing domains. POND took just a few
seconds to solve each instance in the testing domains.
\cps\ is also not competitive with \sgp\ in small instances of 
the first five domains ($BTS1$-$MED$). 
However, when scaling up to larger problems, 
\cps$^{c}$ seems to be better than \sgp. In the last
three domains ($SICK$, $RINGS$, and $DOMS$), \sgp\ is
outperformed by both \cps$^c$ and \cps$^s$.

\section{Conclusion and Future Work}
\label{conclusion}

In this paper, we define an approximation for action theories 
with static causal laws and sensing actions. 
We prove that the newly developed approximation is 
sound with respect to the possible world semantics and 
is deterministic when non-sensing actions are executed. We also 
show that the approximation reduces the complexity of the 
conditional planning problem. 

We use the approximation to develop an answer set programming 
based conditional planner, called \cps. \cps\ differs from 
previously developed model-based planners for domains 
with incomplete initial state (e.g. 
\cite{gef98,cim99,eit03,smi98}) in that it is capable of dealing 
with sensing actions and generating both conditional and 
conformant plans. We prove the correctness of \cps\ by 
showing that plans generated by \cps\ are solutions of 
the encoded planning problem instances. 
Furthermore, we prove that \cps\ will generate a solution 
to $\calp$ if it has a solution with respect to the given
approximation. We also discuss the use of \cps\ in 
reasoning about effects of conditional plans. 

We compare \cps\ with several planners. 
These results provide evidence for the usefulness of answer set 
planning in dealing with sensing actions and incomplete information. 
Our experiments also show that there are situations in which \cps\ 
does not work as well as other state-of-the-art planners. 
In the future, we would like to 
investigate methods such as the use of domain knowledge to 
speed up the planning process (see e.g. \cite{SonBTM05}). 

\medskip
\ni {\bf Acknowledgment.} We would like to thank Michael Gelfond 
for his valuable comments on an earlier draft of this paper. 
We would also like to thank the anonymous reviewers of this paper 
and an extended 
abstract of this paper, which appeared in  \cite{SonTB04}, 
for their constructive comments and suggestions. 
The first two authors were partially supported by 
NSF grant EIA-0220590.

\section*{Appendix A -- Proofs related to the 0-Approximation}

This appendix contains the proofs for the propositions and
theorems given in the paper. As stated, we assume that the body of 
each static law (\ref{static}) is not an empty set and 
$\calg \ne \emptyset$ for every planning problem $(\cald,\cali,\calg)$. 

We begin with a lemma 
about the operator $Cl_\cald$ that will be
used in these proofs. We need the following definition.
Given a domain description $\cald$, for a set of literals $\sigma$, 
let $$\Gamma(\sigma) = \sigma \cup \{ l \mid \exists 
\caused(l, \varphi) \in \cald
\textnormal{ such that } \varphi \subseteq \sigma \}.$$
Let $\Gamma^0(\sigma) = \Gamma(\sigma)$ and 
$\Gamma^{i+1}(\sigma) = \Gamma(\Gamma^i(\sigma))$ 
for $i \ge 0$. Since, by the definition of $\Gamma$,
for any set of literals $\sigma'$ we have 
$\sigma' \subseteq \Gamma(\sigma')$, the sequence 
$\langle \Gamma^i(\sigma) \rangle_{i=0}^{\infty}$ is 
monotonic with respect to the set inclusion operation. 
In addition, $\langle \Gamma^i(\sigma) 
\rangle_{i=0}^{\infty}$ is bounded by the set of fluent 
literals. Thus, there exists $\sigma^{\limit}$ such that 
$\sigma^{\limit}_{\cald} = \bigcup_{i=0}^{\infty} 
\Gamma^i(\sigma)$. Furthermore, $\sigma^{\limit}_{\cald}$ is unique 
and satisfies all static causal laws in $\cald$. 

\begin{lemma}
\label{lm1}
For any set of literals $\sigma$, we have
$\sigma^{\limit}_{\cald} = Cl_{\cald}(\sigma)$.
\end{lemma}
\begin{proof}
By induction we can easily
show that $\Gamma^i(\sigma) \subseteq Cl_{\cald}(\sigma)$
for all $i \ge 0$. Hence, we have
$$\sigma^{\limit}_{\cald} \subseteq Cl_{\cald}(\sigma)$$
Furthermore, from the construction of $\Gamma^i(\sigma)$,
it follows that $\sigma^{\limit}$ satisfies all static causal
laws in $\cald$. Because of the minimality 
property of $Cl_{\cald}(\sigma)$, we have 
$$Cl_{\cald}(\sigma) \subseteq \sigma^{\limit}_{\cald}$$

Accordingly, we have $$\sigma^{\limit}_{\cald}
= Cl_{\cald}(\sigma)$$
\end{proof}
The following corollary follows immediately from the above 
lemma.
\begin{corollary}
\label{corol1}
For two sets of literals $\sigma \subseteq \sigma'$, 
$Cl_{\cald}(\sigma) \subseteq Cl_{\cald}(\sigma')$.
\end{corollary}

For an action $a$ and a state $s$, let $e(a,s) = Cl_\cald(E(a,s))$.
We have the following lemma:

\begin{lemma}
\label{lm1b}
Let $a$ be an action and $s,s'$ be states. Then, we have
$$Cl_{\cald}(E(a,s) \cup (s \cap s')) = Cl_{\cald}(e(a,s) \cup (s \cap s'))$$
\end{lemma}
\begin{proof}
Let $\gamma = E(a,s) \cup (s \cap s')$ and $\gamma' = e(a,s) \cup (s \cap s')$.
As $\gamma \subseteq \gamma'$, it follows from Corollary \ref{corol1} 
that to prove this lemma, it suffices to prove that 
$$Cl_{\cald}(\gamma') \subseteq Cl_{\cald}(\gamma)$$

It is easy to see that
$$\gamma' = Cl_{\cald}(E(a,s)) \cup (s \cap s') \subseteq 
Cl_{\cald}(E(a,s) \cup (s \cap s')) = Cl_{\cald}(\gamma)$$
Therefore, by Corollary \ref{corol1}, we have
$$Cl_{\cald}(\gamma') \subseteq Cl_{\cald}(Cl_{\cald}(\gamma))
= Cl_{\cald}(\gamma)$$

Proof done.
\end{proof}
\subsection*{Proof of Proposition \ref{prop-res-nonsense}}
\begin{lemma}
\label{lm-prop-res-nonsense}
For every state $s' \in Res^c_{\cald}(a,s)$, we have
$$s' \setminus (e(a,s) \cup (s \cap s')) \subseteq pc(a,\delta)$$
\end{lemma}
\begin{proof}
Let $\sigma$ denote $e(a,s) \cup (s \cap s')$.
By Corollary \ref{corol1}, since $e(a,\delta) \subseteq e(a,s) \subseteq
\sigma$, we have
\begin{eqnarray}
Cl_{\cald}(e(a,\delta)) \subseteq Cl_{\cald}(\sigma) = s' \label{lm-sound1-0}
\end{eqnarray}

We now show that, for every $i \ge 1$, 
\begin{eqnarray}
\Gamma^{i}(\sigma) \setminus \Gamma^{i-1}(\sigma) \subseteq 
pc^{i}(a,\delta) \label{lm-sound1-1}
\end{eqnarray}
by induction on $i$.
\begin{itemize}
\item[1.]{\bf Base case: } $i = 1$. Let
$l$ be a literal in $\Gamma^1(\sigma) \setminus \Gamma^0(\sigma)$. We
need to prove that $l \in pc^1(a,\delta)$.

By the definition of $\Gamma$, it follows that
\begin{eqnarray}
l \not \in \Gamma^0(\sigma) = \sigma \label{lm-sound1-2} \\
l \in \Gamma^1(\sigma) \subseteq s' \label{lm-sound1-2_1} 
\end{eqnarray}
and, in addition, there exists a static causal law
$$\caused(l,\varphi)$$
in $\cald$ such that
\begin{eqnarray}
\varphi \subseteq \Gamma^0(\sigma) = \sigma \label{lm-sound1-3} 
\end{eqnarray}
By (\ref{lm-sound1-2}), we have $l \not \in (s \cap s')$.
By (\ref{lm-sound1-2_1}), we have $l \in s'$. Accordingly,
we have $l \not \in s$. On the other hand, because $\delta 
\subseteq s$, we have
\begin{eqnarray}
l \not \in \delta \label{lm-sound1-3_1} 
\end{eqnarray}
It follows from (\ref{lm-sound1-3}) that  $\varphi \subseteq s'$
since $\sigma \subseteq s'$. Because of the completeness of
$s'$, we have $\myneg{\varphi} \cap s' = \emptyset$.
On the other hand, by (\ref{lm-sound1-0}), we have 
$Cl_{\cald}(e(a,\delta)) \subseteq s'$. As a result, we have 
\begin{eqnarray}
& \myneg{\varphi} \cap Cl_{\cald}(e(a,\delta)) = 
\emptyset \label{lm-sound1-4}
\end{eqnarray}
We now show that $\varphi \not \subseteq s$.
Suppose otherwise, that is, $\varphi \subseteq s$. 
This implies that $l \in s$. By (\ref{lm-sound1-2_1}), 
it follows that $l \in (s \cap s') \subseteq \sigma$ 
and this is a contradiction to (\ref{lm-sound1-2}).
Thus, $\varphi \not \subseteq s$. 

On the other hand, we know that $\varphi \subseteq \sigma = e(a,s)
\cup (s \cap s')$ and thus we have $\varphi \cap (e(a,s) \setminus
s) \ne \emptyset$. In addition, it is easy to see that
$e(a,s) \setminus s \subseteq e(a,s) \setminus \delta \subseteq
pc^0(a,\delta)$. Therefore, we have
\begin{eqnarray}
\varphi \cap pc^0(a,\delta) \ne \emptyset \label{lm-sound1-5}
\end{eqnarray}
From (\ref{lm-sound1-3_1}) -- (\ref{lm-sound1-5}), 
and by the definition of $pc^1(a,\delta)$, 
we can conclude that $l \in pc^1(a,\delta)$. 
The base case is thus true.

\item[2.]{\bf Inductive Step: } Assume that (\ref{lm-sound1-1}) is true
for all $i \le k$. We need to prove that it is true for $i = k+1$.
Let $l$ be a literal in $\Gamma^{k+1}(\sigma) \setminus \Gamma^{k}(\sigma)$.
We will show that $l \in pc^{k+1}(a,\delta)$.

By the definition of $\Gamma$, there exists a static causal law
$$\caused(l,\varphi)$$
in $\cald$ such that
\begin{eqnarray}
& \varphi \subseteq \Gamma^k(\sigma) \subseteq s' \label{lm-sound1-6} 
\end{eqnarray}
Because $\varphi \subset s'$, we have $\myneg{\varphi} \cap s'
= \emptyset$. In addition, by (\ref{lm-sound1-0}),
$Cl_{\cald}(e(a,\delta))$ is a subset of $s'$. A a result, we have
\begin{eqnarray}
\myneg{\varphi} \cap Cl_{\cald}(e(a,\delta)) = \emptyset \label{lm-sound1-8}
\end{eqnarray}
It is easy to see that $\varphi \not \subseteq \Gamma^{k-1}(\sigma)$ 
for if otherwise then, by the definition of $\Gamma$, $l$
must be in $\Gamma^k(\sigma)$, 
which is impossible. In other words,
there exists $l' \in \varphi$ such that $l' \not \in \Gamma^{k-1}(\sigma)$ 
but $l' \in \Gamma^{k}(\sigma)$. By the inductive hypothesis, we have
$l' \in pc^k(a,\delta)$, which implies that
\begin{eqnarray}
\varphi \cap pc^k(a,\delta) \ne \emptyset \label{lm-sound1-9}
\end{eqnarray}
Because $l \not \in \Gamma^k(\sigma)$, 
we have $l \not \in \sigma$. As a result,
$l \not \in (s \cap s')$. On the other hand, since
$l \in \Gamma^{k+1}(\sigma) \subseteq s'$, it follows that 
$l \not \in s$. Thus, we have
\begin{eqnarray}
l \not \in \delta \label{lm-sound1-10}
\end{eqnarray}
 From (\ref{lm-sound1-8}) -- (\ref{lm-sound1-10}), and
by the definition of $pc^{k+1}(a,\delta)$, it follows that 
$l \in pc^{k+1}(a,\delta)$. So the inductive step is proven.
\end{itemize}

As a result, it is always the case that (\ref{lm-sound1-1})
holds. Hence, we have
$$\Gamma^i(\sigma) \setminus \sigma \subseteq 
\bigcup_{j=0}^i(pc^j(a,\delta)) = pc^i(a,\delta)$$
and thus,
$$\bigcup_{i=0}^{\infty} (\Gamma^i(\sigma) \setminus \sigma)
\subseteq \bigcup_{i=0}^{\infty} pc^i(a,\delta)$$
Accordingly, by Lemma (\ref{lm1}) and
by the definition of $pc(a,\delta)$, we have
$$(s' \setminus \sigma) \subseteq pc(a,\delta).$$
The lemma is thus true.
\end{proof}

We now prove Proposition \ref{prop-res-nonsense}.
Let 
$$\gamma = e(a,\delta) \cup (\delta \setminus \myneg{pc(a,\delta)})
\; \; \; \; \; \; \; \delta' = Cl_\cald(\gamma)$$
Let $s'$ be some state in $Res^c_{\cald}(a,s)$. Such an $s'$ 
exists because $\cald$ is consistent. By Lemma \ref{lm1b}
and by Definition \ref{def_compl}, we have 
\begin{eqnarray}
s' = Cl_{\cald}(\sigma) \label{lm-sound1ness-a}
\end{eqnarray}
where
$$\sigma = e(a,s) \cup (s \cap s')$$

To prove Proposition \ref{prop-res-nonsense}, it suffices
to prove that $\delta' \subseteq s'$. But first of all, 
let us prove, by induction, the following
\begin{eqnarray}
\Gamma^i(\gamma) \subseteq s' \label{lm-sound1ness-0}
\end{eqnarray}
for every integer $i \ge 0$.
\begin{itemize}
\item[1.] {\bf Base Case: }$i = 0$. Assume that 
$l \in \Gamma^0(\gamma) = \gamma$. We need to show that 
$l \in s'$. There are two possibilities for $l \in \gamma$.
\begin{itemize}
\item[a)] {\em $l \in e(a,\delta)$}. It is easy to see that 
$l \in s'$ because
$$e(a,\delta) \subseteq
e(a,s) \subseteq \sigma \subseteq Cl_{\cald}(\sigma) = s'.$$

\item[b)] {\em $l \not \in e(a,\delta)$, $l \in \delta$,
and $\myneg{l} \not \in pc(a,\delta)$}. Since $\delta \subseteq
s$, we have $l \in s$. Because of the completeness of $s$,
it follows that $\myneg{l} \not \in s$. Accordingly, we have
\begin{eqnarray}
& \myneg{l} \not \in (s \cap s') \label{lm-sound1ness-1}
\end{eqnarray}
On the other hand, because $\myneg{l} \not \in pc(a,\delta)$,
$\myneg{l} \not \in s$, and $(e(a,s) \setminus s) 
\subseteq pc^0(a,\delta) \subseteq pc(a,\delta)$,
we have
\begin{eqnarray}
& \myneg{l} \not \in e(a,s) \label{lm-sound1ness-2}
\end{eqnarray}
    From (\ref{lm-sound1ness-1}) and (\ref{lm-sound1ness-2}),
it follows that $\myneg{l} \not \in \sigma$.
In addition, since $\myneg{l} \not \in pc(a,\delta)$, by Lemma 
\ref{lm-prop-res-nonsense}, 
we have $\myneg{l} \not \in s' \setminus \sigma$.
Accordingly, we have $\myneg{l} \not \in s'$. Because $s'$ is complete, 
we can conclude that $l \in s'$.
\end{itemize}
\item[2.] {\bf Inductive Step: } Assume that (\ref{lm-sound1ness-0})
is true for all $i \le k$. We need to show that 
$\Gamma^{k+1}(\gamma)\subseteq s'$.
Let $l$ be a literal in $\Gamma^{k+1}(\gamma)$. By the definition of
$\Gamma^{k+1}(\gamma)$, there are two possibilities for $l$:

\begin{itemize}
\item[a)] {\em $l \in \Gamma^{k}(\gamma)$}. Clearly, in this case, we have 
$l \in s'$. 
\item[b)] {\em there exists a static causal law
$$\caused(l,\varphi)$$ 
in $\cald$ such that 
$\varphi \subseteq \Gamma^k(\gamma)$}. 

By the inductive hypothesis, we have $\varphi \subseteq s'$.
Hence, $l$ must hold in $s'$.
\end{itemize}
Therefore, in both cases, we have $l \in s'$. This implies that
$\Gamma^{k+1}(\gamma) \subseteq s'$.
\end{itemize}
As a result, (\ref{lm-sound1ness-0}) always holds. 
By Lemma \ref{lm1}, we have
$$\delta' = \bigcup_{i=0}^{\infty} \Gamma^i(\gamma) \subseteq
s'$$
Since $s'$ is a state, $\delta'$ is consistent. Thus,
by the definition of the $Res$-function, we have
$$Res_\cald(a,\delta) = \{ \delta'\}$$
Furthermore, $\delta' \subseteq s'$ for every $s' \in
Res^c_\cald(a,s)$.

The proposition is proven.
\subsection*{Proof of Proposition \ref{prop-res-sense}}
Since $\delta$ is valid, there exists a state $s$ such
that $\delta \subseteq s$.

On the other hand, we assume that in every state of the world,
exactly one literal in $\theta$ holds, there exists a literal
$g \in \theta$ such that $g$ holds in $s$ and for all
$g' \in \theta \setminus \{g\}$, $g'$ does not hold in
$s$.

Accordingly, we have $\delta \cup \{ g \} \subseteq s$.
By Corollary \ref{corol1}, we have 
$\delta' = Cl_\cald(\delta \cup \{g\}) \subseteq Cl_\cald(s)
= s$. Hence, $\delta'$ is consistent. By the definition
of the $Res-$function, we have $\delta' \in Res_\cald(a,\delta)$. 
Since $\delta' \subseteq s$, $\delta'$ is a valid a-state.

The proposition is thus true.
\subsection*{Proof of Proposition \ref{prop-cons-action-theory}}
Let us prove this proposition by using structural induction on $p$.
\begin{itemize}
\item[1.] {\em $p = []$}. 
Trivial.
\item[2.] {\em $p = [a;q]$, where $q$ is a conditional plan and
$a$ is a non-sensing action}. 

Assume that Proposition 
\ref{prop-cons-action-theory} is true for $q$. 
We need to prove that it is also true for $p$.

Suppose $\hat \Phi(p,\delta) \ne \bot$. Clearly we have
$\Phi(a,\delta) \ne \bot$.

Therefore, we have $\Phi(a,\delta) = Res_\cald(a,\delta)$.
On the other hand, since $\delta$ is a valid a-state,
it follows from Proposition \ref{prop-res-nonsense} that
$Res_\cald(a,\delta) = \{ \delta' \}$ for some
valid a-state $\delta'$.

As a result, we have $\hat \Phi(q,\delta')$ 
contains at least one valid a-state. 
Hence, $\hat \Phi(p,\delta) \ne \bot$ contains at least 
one valid a-state.
\item[3.] {\em $p = [a;\kcases(\{g_j \rightarrow p_j\}_{j=1}^n)]$, where
a is a sensing action that senses $g_1,\dots,g_n$}.

Assume that Proposition \ref{prop-cons-action-theory}
is true for $p_j$'s. We need to prove that it is also true
for $p$.

Because $\hat \Phi(p,\delta) \ne \bot$, we have
$\Phi(a,\delta) \ne \bot$. By the definition of the
$\Phi$-function, we have $\Phi(a,\delta) = Res_\cald(a,\delta)$.
As $\delta$ is valid, by Proposition \ref{prop-res-sense},
$Res_\cald(a,\delta)$ contains at least one valid a-state 
$\delta'$.

By the definition of the $Res-$function for sensing actions,
we know that $\delta' = Cl_\cald(\delta \cup \{ g_k \})$
for some $k$. This implies that $g_k$ holds in $\delta'$.

By the inductive hypothesis, we have $\hat \Phi(p_k,\delta')$
contains at least one valid a-state.

By the definition of the $\hat\Phi$-function, 
we have $\hat \Phi(p_k,\delta') \subseteq \hat \Phi(p,\delta)$.
Thus, $\Phi(p,\delta)$ contains at least one valid a-state.
\end{itemize}
\subsection*{Proof of Proposition \ref{prop-trans1}}
Let $n$ denote the size of $\cald$. Because of Lemma \ref{lm1},
we can conclude that for any set of literals $\sigma$,
computing $Cl_\cald(\sigma)$ can be done in polynomial time
in $n$. 

Observe that for a non-sensing action $a$ and 
an a-state $\delta$, computing 
$e(a,\delta)$ and  $pc(a,\delta)$ can be done in 
polynomial time in $n$. Thus,  computing 
$\Phi(a,\delta)$ can be done in polynomial time in $n$.

Likewise, computing $\Phi(a,\delta)$ for a sensing
action $a$ can also be done in polynomial time in $n$.

Hence, Proposition \ref{prop-trans1} holds.
\subsection*{Proof of Theorem \ref{t-pp}}
The proof is similar to the proof of Theorem 3 in \cite{baral:ijcai99:aij} 
which states that the conditional planning problem  
with respect to the 0-approximation in \cite{sonbaral00} is
{\bf NP}-complete. Membership follows from Corollary \ref{cor1}. 
Hardness follows from the fact that the approximation proposed in this 
paper coincides with the 0-approximation in \cite{sonbaral00}, i.e,
the conditional planning problem considered in this paper coincides 
with the planning problem with limited-sensing  
in \cite{baral:ijcai99:aij} which is {\bf NP}-complete. By the 
restriction principle, we conclude that the problem considered in 
this paper is also {\bf NP}-complete.

\section*{Appendix B -- Proofs related to $\pi$}

This section contain proofs related to the correctness of $\pi$.
Before we present the proofs, let us introduce some notations 
that will be used throughout 
the rest of the appendix. Given a program $\Pi$, by $lit(\Pi)$ we mean the
set of atoms in $\Pi$. If $Z$ is a splitting set for $\Pi$
and $\Sigma$ is a set of atoms then by $b_Z(\Pi)$ and
$e_Z(\Pi \setminus b_Z(\Pi),\Sigma)$, we mean the bottom part of 
$\Pi$ w.r.t. $Z$ and the evaluation of the top part w.r.t. $(Z,\Sigma)$
(see \cite{lif94a} for more information about these notions). 

\begin{lemma}
\label{l3}
\begin{itemize}
\item[1.] Let $\Pi$ be a logic program. Suppose $\Pi$ can be divided into two
disjoint subprograms $\Pi_1$ and $\Pi_2$, i.e., $\Pi = \Pi_1 \cup \Pi_2$
and $lit(\Pi_1) \cap lit(\Pi_2) = \emptyset$. Then $S$ is an answer
set for $\Pi$ if and only if there exist two sets $S_1$ and $S_2$ 
of atoms such that $S = S_1 \cup S_2$ and $S_1$ and $S_2$ are 
answer sets for $\Pi_1$ and $\Pi_2$ respectively.
\item[2.] The result in Item 1 can be generalized to $n$ disjoint 
subprograms, where $n$ is an arbitrary integer.
\end{itemize}
\end{lemma}
\begin{proof}
The first item can easily proved by using the splitting set 
$Z=lit(\Pi_1)$. The second item immediately follows from this result.
\end{proof}

\subsection*{Proof of Theorem \ref{t1}}
\label{proof1}
Suppose we are given a planning problem instance
$\calp = (\cald,\cali,\calg)$ and $\pi_{h,w}(\calp)$, where $h \ge 1$ and
$w \ge 1$ are some integers, returns an answer set $S$. 
The proof is primarily based on the splitting set and 
splitting sequence theorems described in \cite{lif94a}.
It is organized as follows. 
We first prove a lemma related to the closure of a set of 
literals (Lemma \ref{l2}). Together with Lemma \ref{l3},
this lemma is used to prove some properties of 
$\pi_{h,w}(\calp)$ (Lemmas \ref{l4},
\ref{l51} \& \ref{l52}). Based on these results, we prove the 
correctness of $\pi_{h,w}(\calp)$ in implementing the $\Phi$
and $\hat \Phi$ functions (Lemma \ref{l6} \& Lemma \ref{l7}).
Theorem \ref{t1} can be derived directly from Lemma \ref{l7}. 

Recall that we have made certain 
assumptions for action theories given to \cps:
{\em (a)} for every k-proposition 
$\determines(a,\eta)$, $\eta$ contains at least two elements;
and {\em (b)} for every static causal law
$\caused(f, \phi)$, $\phi$ is not an empty set.

The following lemma shows a code fragment that correctly encodes 
the closure of a set of literals.\begin{lemma}
\label{l2}
Let $i$ and $k$  be two integers greater than 0,
and $x$ be a 3-ary predicate. For any set $\sigma$ of
literals, the following program
$$
\begin{array}{lll}
x(l,i,k) \la & \; \; \; & (l \in \sigma)\\
x(l,i,k) \la x(\varphi,i,k) & \; \; \; & 
(\caused(l, \varphi) \in \cald)
\end{array}
$$
has the unique answer set $\{ x(l,i,k) \mid l \in 
Cl_{\cald}(\sigma) \}$.
\end{lemma}
\begin{proof}
By the definition of a model of a positive program, it is easy
to see that the above program has the unique answer set 
$\{x(l,i,k) \mid l \in \sigma^{\limit}_{\cald} \} = 
\{ x(l,i,k) \mid l \in Cl_{\cald}(\sigma)\}$ (see Lemma \ref{lm1}).
\end{proof}
Before showing some lemmas about the properties of 
$\pi_{h,w}(\calp)$, let us
introduce some notions and definitions that will be used 
throughout the rest of this section. We first define some sets of atoms
which will frequently be used in the proofs of Lemmas 
\ref{l4} -- \ref{l7} and Theorem \ref{t1}.
Then we divide the program $\pi_{h,w}(\calp)$ into 
small parts to simplify the proofs. In particular,
$\pi_{h,w}(\calp)$ is divided into two programs $\pi^*_{h,w}(\calp)$
and $\pi^c_{h,w}(\calp)$. The former consists of normal logic program
rules while the latter consists of constraints in $\pi_{h,w}(\calp)$.
Then we use the splitting set theorem to remove from 
$\pi^*_{h,w}(\calp)$ auxiliary atoms such as $fluent(\dots)$, 
$literal(\dots)$, $time(\dots)$, $path(\dots)$, etc.
The resulting program, denoted by $\pi_0$, consists of ``main''
atoms only. We then use the splitting sequence theorem to
further split $\pi_0$ into a set of programs $\pi_i$'s. Intuitively,
each $\pi_i$ corresponds to a ``cut'' of $\pi_0$ at time point 
$i$. Finally, each $\pi_i$ is divided into disjoint subprograms
$\pi_i^k$'s, each of which, intuitively, is a ``cut'' of 
$\pi_i$ at a specific path. 

For $1 \le i \le h+1$ and $1 \le k \le w$, let $A_{i,k}$ be 
the set of all the atoms of the form $occ(a,i,k)$, $poss(a,i,k)$, 
$used(i,k)$, $goal(i,k)$, $holds(l,i,k)$, $br(g,i,k,k')$ ($k' \ge k$), 
$e(l,i,k)$, $pc(l,i,k)$, i.e.,
\begin{eqnarray}
A_{i,k} & = & \{occ(a,i,k), poss(a,i,k) \mid a \in {\bf A}\} \cup \nonumber \\
& & \{ holds(l,i,k), e(l,i,k), pc(l,i,k) \mid l 
    \textnormal{ is a literal} \} \cup \nonumber\\
& & \{br(g,i,k,k') \mid g \textnormal{ is a sensed-literal}, 
    k \le k' \le w\} \cup \nonumber\\
& & \{ used(i,k), goal(i,k) \} \label{eq2}
\end{eqnarray}
and let
\begin{eqnarray}
A_{i} = \bigcup_{k=1}^{w} A_{i,k}, \hspace{.5in} 
A = \bigcup_{i=1}^{h+1} A_{i} \label{eq3}
\end{eqnarray}
For a set of atoms $\Sigma \subseteq A$ and a set of predicate 
symbols $X$, by $\Sigma^X$ we denote the set of atoms in $\Sigma$ 
whose predicate symbols are in $X$ and by $\delta_{i,k}(\Sigma)$, 
we mean $\{l \mid holds(l,i,k) \in \Sigma\}$.

Observe that $\pi_{h,w}(\calp)$ can be divided into two parts (1)
$\pi^*_{h,w}(\calp)$ consisting of normal logic program rules, and (2)
$\pi^c_{h,w}(\calp)$ consisting of constraints. Since $S$ is 
an answer set for $\pi_{h,w}(\calp)$\footnote{Recall that at at the 
beginning of this section, we state that
$\pi_{h,w}(\calp)$ returns $S$ as an answer set.},
$S$ is also an answer set for $\pi^*_{h,w}(\calp)$ and
does not violate any constraint in $\pi^c_{h,w}(\calp)$.

Let $V$ be the set of atoms in $\pi_{h,w}(\calp)$ 
whose parameter list does not contain either the time  
or path variable. Specifically, $V$ is the following set of
atoms
\begin{eqnarray}
& \{ fluent(f), literal(f), literal(\neg f),
   contrary(f,\neg f), contrary(\neg f, f) \mid f \in {\bf F} \} 
   \cup \nonumber\\
& \{ sensed(g) \mid \exists \determines(a,\theta) \in \cald.
   g \in \theta\} \cup  \{ action(a) \mid a \in {\bf A} \} \cup  
  \nonumber \\
& \{time(t) \mid t \in \{1..h\}\} \cup \{time1(t) \mid t \in \{1..h+1\}\} \cup 
        \{path(p) \mid p \in \{1..w\}\} \label{def_V}
\end{eqnarray}
It is easy to see that $V$ is a splitting set for $\pi^*_{h,w}(\calp)$. 
Furthermore, the bottom part $b_V(\pi^*_{h,w}(\calp))$
is a positive program and has only one answer set $X_0=V$.
The partial evaluation of the top part of $\pi^*_{h,w}(\cal P)$ 
with respect to $X_0$, 
$$\pi_0 = e_V(\pi^*_{h,w}(\calp) \setminus b_V(\pi^*_{h,w}(\calp)), X_0),$$ 
is the following set of rules (the condition for each rule follows
that rule; and, by default $t$ and $p$ are in ranges
$1 \dots h$ and $1 \dots w$ unless otherwise specified):
\begin{eqnarray}
holds(l,1,1) & \la &  \label{r0_1} \\ 
  & & ( \initially(l) \in \cali ) \nonumber \\
poss(a,t,p) & \la & holds(\psi,t,p) \label{r0_2}\\
  & & ( \executable(a,\psi) \in \cald ) \nonumber \\
e(l,t,p) & \la & occ(a,t,p), holds(\phi,t,p) \label{r0_3} \\
  & & ( \causes(a,l,\phi) \in \cald ) \nonumber \\
pc(l,t,p) & \la & occ(a,t,p), \naf holds(l,t,p), 
  \naf holds(\phi,t,p) \label{r0_4} \\
  & & ( \causes(a,l,\phi) \in \cald ) \nonumber \\
br(g,t,p,p) \mid \dots & & \nonumber \\
  \mid br(g,t,p,w) & \la & occ(a,t,p) \label{r0_6} \\
  & & ( \determines(a,\theta) \in \cald, 
  g \in \theta )  \nonumber \\
pc(l,t,p)& \la & \naf holds(l,t,p), pc(l',t,p),
  \naf e(\neg \varphi,t,p) \label{r0_8}\\
  & & ( \caused(l,\varphi) \in \cald, l' \in \varphi ) \nonumber\\
e(l,t,p)& \la & e(\varphi,t,p) \label{r0_9}\\
  & & ( \caused(l,\varphi) \in \cald ) \nonumber \\
holds(l,t,p)& \la & holds(\varphi,t,p) \label{r0_10}\\
  & & ( \caused(l,\varphi) \in \cald, 1 \le t \le h+1 ) \nonumber \\
goal(t,p) & \la & holds(\calg,t,p) \label{r0_11} \\
 & & (1 \le t \le h+1) \nonumber\\
goal(t,p) & \la & holds(f,t,p), holds(\myneg f,t,p) \label{r0_11_2} \\
 & & (1 \le t \le h+1) \nonumber\\
holds(l,t{+}1,p) & \la & e(l,t,p) \label{r0_13}\\
holds(l,t{+}1,p) & \la & h(l,t,p), \naf pc(\myneg{l},t,p) \label{r0_14}\\
used(t{+}1,p) & \la &  br(g,t,p_1,p) \label{r0_19} \\
  & & ( p_1 < p) \nonumber \\
holds(g,t{+}1,p) & \la &  br(g,t,p_1,p) \label{r0_20} \\
  & & ( p_1 \le p ) \nonumber \\
holds(l,t{+}1,p) & \la &  br(g,t,p_1,p), holds(l,t,p_1) \label{r0_21} \\
  & & ( p_1 < p) \nonumber \\
occ(a_1,t,p) \mid \dots & & \nonumber \\
 occ(a_m,t,p) & \la & used(t,p), \naf goal(t,p) \label{r0_22} \\
used(1,1) & \la &  \label{r0_29} \\
used(t{+}1,p) & \la & used(t,p) \label{r0_30} 
\end{eqnarray}
And $\pi^c_{h,w}(\calp)$ is the following collection of constraints
\begin{eqnarray}
& \la & occ(a,t,p), \naf br(\theta,t,p,p) \label{r0_5}\\
  & & ( \determines(a,\theta) \in \cald ) \nonumber\\
& \la & occ(a,t,p), br(g,t,p,p_1), br(g,t,p,p_2) \label{r0_6_2} \\
  & & ( \determines(a,\theta) \in \cald, g \in \theta, p \le p_1 < p_2)  
  \nonumber \\
& \la & occ(a,t,p), holds(g,t,p) \label{r0_7} \\
  & & ( \determines(a,\theta) \in \cald, g \in \theta )  \nonumber \\
& \la & used(h{+}1,p), \naf goal(h{+}1,p) \label{r0_12} \\
& \la & br(g_1,t,p_1,p), br(g_2,t,p_2,p) \label{r0_16} \\
  & & ( p_1 < p_2 < p ) \nonumber \\
& \la & br(g_1,t,p_1,p), br(g_2,t,p_1,p) \label{r0_17} \\
  & & ( g_1 \ne g_2, p_1 \le p) \nonumber \\
& \la & br(g,t,p_1,p), used(t,p) \label{r0_18} \\
  & & ( p_1 < p ) \nonumber \\
& \la & used(t,p), \naf goal(t,p),
 occ(a_i,t,p), occ(a_j,t,p) \label{r0_22_2} \\
  & & (1 \le i < j \le m) \nonumber\\
& \la &  occ(a,t,p), \naf poss(a,t,p) \label{r0_23}
\end{eqnarray}
Note that choice rules of the form
$$1\{L_1,\dots,L_n\}1 \leftarrow Body$$
have been translated into
$$L_1 \mid \dots \mid L_n \leftarrow Body$$
and
$$\leftarrow Body, L_i, L_j \ \ \ (1 \le i < j \le n)$$

By the splitting set theorem, there exists an answer set $S_0$
for $\pi_0$ such that $S = S_0 \cup X_0$.
Let $U_i$ be the set of atoms in $\pi_0$ whose time parameter 
is less than or equal to $i$, i.e., 
\begin{equation}
U_i = \bigcup_{j=1}^{i} A_{j} \label{eq4}
\end{equation}
It is easy to see that the sequence $\langle U_i 
\rangle_{i=1}^{h{+}1}$ is a splitting sequence for $\pi_0$. 
By the splitting sequence theorem, since $S_0$ is an answer set 
for $\pi_0$, there must be a sequence of sets of literals 
$\langle X_i \rangle_{i=1}^{h{+}1}$
such that $X_i \subseteq U_i \setminus U_{i{-}1}$, and
\begin{itemize}
\item $S_0 = \bigcup^{h{+}1}_{i=1} X_i$
\item $X_1$ is an answer set for 
\begin{equation}
\pi_1 = b_{U_1}(\pi_0) \label{def_pi_1}
\end{equation}
\item for every $1 < i \le h+1$, $X_{i}$ is an answer set for
\begin{equation}
\pi_{i} = e_{U_{i}}(b_{U_{i}}(\pi_0) \setminus b_{U_{i{-}1}}(\pi_0), 
\bigcup_{1 \le t \le i{-}1}X_t) \label{def_pi_i}
\end{equation}
\end{itemize}
Given a set of atoms $\Sigma$, consider rules of the following forms:
\begin{eqnarray}
holds(l,1,1) & \la &  \label{r1_1} \\ 
  & & ( \initially(l) \in \cali ) \nonumber \\
poss(a,t,p) & \la & holds(\psi,t,p) \label{r1_2}\\
  & & (\executable(a,\psi) \in \cald) \nonumber\\
e(l,t,p) & \la & occ(a,t,p), holds(\phi,t,p) \label{r1_3}\\
  & & (\causes(a,l,\phi) \in  \cald) \nonumber\\
pc(l,t,p) & \la & occ(a,t,p), \naf holds(l,t,p), \nonumber \\
  & & \naf holds(\neg \phi,t,p) \label{r1_4}\\
  & & (\causes(a,l,\phi) \in  \cald)  \nonumber\\
br(g,t,k,p) \mid \dots & & \nonumber\\
br(g,t,k,w) & \la & occ(a,t,p) \label{r1_6}\\
  & & ( \determines(a,\theta) \in \cald , g \in \theta) \nonumber\\
pc(l,t,p) & \la & \naf holds(l,t,p), pc(l',t,p), 
  \naf e(\neg \varphi,t,p) \label{r1_8}\\
  & & (\caused(l,\varphi) \in \cald, l' \in \varphi) \nonumber\\
e(l,t,p) & \la & e(\varphi,t,p) \label{r1_9}\\
  & & (\caused(l,\varphi) \in \cald) \nonumber\\
holds(l,t,p) & \la & holds(\varphi,t,p) \label{r1_10}\\
  & & (\caused(l,\varphi) \in \cald) \nonumber\\
goal(t,p) & \la & holds(\calg,t,p) \label{r1_11}\\
goal(t,p) & \la & holds(f,t,p), holds(\neg f,t,p) \label{r1_11_2} \\
holds(l,t,p) & \la &  \label{r1_13}\\
  & & (e(l,t{-}1,p) \in \Sigma) \nonumber\\
holds(l,t,p) & \la & \label{r1_14}\\
  & & (holds(l,t{-}1,p) \in \Sigma , 
  pc(\myneg{l},t{-}1,p) \not \in \Sigma) \nonumber \\
used(t,p) & \la & \label{r1_19}\\
  & & (\exists \langle g,p' \rangle. p' < p \wedge
      br(g,t{-}1,p',p) \in \Sigma) \nonumber\\
holds(g,t,p) & \la & \label{r1_20}\\
  & & (\exists \langle g,p' \rangle. p' \le p \wedge
      br(g,t{-}1,p',p) \in \Sigma) \nonumber\\
holds(l,t,p) & \la & \label{r1_21}\\
  & & \exists \langle g,p' \rangle. p' < p \wedge
      br(g,t{-}1,p',p) \in \Sigma \wedge \nonumber \\
  & & holds(l,t{-}1,p') \in \Sigma) \nonumber\\
occ(a_1,t,p) \mid \dots \nonumber\\
  \mid occ(a_m,t,p) & \la & used(t,p), \naf goal(t,p)\label{r1_22}\\
used(1,1) & \la &  \label{r1_29} \\
used(t,p) & \la & \nonumber \label{r1_30}\\
  & & (used(t{-}1,p) \in \Sigma)
\end{eqnarray}

Then for each $i \in \{1,\dots,h+1\}$, 
$\pi_i$ can be divided into $w$ disjoint subprograms 
$\pi_i^k$, $1 \le k \le w$, where $\pi_i^k$ is defined as follows
\begin{equation}
\pi_i^k =\left \{
  \begin{array}{ll}
    \{ (\ref{r1_1})-(\ref{r1_11_2}), (\ref{r1_22})-(\ref{r1_29}) \mid t=1, p=1\}
       & \textnormal{if } i=1, k=1 \\
    \{ (\ref{r1_2})-(\ref{r1_11_2}), (\ref{r1_22}) \mid t=1, p=k\} 
       & \textnormal{if } i=1, k > 1 \\
    \{ (\ref{r1_2})-(\ref{r1_22}), (\ref{r1_30}) \mid t=i, p=k, \Sigma = X_{i-1}\} 
       & \textnormal{if } 1 < i \le h \\
    \{ (\ref{r1_10})-(\ref{r1_21}), (\ref{r1_30}) \mid t=h+1, p=k, \Sigma = X_{h}\} & \textnormal{otherwise}
  \end{array}
\right.
\label{def_pi_i_k}
\end{equation}
Let $X_{i,k}$ denote $X_i \cap A_{i,k}$. From Lemma \ref{l3}, it 
follows that $X_{i,k}$ is an answer set for $\pi_i^k$. Hence, we have
\begin{eqnarray*}
\delta_{i,k}(S) = \delta_{i,k}(S_0) = \delta_{i,k}(X_i) = \delta_{i,k}(X_{i,k})
\end{eqnarray*}
Due to this fact, from now on, we will use $\delta_{i,k}$ to refer
to either $\delta_{i,k}(S)$, $\delta_{i,k}(S_0)$, $\delta_{i,k}(X_i)$,
or $\delta_{i,k}(X_{i,k})$.

We have the following lemma
\begin{lemma}
\label{l4}
For $1 \le i \le h+1$ and $1 \le k \le w$,
\begin{itemize}
\item[1.] if $used(i,k) \not \in S$ then $S$ does not
contain any atoms of the forms $holds(l,i,k)$,
$e(l,i,k)$, $br(g,i,k,k)$;
\item[2.] if $used(h+1,k) \in S$ and $\delta_{h+1,k}$ is
consistent then 
$$\delta_{h+1,k} \models \calg.$$
\end{itemize}
\end{lemma}
\begin{proof}
\begin{itemize}
\item[1.] We will use induction on $i$ to prove this item.
\begin{itemize}
\item[a.] {\bf Base case:} $i=1$. Let $k$ be an integer such
that $used(1,k) \not \in S$. Clearly we have $k > 1$. 
On the other hand, it is easy to see that (by using the splitting set 
$Z = A_{1,k}^{\{holds,e,br,occ,goal,used\}}$) 
if $k > 1$ then $S$ does not contain atoms of the forms 
$holds(l,1,k)$, $e(l,1,k)$, and $br(g,1,k,k)$. Thus, the base case is true.
\item[b.] {\bf Inductive step:} Assume that Item 1 is true
for $i \le j{-}1$, where $j > 1$. We will prove that it is also true for 
$i = j$. Let $k$ be an integer such that $used(j,k) \not \in S$.

Clearly, to prove Item 1 we only need to prove that atoms of the forms
$e(l,j,k)$, $holds(l,j,k)$, $br(g,j,k,k)$ do not belong to $X_{j,k}$.
Consider the program $\pi_j^k$ (see (\ref{def_pi_i_k})). We know that
$X_{j,k}$ is an answer set for $\pi_j^k$.

Because of rule (\ref{r1_30}), we have 
$used(j{-}1,k) \not \in X_{j{-}1}$. From (\ref{r1_19}), 
it follows that $br(g,j{-}1,k',k) \not \in X_{j{-}1}$ for
every pair $\langle g,k' \rangle$ such that $k' < k$. 
In addition, by the inductive hypothesis, 
we have that for any $l$ and $g$, $e(l,j{-}1,k)$,
$holds(l,j{-}1,k)$, and $br(g,j{-}1,k,k)$ are not in $X_{j{-}1}$.
As a result, rules (\ref{r1_13})-(\ref{r1_21}) do not exist in 
$\pi_{j}^k$. If we split $\pi_j^k$
by the set $Z = A_{j,k}^{\{holds,e,br,occ,used,goal\}}$ then 
$b_Z(\pi_j^k)$ is the set of rules of the forms 
\begin{itemize}
\item[i.] (\ref{r1_3}), (\ref{r1_6}), (\ref{r1_9})--(\ref{r1_11}), 
(\ref{r1_22}) if $i \le h$ 
\item[ii.] (\ref{r1_10})--(\ref{r1_11}) if $i = h+1$
\end{itemize}
It is not difficult to
show that this program has the empty set as its only answer 
set (recall that $\calg \ne \emptyset$). From this, we can 
conclude the inductive step.
\end{itemize}
\item[2.] It is obvious because of the rules (\ref{r0_11}),
(\ref{r0_11_2})
and the constraint (\ref{r0_12}).
\end{itemize}
\end{proof}

\begin{lemma}
\label{l51}
For $1 \le i \le h$ and $1 \le k \le w$, if 
$occ(a,i,k) \in S$ then $a$ is executable
in $\delta_{i,k}$ and there is no $b \ne a$ such that
$occ(b,i,k) \in S$.
\end{lemma}
\begin{proof} From constraint (\ref{r0_23}), it follows that
$poss(a,i,k) \in S$. Notice only rules of
the form (\ref{r0_2}) may have $poss(a,i,k)$ as its head.
Hence, there must be a proposition (\ref{exec})
in $\cald$ such that $\psi$ holds in $\delta_{i,k}$.
This means $a$ is executable in $\delta_{i,k}$.

If there exists $b \ne a$ such that $occ(b,i,k) \in S$ then
constraint (\ref{r0_22_2}) could not be satisfied.
\end{proof}

\begin{lemma}
\label{l52}
for $1 \le i \le h$ and $1 \le k \le w$ 
\begin{itemize}
\item[1.] if $occ(a,i,k) \in S$ and $a$ is a non-sensing action then
\begin{itemize}
\item[a.] $e(l,i,k) \in S$ iff $l \in  e(a,\delta_{i,k})$
\item[b.] $pc(l,i,k) \in S$ iff $l \in pc(a,\delta_{i,k})$
\item[c.] $\neg \exists \langle g,k'\rangle. br(g,i,k',k) \in S$
\end{itemize}
\item [2.] if $occ(a,i,k) \in S$ and $a$ is a sensing 
action $a$ with occurring in a k-proposition of the form 
(\ref{knowledge}) in $\cald$
and $\theta = \{ g_1, \dots, g_n \}$ then there exist $n$ distinct
integers $k_1,\dots,k_n$ greater than or equal to $k$ such that
\begin{itemize}
\item[a.] $X_{i,k}^{\{br\}} = \{ br(g_j,i,k,k_j) \mid j \in \{1,\dots, n\}\}$
\item[b.] $g_j$ does not hold in $\delta_{i,k}$,
\item[c.] if $k_j > k$ then $S$ does not contain any atoms of the form
$holds(l,i,k_j)$
\end{itemize}
\item[3.] if $occ(a,i,k) \not \in S$ for every action $a$ then
\begin{itemize}
\item[a.] $\forall l.pc(l,i,k) \not \in S \wedge e(l,i,k) \not \in S$
\item[b.] $\forall\langle g,k'\rangle.br(g,i,k,k') \not\in S$
\end{itemize}
\end{itemize} 
\end{lemma}
\begin{proof}
Let us split $\pi_i^k$ by the set 
$Z_1 = A_{i,k}^{\{used,goal,occ,holds,poss\}}$. 
By the splitting set theorem,
$X_{i,k} = M \cup N$ where $M$ is an answer set for 
$b_{Z_1}(\pi_i^k)$ and $N$ is an answer set 
for $\Pi_1 = e_{Z_1}(\pi_i^k \setminus b_{Z_1}(\pi_i^k),M)$,
which consists of the following rules
\begin{eqnarray}
e(l,i,k) & \la & \label{l5_r1_3}\\
  & & (occ(a,i,k) \in M, \causes(a,l,\phi) \in \cald, \nonumber\\
  & & holds(\phi,i,k) \subseteq M) \nonumber\\
pc(l,i,k) & \la & \label{l5_r1_4}\\
  & & (occ(a,i,k) \in M, holds(l,i,k) \not \in M, \nonumber \\
  & & \causes(a,l,\phi) \in  \cald, 
      holds(\neg \phi,i,k) \cap M = \emptyset)  \nonumber\\
br(g,i,k,k) \mid \dots & & \nonumber\\
br(g,i,k,w) & \la &  \label{l5_r1_6}\\
  & & (occ(a,i,k) \in M, 
      \determines(a,\theta) \in \cald, g \in \theta) \nonumber\\
pc(l,i,k) & \la & pc(l',i,k), 
  \naf e(\neg \varphi,i,k) \label{l5_r1_8}\\
  & & (\caused(l,\varphi) \in \cald, 
       holds(l,i,k) \not \in M, l' \in \varphi) \nonumber\\
e(l,i,k) & \la & e(\varphi,i,k) \label{l5_r1_9}\\
  & & (\caused(l,\varphi) \in \cald) \nonumber
\end{eqnarray}
    From the splitting set theorem, it follows that
$\delta_{i,k}(M) = \delta_{i,k}$
\begin{itemize}
\item[1.] Assume that $occ(a,i,k) \in S$ and $a$ is a non-sensing
action. By Lemma \ref{l51}, we know that there exists no sensing
action\footnote{Recall that the sets of non-sensing actions
and sensing actions are disjoint from each other. Hence, $a$ itself is
not a sensing action.} $b$ such that $occ(b,i,k) \in S$. This means
that rules of form (\ref{l5_r1_6}) does not exist. Therefore, 
$\Pi_1$ can be rewritten to
\begin{eqnarray*}
e(l,i,k) & \la & \label{l5_r2_3}\\
  & & (\causes(a,l,\phi) \in \cald, 
      holds(\phi,i,k) \subseteq M) \nonumber\\
pc(l,i,k) & \la & \label{l5_r2_4}\\
  & & (holds(l,i,k) \not \in M, \causes(a,l,\phi) \in  \cald, \nonumber \\
  & & holds(\neg \phi,i,k) \cap M = \emptyset)  \nonumber\\
pc(l,i,k) & \la & pc(l',i,k), 
  \naf e(\neg \varphi,i,k) \label{l5_r2_8}\\
  & & (\caused(l,\varphi) \in \cald, 
       holds(l,i,k) \not \in M, l' \in \varphi) \nonumber\\
e(l,i,k) & \la & e(\varphi,i,k) \label{l5_r2_9}\\
  & & (\caused(l,\varphi) \in \cald) \nonumber
\end{eqnarray*}
If we continue splitting the above program using 
$Z_2 = A_{i,k}^{\{e\}}$ then by Lemma \ref{l2},
the bottom part has the only answer set 
$$\{ e(l,i,k) \mid l \in e(a,\delta_{i,k})\}$$
and the evaluation of the top part has the only answer set
$$\{pc(l,i,k) \mid l \in pc(a,\delta_{i,k})\}$$
Due to the fact that $M$ does not contain any atoms of
the form $e(l,i,k)$ or $pc(l,i,k)$, we therefore can conclude
Items (a) and (b).

We now show that $\neg \exists \langle g, k' \rangle. br(g,i,k',k) \in S$.
Suppose otherwise, i.e., there exists $g$ and $k'$ such
that $br(g,i,k',k) \in S$. Notice that only rule (\ref{r0_22})
with $t = i$ and $p = k$ has $occ(a,i,k)$ in its head. 
Hence, its body must be satisfied by $S$. That implies
$used(i,k) \in S$.

On the other hand, since only rules of the form
(\ref{r0_6}) with $p = k'$ may have $br(g,i,k',k)$ in 
its head, there exists a sensing action $b$ such that 
$occ(b,i,k') \in S$ and in addition, $k' \le k$. 
As the sets of non-sensing actions and
sensing actions are disjoint from each other, we have $b \ne a$. From 
Lemma \ref{l51}, it follows that $k' < k$.

Accordingly, we have $used(i,k) \in S, br(g,i,k',k) \in S$ and 
$k' < k$. Constraint (\ref{r0_18}) with $t = i$, $p=k$, and 
$p_1 = k'$ is thus violated. Thus, Item (c) holds.

\item[2.]
Assume that $occ(a,i,k) \in S$ and $a$ is a sensing 
action occurring in a k-proposition of the form (\ref{knowledge}) 
in $\cald$ with$\theta = \{ g_1, \dots, g_n \}$.

In this case, since rules of the forms (\ref{l5_r1_3}) 
and (\ref{l5_r1_4}) do not exist, $\Pi_1$ is the 
following set of rules
\begin{eqnarray*}
br(g_1,i,k,k) \mid \dots & & \nonumber\\
br(g_1,i,k,w) & \la &  \label{l5_r3_6_1}\\
\dots & \dots & \dots \nonumber \\
br(g_n,i,k,k) \mid \dots & & \nonumber\\
br(g_n,i,k,w) & \la &  \label{l5_r3_6_n}\\
pc(l,i,k) & \la & pc(l',i,k), 
  \naf e(\neg \varphi,i,k) \label{l5_r3_8}\\
  & & (\caused(l,\varphi) \in \cald,holds(l,i,k)
       \not \in M, l' \in \varphi) \nonumber\\
e(l,i,k) & \la & e(\varphi,i,k) \label{l5_r3_9}\\
  & & (\caused(l,\varphi) \in \cald) \nonumber
\end{eqnarray*}
By further splitting the above program using the set 
$A_{i,k}^{\{e,pc\}}$, we will see that the bottom part 
has the empty set as its only answer set (recall that we
are assuming that the body of each static law of the form (\ref{static})
is not empty). Therefore, the answer set for the above 
program is also the answer set for the following program
and vice versa.
\begin{eqnarray*}
br(g_1,i,k,k) \mid \dots & & \nonumber\\
br(g_1,i,k,w) & \la &  \label{l5_r4_6}\\
\dots & \dots & \dots \nonumber \\
br(g_n,i,k,k) \mid \dots & & \nonumber\\
br(g_n,i,k,w) & \la &  \label{l5_r4_6_2}
\end{eqnarray*}
Thus, there exist $n$ integers $k_1$, $\dots$, $k_n$
greater than or equal to $k$ such that
$$N = \bigcup_{j=1}^n \{ br(g_j,i,k,k_j) \}$$
It is easy to see that $X_{i,k}^{\{br\}} = N^{\{br\}}$.
In addition, by constraints of the form (\ref{r0_17}), 
$k_j$'s must be distinct. Thus, Items (a) is true. 

Item (b) can be drawn from constraints of the form (\ref{r0_7}).

Assume $k_j > k$. Because of constraints of the form (\ref{r0_18}),
we have $used(i,k_j) \not \in S$. From Lemma \ref{l4}, it follows
that $S$ does not contain any atoms of the form $holds(l,i,k_j)$.
Item (c) is thus true.
\item[3.] $occ(a,i,k) \not \in S$ for every action $a$.
In this case, $\Pi_1$ is the following set of rules
\begin{eqnarray*}
pc(l,i,k) & \la & pc(l',i,k), 
  \naf e(\neg \varphi,i,k) \label{l5_r5_8}\\
  & & (\caused(l,\varphi) \in \cald, holds(l,i,k)
       \not \in M, l' \in \varphi) \nonumber\\
e(l,i,k) & \la & e(\varphi,i,k) \label{l5_r5_9}\\
  & & (\caused(l,\varphi) \in \cald) \nonumber
\end{eqnarray*}
which has an empty set as its only answer set. Items (a)--(b)
follow from this.
\end{itemize}
\end{proof}
The following lemma shows that $\pi_{h,w}(\calp)$ correctly 
implements the transition function $\Phi$.
\begin{lemma}
\label{l6}
For $1 \le i \le h$ and $1 \le k \le w$ 
\begin{itemize}
\item[1. ] if there exists a non-sensing action $a$ such
that $occ(a,i,k) \in S$ then
$$\Phi(a,\delta_{i,k}) = 
\left \{ 
\begin{array}{ll}
\emptyset & \textnormal{ if } \delta_{i+1,k} 
\textnormal{ is inconsistent } \\
\{ \delta_{i{+}1,k} \} & \textnormal{ otherwise}
\end{array}
\right.
;$$
\item[2.] if there exists a sensing action $a$ occurring in a
k-proposition of the form (\ref{knowledge}) in $\cald$ with  
$\theta = \{g_1,\dots,g_n\}$ such that $occ(a,i,k) \in S$ then
there exist $n$ integers $\{k_1, \dots, k_n\}$ such that
$$\Phi(a,\delta_{i,k}) = \{ \delta_{i{+}1,k_j} \mid 1 \le j \le n,
\delta_{i+1,k_j} \textnormal{ is consistent}\},$$
and for each $j$, $g_j$ holds in $\delta_{i+1,k_j}$;
\item[3.] if $occ(a,i,k) \not \in S$ for every action $a$,
$$\delta_{i{+}1,k} = \delta_{i,k}.$$
\end{itemize} 
\end{lemma}
\begin{proof}
\begin{itemize}
\item[1.] Assume that there exists a non-sensing action
$a$ such that $occ(a,i,k) \in X_{i}$. 

Observe that $Z_1 = A_{i+1,k}^{\{holds\}}$
is a splitting set for $\pi_{i{+}1}^k$. 
Hence, by the splitting set theorem, 
$X_{i{+}1,k} = M \cup N$, where $M \subseteq Z_1$ is an 
answer set for $\Pi_1 = b_{Z_1}(\pi_{i{+}1}^k)$ and 
$N$ is an answer set for $\Pi_2 = e_{Z_1}(\pi_{i{+}1}^k \setminus \Pi_1,M)$. 

Notice that by Lemma \ref{l52}, rules (\ref{r1_20})--(\ref{r1_21})
for $t=i+1$, $p= k$ do not exist. Thus, $\Pi_1$ is the 
following set of rules:
\begin{eqnarray*}
holds(l,i{+}1,k) & \la & holds(\varphi,i{+}1,k) \label{l5_r7_10}\\
  & & (\caused(l,\varphi) \in \cald) \nonumber\\
holds(l,i{+}1,k) & \la &  \label{l5_r7_13}\\
  & & (e(l,i,k) \in X_{i}) \nonumber\\
holds(l,i{+}1,k) & \la & \label{l5_r7_14}\\
  & & (holds(l,i,k) \in X_{i} , 
  pc(\myneg{l},i,k) \not \in X_{i}) \nonumber
\end{eqnarray*}
Also by Lemma \ref{l52}, the conditions for the second and third
rules can be written as $(l \in e(a,\delta_{i,k}))$ and
$(l \in \delta_{i,k} , \myneg{l} \not \in 
pc(a,\delta_{i,k}))$ respectively.
Thus, by Lemma \ref{l2}, $\Pi_1$ has the unique answer set
$$M = \{ holds(l,i{+}1,k) \mid l \in Cl_\cald(a,\delta_{i,k})\}$$
On the other hand, by Lemma \ref{l51}, $a$ is executable in 
$\delta_{i,k}$. From the definition of the 
$Res_{\cald}$ and $\Phi$ functions, it follows that
$$\Phi(a,\delta_{i,k}) = 
\left \{ 
\begin{array}{ll}
\emptyset & \textnormal{ if } \delta_{i+1,k} 
\textnormal{ is inconsistent } \\
\{ \delta_{i{+}1,k} \} & \textnormal{ otherwise}
\end{array}
\right.$$
\item[2.] Assume that there exists a sensing action $a$
with a k-proposition of the form 
(\ref{knowledge}) and $\theta = \{ g_1,\dots,g_n\}$
such that $occ(a,i,k) \in S$.

By Lemma \ref{l52}, for each $j \in \{1\dots n \}$, 
there exists $k_j \ge k$ such that $br(g_j,i,k,k_j) \in X_i$. 
It is easy to see that $Z_2 = A_{i+1,k_j}^{\{holds\}}$ is a 
splitting set for $\pi_{i{+}1}^{k_j}$. 
Considering cases $k_j = k$ and $k_j > k$ in turn and
observe that $holds(l,i,k_j) \not \in S$ if $k_j > k$, 
we will see that in both cases $b_{Z_2}(\pi_{i{+}1}^{k_j})$ 
is the following set of rules:
\begin{eqnarray*}
holds(l,i{+}1,k_j) & \la & holds(\varphi,i{+}1,k_j) \label{r8_10}\\
  & & (\caused(l,\varphi) \in \cald) \nonumber\\
holds(l,i{+}1,k_j) & \la & \label{r8_14}\\
  & & (holds(l,i,k) \in X_{i}) \nonumber \\
holds(g_j,i{+}1,k_j) & \la & \label{r8_20}
\end{eqnarray*}
By Lemma \ref{l2}, the only answer set for the above program
is 
$$M = \{ holds(l,i+1,k_j) \mid l \in Cl_\cald(\delta_{i,k} \cup \{g_j\})\}$$
On the other hand, by Lemma \ref{l51}, $a$ is executable in $\delta_{i,k}$
and by Lemma \ref{l52}, $g_j$ does not hold in $\delta_{i,k}$.
Thus, according to the definition of the transition function, we have 
$$\Phi(a,\delta_{i,k}) =  \{ Cl_{\cald}(\delta_{i,k} \cup \{ g_j \}) \mid
1 \le j \le n, Cl_{\cald}(\delta_{i,k} \cup \{ g_j \}) 
\textnormal{ is consistent}\}$$
Hence, we have
$$\Phi(a,\delta_{i,k}) = \{ \delta_{i{+}1,k_j} (M)
\mid 1 \le j \le n, \delta_{i{+}1,k_j} (M) \textnormal{ is consistent}\} = $$
$$\{ \delta_{i{+}1,k_j} \mid 1 \le j \le n,
\delta_{i+1,k_j} \textnormal{ is consistent}\}$$
and obviously, $g_j$ holds in $\delta_{i+1,k_j}$.
\item[3.] Assume that $occ(a,i,k) \not \in S$ for every action $a$.

Similar to the first case, $Z_1$ is a splitting set for $\pi_{i{+}1}^k$. 
$b_{Z_1}(\pi_{i{+}1}^k)$ is the following set of rules:
\begin{eqnarray*}
holds(l,i{+}1,k) & \la & holds(\varphi,i{+}1,k) \label{r9_10}\\
  & & (\caused(l,\varphi) \in \cald) \nonumber\\
holds(l,i{+}1,k) & \la & \label{r9_14}\\
  & & (holds(l,i,k) \in X_{i}) \nonumber
\end{eqnarray*}

Because that $\delta_{i,k}$ is an a-state (Lemma \ref{l4}), by 
Lemma \ref{l2} the only answer set for this program is
$$M = \{ holds(l,i+1,k) \mid l \in \delta_{i,k}\}$$
Thus, we have
$$\delta_{i{+}1,k} = \delta_{i{+}1,k}(M) = 
\delta_{i,k}$$
\end{itemize}
\end{proof}
The following lemma shows that $\pi_{h,w}(\calp)$ correctly implements
the extended transition function.
\begin{lemma}
\label{l7}
We have
\begin{list}{}{}
\item[1. ] $\delta_{1,1}$ is the initial a-state for $\cal P$.
\item[2. ] For every pair of integers $1 \le i \le h {+} 1$,
$1 \le k \le w$, if $used(i,k) \in S$ then 
\begin{list}{}{}
\item[a) ] $p_i^k(S)$ is a conditional plan
\item[b) ] furthermore, if $\delta_{i,k}$ is consistent then
for every $\delta \in \hat \Phi(p_i^k(S),\delta_{i,k})$,
$\delta \models \calg$.
\end{list}
\end{list}
\end{lemma}
\begin{proof}
\begin{itemize}
\item[1. ]
$Z_1 = A_{1,1}^{\{holds\}}$ is a splitting set for $\pi_1^1$.
The bottom part, $b_{Z_1}(\pi_1^1)$, consists of the following rules:
\begin{eqnarray*}
holds(l,1,1) & \la & \\ 
& & \{ \initially(l) \in {\cal I} \}\\
holds(l,1,1) & \la & holds(\varphi,1,1) \\
& & \{ \caused(l,\varphi) \in {\cal D} \}
\end{eqnarray*}
By Lemma \ref{l2}, the only answer set for the above program is
$$M = \{holds(l,1,1) \mid l \in \delta_1 \}$$
where $\delta_1$ is the initial a-state of $\calp$. Thus,
$\delta_{1,1} = \delta_{1,1}(M)$ is the initial a-state of $\cal P$.
\item[2.]
We now prove Item 2 by induction on parameter $i$.
\begin{itemize}
\item[a. ] {\bf Base case:} $i = h {+}1$. Let $k$ be an arbitrary
integer between 1 and $w$ such that $used(i,k) \in S$.
Clearly $p_i^k(S)= []$ is a conditional plan. 

Now suppose that $\delta_{i,k}$ is consistent. 
According to the definition of the extended transition function, 
we have $$\hat \Phi(p^k_{i}(S),\delta_{i,k}) = \hat 
\Phi([],\delta_{i,k}) = \{ \delta_{i,k} \}$$
On the other hand, by Lemma \ref{l4}, we have that 
$\delta_{i,k} \models \calg$. 
Thus, Item 2 is true for $i = h{+}1$.

\item[b. ] {\bf Inductive step:} Assume that Item 2 is true for all 
$h+1 \ge i > t$. We will show that it is true for $i = t$. 
Let $k$ be an integer between 1 and $w$ such that $used(t,k) \in S$.
Consider three possibilities:
\begin{list}{}{}
\item[i. ] $occ(a,t,k) \in S$ for some non-sensing action $a$.
By the definition of $p^k_t(S)$, we have $p^k_t(S) = [a;p^k_{t{+}1}(S)]$.
In addition, by rule (\ref{r0_30}) we have $used(t{+}1,k)$ $\in S$. 
Thus, according to the inductive hypothesis,
$p^k_{t{+}1}(S)$ is a conditional plan. Accordingly, $p^k_t(S)$ is also 
a conditional plan. 

Now suppose that $\delta_{t,k}$ is consistent. Consider two cases
\begin{itemize}
\item {\em $\delta_{t+1,k}$ is consistent.} 
We have 
$$\hat \Phi(p^k_t(S),\delta_{t,k}) = 
\hat \Phi([a;p^k_{t{+}1}(S)], \delta_{t,k})
= \hat \Phi(p^k_{t{+}1}(S),\delta_{t{+}1,k})$$
(by Lemma \ref{l6} and by the definition of the 
extended transition function).

On the other hand, according to the inductive hypothesis, 
for every $\delta$ in $\hat \Phi(p^k_{t{+}1}(S),\delta_{t{+}1,k})$,
$\delta \models \calg$. Hence, the inductive step is proven.
\item {\em $\delta_{t+1,k}$ is inconsistent.} 
By Lemma \ref{l6}, we have $\hat \Phi(p^k_{t}(S),\delta_{t,k})
= \emptyset$. Thus, the inductive step is proven.
\end{itemize}
\item[ii. ] $occ(a,t,k) \in S$ for some sensing action $a$
with a k-proposition of the form 
(\ref{knowledge}) and $\theta = \{g_1,\dots,g_n\}$.
By Lemma \ref{l52} there exist exactly $n$ integers $k_1,\dots,k_n$
greater than $k$ such that $br(g_j,t,k,k_j) \in S$ for $1 \le j \le n$.
This implies that $used(t+1,k_j) \in S$ (see rules (\ref{r0_19}) and 
(\ref{r0_30})). Thus, by the definition of $p^k_t(S)$, we have 
$p^k_t(S) = [a;\kcases(\{g_j \rightarrow p_{i{+}1}^{k_j}(S) \}_{j=1}^n)]$.
On the other hand, we know by the inductive hypothesis that 
$p_{i{+}1}^{k_j}(S)$ is a conditional
plan for $1 \le j \le n$. As a result, $p^k_t(S)$ is also a conditional plan.

Suppose $\delta_{i,k}$ is consistent. Let $J = \{ j \mid
\delta_{t+1,k_j} \textnormal{ is consistent}\}$.
By Lemma \ref{l6}, we have
$$\Phi(a,\delta_{t,k}) = \{\delta_{t+1,k_j} \mid j \in J\}$$
and $g_j$ holds in $\delta_{t+1,k_j}$
for every $1 \le j \le n$. Hence, by the definition of $\hat\Phi$,
we have
$$\hat \Phi(p^k_{t}(S),\delta_{t,k}) = 
\bigcup_{j \in J} \hat \Phi(p^k_{t{+}1}(S),\delta_{t{+}1,k_j})$$
According to the inductive hypothesis, for every $\delta \in
\hat \Phi(p^k_{t{+}1}(S),\delta_{t{+}1,k_j})$, where $j \in J$,
we have $\delta \models \calg$. This implies
that for every $\delta \in \hat \Phi(p^k_{t}(S),\delta_{t,k})$,
we have $\delta \models \calg$.
\item[iii.] There is no action $a$ such that $occ(a,t,k) \in S$.
According to the definition of $p_t^k(S)$, $p_t^k(S) = []$. Hence,
it is a conditional plan.

It is easy to see that $goal(t,k) \in S$, 
which means that either $\delta_{t,k}$ is inconsistent
or $\delta_{t,k} \models \calg$ (see rules
(\ref{r0_11}), (\ref{r0_11_2}), and (\ref{r0_22})). 
Now suppose that $\delta_{t,k}$ is consistent. This implies
that $\delta_{t,k} \models \calg$. We have
$$\hat \Phi(p^k_{t}(S),\delta_{t,k})= \hat \Phi([],\delta_{t,k})
= \{ \delta_{t,k}\}$$
Thus, the inductive step is proven.
\end{list}
\end{itemize}
\end{itemize}
\end{proof}
Theorem \ref{t1} immediately follows from Lemma \ref{l7}.
\subsection*{Proof of Proposition \ref{prop-reduct}}
First, we prove the following lemma.
\begin{lemma}
\label{lm-reduct}
Let $\calp=(\cald,\cali,\calg)$ be a planning problem instance,
$\delta$ be an a-state and $p$ be a plan.
If $\hat \Phi(p,\delta) \models \calg$ then
$\hat \Phi(\reduct_\delta(p),\delta) \models \calg$.
\end{lemma}
\begin{proof}
Let us prove the lemma by structural induction on $p$.
\begin{itemize}
\item[1.] {\em $p = []$}. 

The proof is trivial since $\reduct_\delta(p) = p  = []$.
\item[2.] {Assume that \em $p = [a;q]$, where $q$ is a conditional plan and
$a$ is a non-sensing action and the lemma is true for $q$}. 

Suppose $\hat \Phi(p,\delta) \models \calg$. We need to show that
$\hat \Phi(\reduct_\delta(p),\delta) \models \calg$.

If $\delta \models \calg$ then
$$\hat \Phi(\reduct_\delta(p),\delta) = \hat
\Phi([],\delta) = \{ \delta \} \models \calg$$

Now consider the case that $\delta \not \models \calg$.

Clearly, we have $\Phi(a,\delta) \ne \bot$. Therefore,
$\Phi(a,\delta) = \{ \delta' \}$ for some $\delta'$.
Hence, by the definition of $\reduct$, we have
$$\reduct_\delta(p) = a;\reduct_{\delta'}(q)$$
Thus,
$$\hat \Phi(\reduct_\delta(p),\delta) = \hat
\Phi(\reduct_{\delta'}(q),\delta')$$

On the other hand, we have
$$\hat \Phi(p,\delta) = \hat \Phi(q,\delta')$$

Because $\hat \Phi(p,\delta) \models \calg$,
we have 
$$\hat \Phi(q,\delta') \models \calg$$

By inductive hypothesis, we have
$$\hat \Phi(\reduct_{\delta'}(q),\delta') \models \calg$$

Hence, 
$$\hat \Phi(\reduct_\delta(p),\delta) \models \calg$$

\item[3.] {Assume that \em $p = 
[a;\kcases(\{g_j \rightarrow p_j\}_{j=1}^n)]$, where
a is a sensing action that senses $g_1,\dots,g_n$,
and the lemma for $p_j$'s}. 

Suppose $\hat \Phi(p,\delta) \models \calg$. We need to show that
$\hat \Phi(\reduct_\delta(p),\delta) \models \calg$.

If $\delta \models \calg$ then
$$\hat \Phi(\reduct_\delta(p),\delta) = \hat
\Phi([],\delta) = \{ \delta \} \models \calg$$

Now consider the case that $\delta \not \models \calg$.
There are two possibilities.
\begin{itemize}
\item[a)] {\em there exists $g_k$ such that $g_k \in \delta$}.
By the definition of $\reduct$, we have
$$\reduct_{\delta}(p) = \reduct_{\delta}(p_k)$$
By the definition of the $\hat \Phi$-function, it is easy to
see that
$$\hat \Phi(p,\delta) = \hat \Phi(p_k,\delta)$$
Since $\hat \Phi(p,\delta) \models \calg$, we have
$\hat \Phi(p_k,\delta) \models \calg$. By the inductive 
hypothesis, we have
$$\hat \Phi(\reduct_{\delta}(p_k),\delta) \models \calg$$
Hence, we have
$$\hat \Phi(\reduct_{\delta}(p),\delta) \models \calg$$

\item[a)] {\em for every $1 \le j \le n$, $g_j \not \in \delta$}. 
By the definition of $\reduct$, we have
$$\reduct_{\delta}(p) = a;\kcases(\{g_j \rightarrow q_j\}_{j=1}^{n})$$
where
$$q_j = 
\left \{
  \begin{array}{ll}
    [] & \textnormal{ if } Cl_{\cald}(\delta \cup \{g_j\})
    \textnormal{ is inconsistent} \\
    \reduct_{Cl_{\cald}(\delta \cup \{g_j\})} (p_j) 
    & \textnormal{ otherwise}
  \end{array}
\right.
$$
For every $1 \le j \le n$, let $\delta_j = Cl_{\cald}(\delta \cup \{g_j\})$.
Let $J = \{j \mid \delta_j \textnormal{ is consistent} \}$. 
It is easy to see that
$$\hat \Phi(p,\delta) = \bigcup_{j \in J} \Phi(p_j,\delta_j)$$
because $g_j$ holds in $\delta_j$ but for every $k \ne j$,
$g_k$ does not hold in $\delta_j$.

Because $\hat \Phi(p,\delta) \models \calg$, we have
$$\Phi(p_j,\delta_j) \models \calg$$ for every $j \in J$.

On the other hand, we have
$$q_j = 
\left \{
  \begin{array}{ll}
    [] & \textnormal{ if } j \not \in J \\
    \reduct_{\delta_j} (p_j) 
    & \textnormal{ otherwise}
  \end{array}
\right.
$$
Thus,
$$\hat \Phi(\reduct_{\delta}(p),\delta) =
\bigcup_{j \in J} \Phi(q_j,\delta_j) =
\bigcup_{j \in J} \Phi(\reduct_{\delta_j}(p_j),\delta_j)
$$
By the inductive hypothesis, for every $j \in J$, 
as $\Phi(p_j,\delta_j) \models \calg$, we have
$\Phi(\reduct_{\delta_j}(p_j),\delta_j) \models \calg$.
As a result, we have 
$$\hat \Phi(\reduct_{\delta}(p),\delta) \models \calg$$
\end{itemize}
\end{itemize}
\end{proof}

We now prove Proposition \ref{prop-reduct}. Let $p$
be a solution to $\calp$. From the construction of
$\reduct$, it is easy to see that $\reduct_\delta(p)$
is unique.

By Lemma \ref{lm-reduct},
we have that $\hat \Phi(\reduct_\delta(p),\delta) \models
\calg$. Thus, $\reduct_\delta(p)$ is also a solution
to $\calp$.

So, we can conclude the proposition.
\subsection*{Proof of Theorem \ref{t2}}
\label{proof2}
The idea of the proof is as follows. Let $q$ be 
$\reduct_{\delta}(p)$, where $\delta$ is the initial a-state
of $\calp$, and let $T_q$ be the labeled tree 
for $q$ numbered according to the principles described 
in Section \ref{planner}.
Let $h$ and $w$ denote the height and width of $T_q$ respectively.
For $1 \le i \le h+1$, $1 \le k \le w$, we define $\delta_{i,k}$ 
to be the a-state at node $(i,k)$\footnote{That is, the node
numbered with $(i,k)$ in $T_q$} of $T_q$ if such a node
exists and $\bot$ otherwise. Based on $T_q$ and 
$\delta_{i,k}$, we construct the set $Y_{i,k}$ 
of atoms that hold at node $(i,k)$. Then we prove that the 
union of these sets, denoted by $S_0'$, 
is an answer set for $\pi_0$ (rules (\ref{r0_1})-(\ref{r0_30})) 
by showing that each $Y_{i,k}$ is an answer
set for a part of $\pi_0$, denoted by ${\pi'}_{i}^{k}$. Furthermore, 
a set $S'$ can be constructed from
$S_0'$ in such a way that it is an answer set for $\pi^*_{h,w}(\calp)$. 
Moreover, $S'$ does not violate any constraints in $\pi^c_{h,w}$
(rules (\ref{r0_5})-(\ref{r0_23})).
As such, it is an answer set for $\pi_{h,w}(\calp)$. Moreover,
$q=p^1_1(S')$. 

Given the numbered tree $T_q$, by $\langle a, i, k \rangle$ 
we mean the node labeled with $a$ and numbered with $(i,k)$
in $T_q$; by $\langle g,i,k,k' \rangle \in T_q$ we mean the
link, whose label is $g$, between the nodes $(i,k)$ and $(i+1,k')$ 
in $T_q$. 

For $1 \le i \le h+1$, $1 \le k \le w$, 
we define the a-state $\delta_{i,k}$ as follows.
\begin{itemize}
\item[i.] if $i=1$ 
\begin{equation}
\delta_{i,k} = \left \{
\begin{array}{lll}
Cl_\cald(\{l \mid \initially(l) \in \cali\}) &
  \hspace{.05in} & \textnormal{if } k=1\\
\bot &
  \hspace{.05in} & \textnormal{if } k > 1
\end{array}
\right.\label{delta_1}
\end{equation}
\item[ii.] if $i > 1$
\begin{equation}
\delta_{i,k} = \left \{
\begin{array}{lll}
Cl_\cald(e(a,\delta_{i{-}1,k}) \cup
(\delta_{i{-}1,k} \setminus pc(a,\delta_{i{-1},k}))) &
  \hspace{.05in} & \textnormal{if } \langle a,i{-}1,k \rangle \in T_q
  \textnormal{ for } \\
  \hspace{1in} \textnormal{a non-sensing action } a &
  \\
Cl_\cald(\delta_{i{-}1,k'} \cup \{g\}) &
  \hspace{.05in} & \textnormal{if } \langle g,i{-}1,k',k \rangle \in T_q \\
\delta_{i{-}1,k}  & \hspace{.05in} & \textnormal{otherwise}
\end{array}
\right. \label{delta_2}
\end{equation}
\end{itemize}
Note that given $(i,k)$, there exists at most one action $a$
such that $\langle a,i{-}1,k \rangle \in T_q$, and furthermore,
at most one pair $\langle g,k' \rangle$ such that 
$\langle g,i{-}1,k',k \rangle \in T_q$. In addition, the
conditions in Equation (\ref{delta_2}) do not overlap each other. 
Thus, $\delta_{i,k}$ is uniquely defined for $1 \le i \le h+1$
and $1 \le k \le w$. In what follows, the undefined situation 
$\bot$ can sometimes be thought of as $\emptyset$, depending the
context in which it is used.

Let us construct the set $Y_{i,k}$ of atoms based on $\delta_{i,k}$
as follows.
\begin{enumerate}
\item $used(1,1) \in Y_{1,1}$
\item $holds(l,i,k) \in Y_{i,k}$ iif $l \in \delta_{i,k}$
\item $poss(a,i,k) \in Y_{i,k}$ iif there exists a proposition of the form
(\ref{exec}) s.t. $\psi \subseteq \delta_{i,k}$
\item $occ(a,i,k) \in Y_{i,k}$ iif $\langle a,i,k \rangle \in T_q$
\item $br(g,i,k,k') \in Y_{i,k}$ iif $\langle g,i,k,k' \rangle \in T_q$
for some $g,k'$
\item $e(l,i,k) \in Y_{i,k}$ iif $\langle a,i,k \rangle \in T_q$ and 
$l \in e(a,\delta_{i,k})$ for some non-sensing action $a$
\item $pc(l,i,k) \in Y_{i,k}$ iif $\langle a,i,k \rangle \in T_q$ and 
$l \in pc(a,\delta_{i,k})$ for some non-sensing action $a$
\item For $i > 1$, $used(i,k) \in Y_{i,k}$ iif either
\begin{enumerate}
\item $used(i{-}1,k) \in Y_{i{-}1,k}$; or
\item there exists $\langle g,k' \rangle$ s.t. $\langle g,i{-}1,k',k \rangle 
\in Y_{i{-}1,k'}$
\end{enumerate}
\item $goal(i,k) \in Y_{i,k}$ iff $\delta_{i,k} \models \calg$
or $\delta_{i,k}$ is inconsistent
\item Nothing else in $Y_{i,k}$
\end{enumerate}
Clearly, $Y_{i,k}$'s are uniquely defined. Furthermore, they are 
disjoint from each other. Let 
$$Y_i = \bigcup_{k=1}^w Y_{i,k} \textnormal{ and }
S_0' = \bigcup_{i=1}^{h+1} Y_i$$

\begin{lemma}
\label{lmb1}
For $1 \le i \le h$ and $1 \le k \le w$, let 
$M = Y_{i,k}^{\{holds,poss,goal,used,occ\}}$
and let $\Pi$ be the following program:
\begin{eqnarray*}
e(l,i,k) & \la & \label{lmb1_r0_3}\\
  & & (occ(a,i,k) \in M, \causes(a,l,\phi) \in \cald,
      holds(\phi,i,k) \subseteq M) \nonumber\\
pc(l,i,k) & \la & \label{lmb1_r0_4}\\
  & & (occ(a,i,k) \in M, \causes(a,l,\phi) \in  \cald, \nonumber \\
  & & holds(l,i,k) \not \in M, 
      holds(\neg \phi,i,k) \cap M = \emptyset)  \nonumber\\
br(g,i,k,k) \mid \dots & & \nonumber\\
br(g,i,k,w) & \la &  \label{lmb1_r0_6}\\
  & & (occ(a,i,k) \in M, 
      \determines(a,\theta) \in \cald, g \in \theta) \nonumber\\
pc(l,i,k) & \la & pc(l',i,k), 
  \naf e(\neg \varphi,i,k) \label{lmb1_r0_8}\\
  & & (\caused(l,\varphi) \in \cald, holds(l,i,k) \not \in M, 
  l' \in \varphi) \nonumber\\
e(l,i,k) & \la & e(\varphi,i,k) \label{lmb1_r0_9}\\
  & & (\caused(l,\varphi) \in \cald) \nonumber
\end{eqnarray*}
Then, $N = Y_{i,k}^{\{e,pc,br\}}$ is an answer set for $\Pi$.
\end{lemma}
\begin{proof}
Given $(i,k)$, there are three cases that may happen at node $(i,k)$.
\begin{itemize}
\item there exists a non-sensing action $a$ such that
$\langle a,i,k \rangle \in T_q$;
\item there exists a sensing action $a$ such that 
$\langle a,i,k \rangle \in T_q$;
\item $\langle a,i,k \rangle \not \in T_q$ for every action $a$
\end{itemize}
Let us consider each of those in turn.
\begin{itemize}
\item[1.] {\em there exists a non-sensing action $a$ such that
$\langle a,i,k \rangle \in T_q$.}

        From the construction of $Y_{i,k}$, 
we know that $occ(a,i,k) \in M$
and there is no $b \ne a$ such that $occ(b,i,k) \in M$. 
Furthermore, due to the fact that $N$ does not contain any
atom of the form $holds(l,i,k)$, we have $holds(l,i,k) \in M$ iff 
$holds(l,i,k) \in Y_{i,k}$. That means $holds(l,i,k) \in M$ iff
$l \in \delta_{i,k}$.

Hence, $\Pi$ can be rewritten to:
\begin{eqnarray*}
e(l,i,k) & \la & \label{lmb1_r5_3}\\
  & & (l \in e(a,\delta_{i,k})) \nonumber\\
pc(l,i,k) & \la & \label{lmb1_r5_4}\\
  & & (l \in pc^0(a,\delta_{i,k}))  \nonumber\\
pc(l,i,k) & \la & pc(l',i,k), 
  \naf e(\neg \varphi,i,k) \label{lmb1_r5_8}\\
  & & (\caused(l,\varphi) \in \cald, l \not \in \delta_{i,k},
  l' \in \varphi) \nonumber\\
e(l,i,k) & \la & e(\varphi,i,k) \label{lmb1_r5_9}\\
  & & (\caused(l,\varphi) \in \cald) \nonumber
\end{eqnarray*}
As have been seen in the proof of Theorem \ref{t1} 
(see the proof of Lemma \ref{l52}, Item 1), 
the only answer set for this program
is $\{ e(l,i,k) \mid l \in e(a,\delta_{i,k}) \} \cup 
\{ pc(l,i,k) \mid l \in pc(a,\delta_{i,k}) \}= N$.
\item[2. ] {\em there exists a sensing action $a$ such that
$\langle a,i,k \rangle \in T_q$.}

We have $occ(a,i,k) \in M$ and there is no non-sensing action $b$
such that $occ(b,i,k) \in M$. As a result, $\Pi$ is 
\begin{eqnarray*}
br(g,i,k,k) \mid \dots & & \nonumber\\
br(g,i,k,w) & \la &  \label{lmb1_r10_6}\\
  & & (occ(a,i,k) \in M, 
      \determines(a,\theta) \in \cald, g \in \theta) \nonumber\\
pc(l,i,k) & \la & pc(l',i,k), 
  \naf e(\neg \varphi,i,k) \label{lmb1_r10_8}\\
  & & (\caused(l,\varphi) \in \cald, holds(l,i,k) \not \in M, 
  l' \in \varphi) \nonumber\\
e(l,i,k) & \la & e(\varphi,i,k) \label{lmb1_r10_9}\\
  & & (\caused(l,\varphi) \in \cald) \nonumber
\end{eqnarray*}
It is easy that an answer set for $\Pi$ is also
an answer set for
\begin{eqnarray*}
br(g,i,k,k) \mid \dots & & \nonumber\\
br(g,i,k,w) & \la &  \label{lmb1_r15_6}\\
  & & (occ(a,i,k) \in M, 
      \determines(a,\theta) \in \cald, g \in \theta) \nonumber
\end{eqnarray*}
and vice versa. On the other hand, 
$$N = Y_{i,k}^{\{e,pc,br\}} = 
\{ br(g,i,k,k') \mid \langle g,i,k,k' \rangle \in T_q\}$$
is an answer set for the latter program. As a result, $N$ is
also an answer set for $\Pi$.
\item[3.] {\em $\langle a,i,k \rangle \not \in T_q$ 
for every action $a$.}

In this case, the first three rules of $\Pi$ do not exist 
because $occ(a,i,k) \not \in M$ for every $a$. Thus, $\Pi$
consists of the last two rules only. It is easy to see that
it has the empty set as its only answer set. On the other hand,
from the construction of $Y_{i,k}$, we have $Y_{i,k}^{\{e,pc,br\}}
= \emptyset$. Accordingly, $Y_{i,k}^{\{e,pc,br\}}$ is an answer
set for $\Pi$.
\end{itemize}
The proof is done.
\end{proof}

\begin{lemma}
\label{lmb2}
For $1 \le i \le h+1$, $1 \le k \le w$, $Y_{i,k}$ is an answer 
set for ${\pi'}^k_{i}$, where ${\pi'}^k_{i}$ is defined in the same 
way as ${\pi}^k_{i}$ except that we replace every occurrence of $X$ 
in Equation (\ref{def_pi_i_k}) by $Y$.
\end{lemma}
\begin{proof}
Let us consider in turn two cases $i = 1$ and $i > 1$.
\begin{itemize}
\item[1. ] $i = 1$.
It is easy to see that the only answer set for 
${\pi'}^k_{1}$, where $k > 1$, is 
$$
Y_{1,k} = \{poss(a,1,k) \mid \executable(a,\emptyset) \in \cald\} 
$$
by using the splitting set $A_{1,k}^{\{holds,occ,
br,used,e,pc\}}$ (see (\ref{eq2}) for the definition of $A_{i,k}$) 
and observe that the bottom part has the empty set as its 
only answer set and $Y_{1,k}$ is the only answer set for 
the evaluation of the top part. 

We now prove that $Y_{1,1}$ is an answer set for ${\pi'}^1_{1}$ which
consists of the rules of the forms 
(\ref{r1_1})-(\ref{r1_11_2}), (\ref{r1_22})-(\ref{r1_29}) where  $t=1$
and $p=1$.
If we use the set $Z_1 = A_{1,1}^{\{holds,occ,poss,goal,used\}}$ 
to split ${\pi'}^1_{1}$ then $b_{Z_1}({\pi'}^1_{1})$ is 
$$\{ (\ref{r1_1})-(\ref{r1_2}),(\ref{r1_10})-(\ref{r1_11_2}),(\ref{r1_22}),
(\ref{r1_29}) \mid t = 1, p = 1 \}$$

    From the definition of $Y_{1,1}$, we can easily show that 
$M = Y_{1,1}^{\{holds,occ,poss,goal,used\}}$
is an answer set for  $b_{Z_1}({\pi'}^1_{1})$. Furthermore, we have 
$$\delta_{1,1}(M) = \delta_{1,1}(Y_{1,1}) = \delta_{1,1}$$

The evaluation of the top part, 
$\Pi_1 = e_{Z_1}({\pi'}^1_1 \setminus b_{Z_1}({\pi'}^1_1),M)$, is 
the following set of rules
\begin{eqnarray*}
e(l,1,1) & \la & \label{lmb2_r3_3}\\
  & & (occ(a,1,1) \in M, \causes(a,l,\phi) \in  \cald, \nonumber\\
  & &  holds(\phi,1,1) \subseteq M) \nonumber\\
pc(l,1,1) & \la & \label{lmb2_r3_4}\\
  & & (occ(a,1,1) \in M, \causes(a,l,\phi) \in  \cald,\nonumber\\
  & & holds(l,1,1) \not \in M, holds(\neg \phi,1,1) 
      \cap M = \emptyset) \nonumber\\
br(g,1,1,k) \mid \dots & & \nonumber\\
br(g,1,1,w) & \la & \label{lmb2_r3_6}\\
  & & ( \determines(a,\theta) \in \cald, g \in \theta, 
  occ(a,1,1) \in M) \nonumber\\
pc(l,1,1) & \la & pc(l',1,1), \naf e(\neg \varphi,1,1) \label{lmb2_r3_8}\\
  & & (\caused(l,\varphi) \in \cald, l' \in \varphi,
  holds(l,1,1) \not \in M) \nonumber\\
e(l,1,1) & \la & e(\varphi,1,1) \label{lmb2_r3_9}\\
  & & (\caused(l,\varphi) \in \cald) \nonumber
\end{eqnarray*}
By Lemma \ref{lmb1}, $N = Y_{1,1}^{\{e,pc,br\}}$ is
an answer set for $\Pi_1$. As a result, $Y_{1,1} = M \cup N$ is an
answer set for ${\pi'}^1_1$.
\item[2. ] $1 < i \le h+1$. 

Using the splitting set $Z_2 = A_{i,k}^{\{holds,occ,goal,used,poss\}}$
to split ${\pi'}^k_i$, we have that
the bottom part $\Pi_2 = b_{Z_2}({\pi'}^k_i)$ consists of rules of the forms
\begin{list}{$\bullet$}{}
\item (\ref{r1_2}), (\ref{r1_10})--(\ref{r1_22}), and (\ref{r1_30}) if
$i \le h$
\item (\ref{r1_10})--(\ref{r1_21}), and (\ref{r1_30}) if $i = h+1$
\end{list}

We now prove that $M = Y_{i,k}^{\{holds,occ,goal,used,poss\}}$
is an answer set for $\Pi_2$. Let us further
split $\Pi_2$ by the set $Z_3 = A_{i,k}^{\{holds\}}$. 
Then, the bottom part $b_{Z_3}(\Pi_2)$ consists of
rules of the forms (\ref{r1_10}), (\ref{r1_13})--(\ref{r1_14}), 
(\ref{r1_20})--(\ref{r1_21}) only.

Consider three cases
\begin{itemize}
\item[a.] {\em there exists a non-sensing action $a$ such that
$occ(a,i{-}1,k) \in Y_{i{-}1}$.}

    From the construction of $Y_{i,k}$'s, it is easy to see that
there exists no $\langle g,k' \rangle$ such that $br(g,i{-}1,k',k) 
\in Y_{i{-}1}$. Thus, $b_{Z_3}(\Pi_2)$ contains rules of
the forms (\ref{r1_10}), (\ref{r1_13})--(\ref{r1_14}) only.
On the other hand, we have 
$$e(l,i{-}1,k) \in Y_{i{-}1} \textnormal{ iff } l \in e(a,\delta_{i{-}1,k})$$
$$pc(\myneg{l},i{-}1,k) \not \in Y_{i{-}1} \textnormal{ iff } \myneg{l} \not 
\in pc(a,\delta_{i{-}1,k})$$
Hence, $b_{Z_3}(\Pi_2)$ is the following collection of rules:
\begin{eqnarray*}
holds(l,i,k) & \la & holds(\varphi,i,k) \\
  & & (\caused(l,\varphi) \in \cald) \nonumber\\
holds(l,i,k) & \la & \\
  & & (l \in e(a,\delta_{i{-}1,k})) \nonumber\\
holds(l,i,k) & \la & \\
  & & (l \in \delta_{i{-}1,k} ,
  \myneg{l} \not \in \delta_{i{-}1,k}) \nonumber
\end{eqnarray*}
By Lemma \ref{l2}, it has the only answer set
$$\{ holds(l,i,k) \mid l \in Cl_\cald(e(a,\delta_{i{-}1,k}) \cup
(\delta_{i{-}1,k} \setminus pc(a,\delta_{i{-1},k})))\} =
Y_{i,k}^{\{holds\}}$$
\item[b.] $\exists \langle g,k' \rangle. br(g,i{-}1,k',k) \in Y_{i{-}1}$.

        From the construction of $Y_{i,k}'s$, such 
$\langle g,k' \rangle$ is unique and in addition $k' \le k$. 
Thus, $b_{Z_3}(\Pi_2)$ is 
\begin{eqnarray*}
holds(l,i,k) & \la & holds(\varphi,i,k)\\
  & & (\caused(l,\varphi) \in \cald) \nonumber\\
holds(l,i,k) & \la & \\
  & & ((l \in \delta_{i{-}1,k}) \vee
       (k' < k  \wedge l \in \delta_{i{-}1,k'})) \nonumber\\
holds(g,i,k) & \la & 
\end{eqnarray*}
or equivalently,
\begin{eqnarray*}
holds(l,i,k) & \la & holds(\varphi,i,k)\\
  & & (\caused(l,\varphi) \in \cald) \nonumber\\
holds(l,i,k) & \la & \\
  & & (l \in \delta_{i{-}1,k'} \cup \{ g \}) \nonumber
\end{eqnarray*}
since if $k'<k$ then $\delta_{i{-}1,k} = \emptyset$.
By Lemma \ref{l2}, this program has the only answer set
$$\{holds(l,i,k) \mid l \in Cl_\cald(\delta_{i{-}1,k'}\cup \{g\}) \}
= \{ holds(l,i,k) \mid l \in \delta_{i,k}\}$$
Hence, $Y^{\{holds\}}_{i,k}$ is the only answer set for
$b_{Z_3}(\Pi_2)$.
\item[c.] {\em $occ(a,i{-}1,k) \not \in Y_{i{-}1}$ for every 
non-sensing action $a$ and $\forall \langle g,k' \rangle. 
br(g,i{-}1,k',k) \not \in Y_{i{-}1}$.}

         From the construction of $Y_{i,k}$'s, it follows that
$e(l,i{-}1,k) \not \in Y_{i{-}1}$ and $pc(l,i{-}1,k) \not \in Y_{i{-}1}$
for every $l$. Hence, $b_{Z_3}(\Pi_2)$ is the following set of rules
\begin{eqnarray*}
holds(l,i,k) & \la & holds(\varphi,i,k)\\
  & & (\caused(l,\varphi) \in \cald) \nonumber\\
holds(l,i,k) & \la & \\
  & & (l \in \delta_{i{-}1,k}) \nonumber
\end{eqnarray*}
whose only answer set is
$$\{holds(l,i,k) \mid l \in \delta_{i{-}1,k}\}= 
\{holds(l,i,k) \mid l \in \delta_{i,k}\} =
Y^{\{holds\}}_{i,k}$$
\end{itemize}
So, in all three cases, we have $Y^{\{holds\}}_{i,k}$
is an answer set for $b_{Z_3}(\Pi_2)$.

Hence, $\Pi_3 = e_{Z_3}(\Pi_2 \setminus b_{Z_3}(\Pi_2), Y^{\{holds\}}_{i,k})$
is the following set of rules:
\begin{eqnarray*}
poss(a,i,k) & \la & \label{lmb1_r12_2}\\
  & & (\executable(a,\psi) \in \cald, \psi \subseteq \delta_{i,k}) \nonumber\\
used(i,k) & \la & \label{lmb1_r12_19}\\
  & & (\exists \langle g,k' \rangle. k' < k ,
      br(g,i{-}1,k',k) \in Y_{i{-}1}) \nonumber\\
goal(i,k) & \la & \label{lmb1_r12_11}\\
  & & (\calg \subseteq \delta_{i,k})\\
goal(i,k) & \la & \label{lmb1_r12_11_2}\\
  & & (\delta_{i,k} \textnormal{ is inconsistent})\\
occ(a_1,i,k) \mid \dots \nonumber\\
  \mid occ(a_m,i,k) & \la & used(i,k), \naf goal(i,k)\label{lmb1_r12_22}\\
used(i,k) & \la & \nonumber \label{lmb1_r12_30}\\
  & & (used(i{-}1,k) \in Y_{i{-}1})
\end{eqnarray*}
It is easy to see that $Y_{i,k}^{\{poss,used,goal,occ\}}$
is an answer set for $\Pi_3$. Accordingly, we have
$M=Y_{i,k}^{\{holds,poss,used,goal,occ\}}$ is an answer set for $\Pi_2$.

$\Pi_4 = e_{Z_2} ({\pi'}_i^k \setminus \Pi_2,M)$ is thus the 
following set of rules:
\begin{eqnarray*}
e(l,i,k) & \la & \label{lmb1_r13_3}\\
  & & (occ(a,i,k) \in M, \causes(a,l,\phi) \in \cald,
      holds(\phi,i,k) \subseteq M) \nonumber\\
pc(l,i,k) & \la & \label{lmb1_r13_4}\\
  & & (occ(a,i,k) \in M, \causes(a,l,\phi) \in  \cald, \nonumber \\
  & & holds(l,i,k) \not \in M, 
      holds(\neg \phi,i,k) \cap M = \emptyset)  \nonumber\\
br(g,i,k,k) \mid \dots & & \nonumber\\
br(g,i,k,w) & \la &  \label{lmb1_r13_6}\\
  & & (occ(a,i,k) \in M, 
      \determines(a,\theta) \in \cald, g \in \theta) \nonumber\\
pc(l,i,k) & \la & pc(l',i,k), 
  \naf e(\neg \varphi,i,k) \label{lmb1_r13_8}\\
  & & (\caused(l,\varphi) \in \cald, holds(l,i,k) \not \in M, 
  l' \in \varphi) \nonumber\\
e(l,i,k) & \la & e(\varphi,i,k) \label{lmb1_r13_9}\\
  & & (\caused(l,\varphi) \in \cald) \nonumber
\end{eqnarray*}
By Lemma \ref{lmb1}, $N = Y_{i,k}^{\{e,pc,br\}}$ is an answer set for
$\Pi_4$.

As a result, $Y_{i,k} = M \cup N$ is an answer set for ${\pi'}_i^k$.
\end{itemize}
\end{proof}
\begin{lemma}
We have
\begin{itemize}
\item[1. ] $S' = \bigcup_{i=1}^{h+1} Y_{i} 
\cup X_0$ is an answer set for $\pi_{h,w}(\cald)$,
where $X_0=V$ is defined in (\ref{def_V}).
\item[2. ] $p^1_1(S') = q$ 
\end{itemize}
\end{lemma}
\begin{proof}
\begin{itemize}
\item[1. ]
Since $Y_{i,k}$ is an answer set for ${\pi'}_i^k$ and ${\pi'}_i^k$'s
are disjoint from each other, we have $Y_i$ is an answer set
for ${\pi'}_i$, where ${\pi'}_i$ is defined in the same way as $\pi_i$ except
that every occurrence of $X$ in Equations (\ref{def_pi_1})
and (\ref{def_pi_i}) is replaced with $Y$. From 
the splitting sequence theorem,
it follows that $S_0' = \bigcup_{i=1}^{h+1} Y_{i}$
is an answer set for $\pi_0$. Thus, $S'$ is an answer set
for $\pi^*_{h,w}(\calp)$. 

On the other hand, it is not difficult to show that
$S'$ satisfies all constraints in $\pi^*_{h,w}(\calp)$
based on the following observations. 
\begin{list}{$\bullet$}{}
\item If $occ(a,i,k) \in Y_{i,k}$ for some sensing action $a$
which occurs in a k-proposition of the form (\ref{knowledge}) 
then there exists $g$
in $\theta$ such that $br(g,i,k,k) \in Y_{i,k}$. Furthermore,
for every $g' \in \theta$, $g'$ does not in $\delta_{i,k}$.
The latter property holds because that $q$ does not contain an action that
senses an already known-to-be-true literal. 
\item If $used(h+1,k) \in Y_{h+1,k}$ then $\delta_{h+1,k} \models \calg$. 
\item $\delta_{i,k}$ is either $\bot$ or an a-state. 
This means that $Y_{i,k}^{\{holds\}}$ does not contain two atoms 
of the forms $holds(l,i,k)$ and $holds(l',i,k)$, where $l$ and $l'$
are contrary literals. 
\item No two branches come to the same node $(i,k)$.
\item If $used(i,k) \in Y_{i}$ then $br(g,i,k',k) \not \in Y_i$
for any pair $\langle g,k' \rangle$, $k' \ne k$.
\item if $\langle a,i,k \rangle \in T_q$ then $a$ must be executable
in $\delta_{i,k}$.
\end{list}
Accordingly, we have $S$ is an answer set for $\pi_{h,w}(\calp)$.
\item[2. ] Immediate from the construction of $Y_{i,k}$.
\end{itemize}
\end{proof}
Theorem \ref{t2} follows directly from this lemma.

\section*{Appendix C -- A Sample Encoding}
This appendix contains the encoding of the planning problem $\calp_1$ in Example
\ref{ex04}. The first subsection describes the input planning problem.
The next subsection presents the corresponding logic program $\pi_{h,w}(\calp_1)$.
The last two subsections are the outputs of \smodels\ and \cmodels when this
logic program is run with the parameters $h=2$ and $w=3$.
\subsection*{Input Domain}
\begin{small}
\begin{Verbatim}
% A possible plan is 
%   check; cases(open-> [];closed->[flip_lock];locked->[])
% fluents
fluent(open).
fluent(closed).
fluent(locked).

% actions
action(check).
action(push_up).
action(push_down).
action(flip_lock).

% executability conditions
executable(check,[]).
executable(push_up,[closed]).
executable(push_down,[open]).
executable(flip_lock,[neg(open)]).

% dynamic laws
causes(push_down,closed,[]).
causes(push_up,open,[]).
causes(flip_lock,locked,[closed]).
causes(flip_lock,closed,[locked]).

% knowledge laws
determines(check,[open,closed,locked]).

% static laws
oneof([open,closed,locked]).

% initial state
initially(neg(open)). % window is not open

% goal
goal(locked). % window is locked
\end{Verbatim}
\end{small}
\subsection*{Encoding}
\begin{small}
\begin{Verbatim}
%%%%%%%%%%%%%%%%%%%%%%%%%%%%%%%%%%%%%%%%%%%%%%%%%%%%%%%%%%%%%%%%%
% Usage:                              
%     lparse -c h=<height> -c w=<width> | smodels
%%%%%%%%%%%%%%%%%%%%%%%%%%%%%%%%%%%%%%%%%%%%%%%%%%%%%%%%%%%%%%%%%
#domain fluent(F).
#domain literal(L;L1).
#domain sense(G;G1;G2).
#domain time(T).
#domain time1(T1).
#domain path(P;P1;P2).
#domain action(A).

% Input parameters
time(1..h).
time1(1..h+1).
path(1..w).

%%%%%%%%%%%%%%%%%%%%%%%%%%%%%%%%%%%%%%%%%%%%%%%%%%%%%%%%%%%%%%%%%
% Action declarations
%%%%%%%%%%%%%%%%%%%%%%%%%%%%%%%%%%%%%%%%%%%%%%%%%%%%%%%%%%%%%%%%%
action(check).
action(push_up).
action(push_down).
action(flip_lock).

%%%%%%%%%%%%%%%%%%%%%%%%%%%%%%%%%%%%%%%%%%%%%%%%%%%%%%%%%%%%%%%%%
% Fluent declarations
%%%%%%%%%%%%%%%%%%%%%%%%%%%%%%%%%%%%%%%%%%%%%%%%%%%%%%%%%%%%%%%%%
fluent(open).
fluent(closed).
fluent(locked).
sense(open).
sense(closed).
sense(locked).

%%%%%%%%%%%%%%%%%%%%%%%%%%%%%%%%%%%%%%%%%%%%%%%%%%%%%%%%%%%%%%%%%
% DOMAIN DEPENDENT RULES 
%%%%%%%%%%%%%%%%%%%%%%%%%%%%%%%%%%%%%%%%%%%%%%%%%%%%%%%%%%%%%%%%%

% Initial situation
holds(neg(open),1,1).

% Executability conditions 
poss(check,T,P).

poss(push_up,T,P) :-
        holds(closed,T,P).
poss(push_down,T,P) :-
        holds(open,T,P).
poss(flip_lock,T,P) :-
        holds(neg(open),T,P).

% Effects of non-sensing actions 
e(closed,T+1,P) :-
        occ(push_down,T,P).
pc(closed,T+1,P) :-
        occ(push_down,T,P).
e(open,T+1,P) :-
        occ(push_up,T,P).
pc(open,T+1,P) :-
        occ(push_up,T,P).
e(locked,T+1,P) :-
        occ(flip_lock,T,P),
        holds(closed,T,P).
pc(locked,T+1,P) :-
        occ(flip_lock,T,P),
        not holds(neg(closed),T,P).
e(closed,T+1,P) :-
        occ(flip_lock,T,P),
        holds(locked,T,P).
pc(closed,T+1,P) :-
        occ(flip_lock,T,P),
        not holds(neg(locked),T,P).

% Effects of sensing actions 
:- occ(check,T,P),
        not br(open,T,P,P),
        not br(closed,T,P,P),
        not br(locked,T,P,P).
1{br(open,T,P,X):new_br(P,X)}1 :-
        occ(check,T,P).
1{br(closed,T,P,X):new_br(P,X)}1 :-
        occ(check,T,P).
1{br(locked,T,P,X):new_br(P,X)}1 :-
        occ(check,T,P).
:- occ(check,T,P),
        holds(open,T,P).
:- occ(check,T,P),
        holds(closed,T,P).
:- occ(check,T,P),
        holds(locked,T,P).

% Static laws 
holds(neg(open),T1,P) :-
        holds(closed,T1,P).

e(neg(open),T+1,P) :-
        e(closed,T+1,P).

pc(neg(open),T+1,P) :-
        pc(closed,T+1,P),
        not holds(neg(open),T,P),
        not e(neg(closed),T+1,P).

holds(neg(open),T1,P) :-
        holds(locked,T1,P).

e(neg(open),T+1,P) :-
        e(locked,T+1,P).

pc(neg(open),T+1,P) :-
        pc(locked,T+1,P),
        not holds(neg(open),T,P),
        not e(neg(locked),T+1,P).

holds(open,T1,P) :-
        holds(neg(closed),T1,P),
        holds(neg(locked),T1,P).

e(open,T+1,P) :-
        e(neg(closed),T+1,P),
        e(neg(locked),T+1,P).

pc(open,T+1,P) :-
        pc(neg(closed),T+1,P),
        not holds(open,T,P),
        not e(closed,T+1,P),
        not e(locked,T+1,P).
pc(open,T+1,P) :-
        pc(neg(locked),T+1,P),
        not holds(open,T,P),
        not e(closed,T+1,P),
        not e(locked,T+1,P).

holds(neg(closed),T1,P) :-
        holds(open,T1,P).

e(neg(closed),T+1,P) :-
        e(open,T+1,P).

pc(neg(closed),T+1,P) :-
        pc(open,T+1,P),
        not holds(neg(closed),T,P),
        not e(neg(open),T+1,P).

holds(neg(closed),T1,P) :-
        holds(locked,T1,P).

e(neg(closed),T+1,P) :-
        e(locked,T+1,P).

pc(neg(closed),T+1,P) :-
        pc(locked,T+1,P),
        not holds(neg(closed),T,P),
        not e(neg(locked),T+1,P).

holds(closed,T1,P) :-
        holds(neg(open),T1,P),
        holds(neg(locked),T1,P).

e(closed,T+1,P) :-
        e(neg(open),T+1,P),
        e(neg(locked),T+1,P).

pc(closed,T+1,P) :-
        pc(neg(open),T+1,P),
        not holds(closed,T,P),
        not e(open,T+1,P),
        not e(locked,T+1,P).
pc(closed,T+1,P) :-
        pc(neg(locked),T+1,P),
        not holds(closed,T,P),
        not e(open,T+1,P),
        not e(locked,T+1,P).

holds(neg(locked),T1,P) :-
        holds(open,T1,P).

e(neg(locked),T+1,P) :-
        e(open,T+1,P).

pc(neg(locked),T+1,P) :-
        pc(open,T+1,P),
        not holds(neg(locked),T,P),
        not e(neg(open),T+1,P).

holds(neg(locked),T1,P) :-
        holds(closed,T1,P).

e(neg(locked),T+1,P) :-
        e(closed,T+1,P).

pc(neg(locked),T+1,P) :-
        pc(closed,T+1,P),
        not holds(neg(locked),T,P),
        not e(neg(closed),T+1,P).

holds(locked,T1,P) :-
        holds(neg(open),T1,P),
        holds(neg(closed),T1,P).

e(locked,T+1,P) :-
        e(neg(open),T+1,P),
        e(neg(closed),T+1,P).

pc(locked,T+1,P) :-
        pc(neg(open),T+1,P),
        not holds(locked,T,P),
        not e(open,T+1,P),
        not e(closed,T+1,P).
pc(locked,T+1,P) :-
        pc(neg(closed),T+1,P),
        not holds(locked,T,P),
        not e(open,T+1,P),
        not e(closed,T+1,P).

%%%%%%%%%%%%%%%%%%%%%%%%%%%%%%%%%%%%%%%%%%%%%%%%%%%%%%%%%%%%%%%%%
% GOAL REPRESENTATION
%%%%%%%%%%%%%%%%%%%%%%%%%%%%%%%%%%%%%%%%%%%%%%%%%%%%%%%%%%%%%%%%%

goal(T1,P) :-
        holds(locked,T1,P).

goal(T1,P) :-
        contrary(L,L1),
        holds(L,T1,P),
        holds(L1,T1,P).

:- used(h+1,P),
        not goal(h+1,P).

%%%%%%%%%%%%%%%%%%%%%%%%%%%%%%%%%%%%%%%%%%%%%%%%%%%%%%%%%%%%%%%%%
% DOMAIN INDEPENDENT RULES 
%%%%%%%%%%%%%%%%%%%%%%%%%%%%%%%%%%%%%%%%%%%%%%%%%%%%%%%%%%%%%%%%%
% Rules encoding the effects of non-sensing actions
holds(L,T+1,P) :-
        e(L,T+1,P).

holds(L,T+1,P) :-
        holds(L,T,P),
        contrary(L,L1),
        not pc(L1,T+1,P).

% Inertial rules for sensing actions
% Cannot branch to the same path
:- P1 < P2,
        P2 < P,
        br(G1,T,P1,P),
        br(G2,T,P2,P).

:- G1 != G2,
        P1 <= P,
        br(G1,T,P1,P),
        br(G2,T,P1,P).

:- P1 < P,
        br(G,T,P1,P),
        used(T,P).

used(T+1,P) :-
        P1 < P,
        br(G,T,P1,P).

holds(G,T+1,P) :-
        P1 <= P,
        br(G,T,P1,P).

holds(L,T+1,P) :-
        P1 < P,
        br(G,T,P1,P),
        holds(L,T,P1).

% Rules for generating action occurrences
1{occ(X,T,P):action(X)}1 :-
        used(T,P),
        not goal(T,P).

:- occ(A,T,P),
        not poss(A,T,P).

% Auxiliary Rules
literal(F).
literal(neg(F)).

contrary(F,neg(F)).
contrary(neg(F),F).

new_br(P,P1) :-
        P <= P1.

used(1,1).
used(T+1,P) :-
        used(T,P).

%%%%%%%%%%%%%%%%%%%%%%%%%%%%%%%%%%%%%%%%%%%%%%%%%%%%%%%%%%%%%%%%%
% HIDE/SHOW ATOMS 
%%%%%%%%%%%%%%%%%%%%%%%%%%%%%%%%%%%%%%%%%%%%%%%%%%%%%%%%%%%%%%%%%
hide.
show occ(A,T,P).
show br(G,T,P,P1).
\end{Verbatim}
\end{small}
\subsection*{Smodels Output}
\begin{small}
\begin{Verbatim}
$ lparse -c h=2 -c w=3 examples/ex2.smo | smodels

smodels version 2.28. Reading...done
Answer: 1
Stable Model: 
br(open,1,1,2) occ(check,1,1) br(closed,1,1,1) 
br(locked,1,1,3) occ(flip_lock,2,1) 
True
Duration: 0.020
Number of choice points: 2
Number of wrong choices: 0
Number of atoms: 313
Number of rules: 893
Number of picked atoms: 257
Number of forced atoms: 31
Number of truth assignments: 4052
Size of searchspace (removed): 12 (65)
\end{Verbatim}
\end{small}
\subsection*{Cmodels Output}
\begin{small}
\begin{Verbatim}
$ lparse -c h=2 -c w=3 examples/ex2.smo | cmodels

cmodels
cmodels version 3.01 Reading...done
Program is not tight.
Calling SAT solver mChaff...
Answer: 1 
Answer set: br(open,1,1,3) occ(check,1,1) br(closed,1,1,1) 
br(locked,1,1,2) occ(flip_lock,2,1) 
Number of Loop Formulas 6
\end{Verbatim}
\end{small}
\bibliographystyle{acmtrans}

\end{document}